\documentclass[11pt]{article} 
\usepackage[paperheight=9.44in,paperwidth=6.69in,left=1.85cm,right=1.8cm,bottom=2.5cm,top=3cm,twoside]{geometry}

\usepackage{subfig} 

\usepackage[linesnumbered,ruled,vlined,boxed]{algorithm2e}
\usepackage{authblk} 
\usepackage{fancyhdr} 
\usepackage{lastpage} 
\usepackage{textcomp} 
\usepackage{amsmath} 
\usepackage{amssymb} 
\usepackage{amsthm} 
\usepackage{graphicx} 
\usepackage{multirow} 
\usepackage{hyperref} 
\usepackage{longtable} 
\usepackage{float} 

\newcommand{\linia}{\noindent\rule{\linewidth}{0.5mm}\hrulefill} 

\SetAlFnt{\small}
\SetAlCapFnt{\small}
\SetAlCapNameFnt{\small}
\SetAlCapHSkip{0pt}
\IncMargin{-\parindent}

\usepackage{times}
\usepackage{titlesec}

\titleformat*{\section}{\large\bfseries}
\titleformat*{\subsection}{\normalsize\bfseries}

\fancypagestyle{firststyle}
{
	\fancyhf{}
	\fancyfoot{}
	\fancyhead{}
	\fancyhead[CE]{\upshape Zeszyty Naukowe WWSI, No 18, Vol. 12, 2018, pp. \thepage-\pageref{LastPage} \newline DOI: 10.26348/znwwsi.18.45}
	\fancyhead[CO]{\upshape Zeszyty Naukowe WWSI, No 18, Vol. 12, 2018, pp. \thepage-\pageref{LastPage} \newline DOI: 10.26348/znwwsi.18.45}
	\fancyfoot[L,R]{\footnotesize\bfseries Manuscript received June 9, 2018}
	\fancyfoot[LE,RO]{\thepage}
	}

\title{\large \bfseries On the clustering of correlated random variables} 
\author{\normalsize Zenon Gniazdowski\thanks{E-mail: zgniazdowski@wwsi.edu.pl} }
\author{\normalsize Dawid Kaliszewski}
\affil{\normalsize Warsaw School of Computer Science}

\date{\vspace{-5ex}}
\providecommand{\keywords}[1]{\textbf{\textit{Keywords ---}} #1}

\begin{document}

\maketitle 
\thispagestyle{firststyle} 

\linia
\begin{abstract}
	\noindent In this work, the possibility of clustering correlated random variables was examined, both because of their mutual similarity and because of their similarity to the principal components.
	The k-means algorithm and spectral algorithms were used for clustering.
	For spectral methods, the similarity matrix was both the matrix of relation established on the level of correlation and the matrix of coefficients of determination.
	For four different sets of data, different ways of measuring the disimilarity of variables were analyzed, and the impact of the diversity of initial points on the efficiency of the k-means algorithm was analyzed.

\end{abstract}
\keywords{\small similarity of variables, clustering of variables, vertical clustering}

\section{Introduction}
In cluster analysis, the set of elements is divided into subsets of similar elements.
In a given subset, there should be those elements that are mutually similar, but at the same time they are different from elements from other groups.
In practice, this is the division of a set of points located in a multidimensional space into subsets of similar points \cite{Hand2001}.
By analogy to the set of points on the plane, their clustering can be called horizontal clustering \cite{Gniazdowski2017}.
Horizontal clustering seems to be one of the important problems of data analysis.
Many algorithms have been developed for his needs.
Examples include the k-means algorithm and hierarchical algorithms \cite{Hand2001} \cite{Larose2005}, as well as spectral algorithms \cite{VonLuxburg2007}.

On the other hand, points in a multidimensional space by their coordinates represent some variables.
If variables are correlated, then (maybe) they are mutually similar.
If they are similar, then the similarity can be identified. In other words, correlated variables can be clustered.
To distinguish from horizontal clustering of points, clustering of variables will be called vertical clustering \cite{Gniazdowski2017}.
This work concerns the possibility of vertical clustering of correlated random variables.
Correlated random variables will be clustered due to their mutual similarity, and will be clustered due to their similarity to previously identified principal components.

The problem covered by the article has its history, which dates back to 2011.
At that time, a question was asked about the relationship between the results of the principal component method and the relation established on the basis of the level of significance of correlation coefficients \cite{Gniazdowski2011}: the relation occurs when the correlation coefficient is significant at a given level of significance.
Unfortunately, this approach had some weaknesses.
The significance level does not provide information on the intensity of the correlation.
When the sample is large, it is easy to show the statistical significance of poor correlation \cite{Blalock1960}.
Information on the intensity of the correlation can be rather found in the value of the angle between random vectors and not in the level of significance \cite{Gniazdowski2013}.
Therefore, a more realistic attempt to answer the above question was made in the master thesis \cite{Weremiejewicz2014}.
In this case, the relation was built based on the angle between random vectors.
It was assumed that random variables are similar when the value of the angle between their random components estimated on the basis of the correlation coefficient was not greater than $\left| \pi / 4\right|$.
It was examined whether the number of connected components of the relation graph is equal to the number of necessary principal components.
The connected components of the graph were counted from its graphical representation, and the number of necessary principal components was estimated, for example, with the use of a scree plot.
In retrospect, it should be recognized that this approach was also quite naive, and the results obtained at that time were not very promising.

A breakthrough in the perception of the problem could occur when in \cite{Gniazdowski2017} it was noted that the coefficient of determination describes a common variance of two random variables.
This common variance can be a very good measure of the similarity of random variables.
This leads to the following conclusions:
\begin{itemize}
	\item The principal components method allows clustering of primary variables due to their similarity to the principal components \cite{Gniazdowski2017}. Clustering can be implemented using the k-means method.
	\item The mutual similarity between the correlated random variables can be described by a binary relation established on the level of correlation.
	      Between two variables, a relation occurs when the value of the determination coefficient between these variables reaches at least an established level.
	      Such a relation will divide random variables into subsets of similar variables.
	      The division of variables will appear in the division of the relation graph into connected components.
	      The nodes from the individual connected components will form disjoint clusters. The relation matrix can be used for clustering using the spectral method.
	\item The coefficients of determination between random variables describe their mutual similarity.
	      This similarity is a generalization of the similarity described by the relation.
	      The matrix of determination coefficients is a similarity matrix that can be used for clustering using spectral methods.
\end{itemize}
As a consequence of the three conclusions presented here, correlated random variables can be clustered in three ways.
For each of these three ways, their variants may also be tested, and then conclusions on their efficiency may be formulated.
In this context, there are several questions that should be answered in the article:
\begin{enumerate}
	\item Is there a relationship between the efficiency of clustering on $k$ clusters and the type of relation that is obtained for the lower limit of the assumed threshold $\epsilon$, at which the relation graph has $k$ connected components?
	\item Is the efficiency of clustering greater for spectral methods or for methods that analyze the similarity of primary variables to principal components?
	\item Which distance measurement method will allow more effective clustering of variables: the Euclidean distance measure or the cosine measure of dissimilarity?
	\item Does the entropy of the set of initial points for the k-means algorithm affect the results of clustering?
\end{enumerate}
First, the answers to the questions presented here should be searched in the context of the analyzed datasets. Next, you can try to draw conclusions that are more universal and are not limited to specific datasets.

\section{Preliminaries}
To discuss the problem, the most important concepts and algorithms used in this work should be presented. Based on them, the variables will be clustered, and then the results of clustering will be compared. The assumptions as well as the accepted symbols will also be presented.
\subsection{Binary relation}
The relation is a subset of the Cartesian product of two non-empty sets \cite{Rasiowa1973}\cite{Ross1992}. It is considered a set $X = \left\lbrace x_1, x_2, ..., x_n\right\rbrace $, consisting of $n$ elements. A binary relation $\rho$ on a set $X$ is a subset of the Cartesian product $X^2 = X \times X$. This relation can be represented in the form of a square binary matrix $R$ of size $n \times n$. Its elements $R_{ij}$ are defined by the following formula:
\begin{equation}\label{eq01}
	\forall_{i,j} R_{ij}=
	\begin{cases}
		1, \text{when } {x_i} {\rho} {x_j} \\
		0, \text{when } \neg \left( x_i \rho x_j\right)
	\end{cases}.
\end{equation}

\noindent The binary relation can have the following properties: it can be reflexive, irreflexive, symmetric, antisymmetric, transitive and connex \cite{Rasiowa1973} \cite{Ross1992}. The properties of the relation will be manifested in the matrix.
From the point of view of this article, the reflexivity and symmetry of the relation, as well as - if it occurs - transitivity are important:
\begin{itemize}
	\item Relation $R$ on the set $X$ is reflexive when $\forall{x}\in{X}:x \rho x$. In matrix notation, the reflexivity manifests itself in the fact that all elements on the diagonal of matrix $R$ are equal to $1$.
	\item Relation $R$ on the set $X$ is symmetrical when $\forall{x,y}\in{X}: x \rho y \Rightarrow y \rho x$. In the matrix notation, the symmetry of the relation manifests itself in the symmetry of the relation matrix: $R = R ^ T$.
	\item Relation $R$ on the set $X$ is transitive, if $\forall {x,y,z} \in{X}: \left( x \rho y \land y \rho z\right) \Rightarrow x \rho z$. The matrix of this relation is also the adjacency matrix of the graph of this relation. So, the condition of transitivity can also be formulated in the language of graph theory: if between two nodes of the graph of relation there is a path with a length of two arcs, then in the graph of the transitive relation, between these nodes there exists a path with a length of one arc. In other words: each path with a length of two arcs should be accompanied by a shortcut with the length of one arc.The square of the adjacency matrix counts all paths with the length of two arcs \cite{Ross1992}.Because for testing of transitivity there is not essential how many paths with a length of two arcs are there, but whether such paths exist, therefore the matrix multiplication $R^2$ can be replaced by a Boolean multiplication $R * R$. The condition of transitivity of relation is fulfilled when for each $(R*R)_{ij}=1$ there is $R_{ij}=1$. This condition can be expressed as follows: $R * R \le R$ \cite{Ross1992}.

\end{itemize}
The properties of a binary relation determine its type:
\begin{itemize}
	\item If the relation is reflexive, symmetric, and transitive, it is an equivalence relation. Equivalence relation divides the set into disjoint classes of equivalence. This relation allows to investigate whether the two elements in the set are equivalent in the sense of relation (belong to the same equivalence classes), or are not equivalent (belong to different classes of equivalence) \cite{Rasiowa1973}\cite{Ross1992}.
	\item If the relation is reflexive and symmetrical, and it is not transitive, then it is called a relation of similarity (or tolerance) \cite{Peters2012}. The similarity relation will divide the set into subsets of similar objects. By analogy to the equivalence classes, these subsets can be called similarity classes.
\end{itemize}
Finally, it can be seen that the relation of equivalence because it is reflexive and symmetrical is a special case of the relation of similarity.

\subsection{Binary relation established on the level of correlation}
The level of the correlation (or similarity) between the random variables is represented by the matrix of determination coefficients \cite{Gniazdowski2017}.
Assuming the value of the similarity threshold $\epsilon$, it is possible to construct a matrix of relation represented by the so-called "$\epsilon$-neighborhood graph" \cite{VonLuxburg2007}. If the similarity between two different variables $v_i$ and $v_j$ is not less than the assumed threshold value $\epsilon$, then the variables $v_i$ and $v_j$ are considered to be in relation to each other. This means that in the appropriate place in the relation matrix $R_{c \times c}$ it is inserted the value of $1$: $R_{ij}:=1$, $R_{ji}:=1$. Otherwise, a zero value is inserted: $R_{ij}:=0$, $R_{ji}:=0$. Because each variable is identical (indistinguishable) from itself, so the diagonal of the matrix will be set by unit values.

The value of $\epsilon$ determines the limit value of the determination coefficient, sufficient for the existence of a relation. The assumed value of $\epsilon$ can not be too small or too big. If the value of $\epsilon$ is too small, there will be more unit values in the relation matrix. In this case, the relation may have too few equivalence (or similarity) classes. This will manifest itself in a small amount of connected componenst of the graph.
When the threshold value is higher, then there will be more zeros in the matrix of the relation, but similarity or equivalence classes (connected components of the graph) may also arrive. It is more likely that isolated nodes can appear in the graph.

It is assumed that the variables are similar when they have not less than half of the common variance measured by the coefficient of determination \cite{Gniazdowski2017}. Therefore, threshold values $\epsilon$ close to $0.5$ are accepted. If the graph has too many or too few connected components, threshold values $\epsilon$ lower or higher than $0.5$ may be considered.

The obtained matrix $R_ {c \times c}$ represents the reflexive relation (only unit values on the diagonal) and symmetrical relation (symmetric matrix). The reflexive and symmetric relation is a similarity relation \cite{Peters2012}, which divides the set into subsets of similar objects (in the sense of the accepted similarity threshold). These subsets are the mentioned above similarity classes. If this relation is transitive, then it is the equivalence relation that divides the set into equivalence classes.

\subsection{Spectral clustering}\label{Luxburg}
The description of spectral clustering methods is discussed in \cite{VonLuxburg2007}. This article will use some of the ideas contained there.
It is assumed that there is a square symmetric similarity matrix $S$, defined for a certain $c$-element set of objects $v = {v_1, v_2, ... v_c}$. In particular, as a similarity matrix the adjacency matrix of the graph of similarity relation can be considered. The nodes of this graph are objects from the set $v$. The value $s_{ij}$ is equal to unit value when the nodes $v_i$ and $v_j$ are connected by an arc. Clustering consists in dividing the set of nodes $v$ into disjoint subsets of similar nodes. For this purpose, for the matrix $S$, a diagonal matrix $D$ of degrees of the nodes is defined. Its diagonal elements $D_{ii}$ are equal:
\begin{equation}\label{eq02}
	D_{ii}=\sum_{j=1}^{c}S_{ij}.
\end{equation}
For the graph, the Laplacian matrix $L$ is also defined:
\begin{equation}\label{eq03}
	L=D-S.
\end{equation}
For Laplacian matrix L, the eigenproblem is solved. The smallest eigenvalue of the Laplacian matrix has a zero value. If the graph is not connected, the number of its connected components is equal to the number of the Laplacian matrix zero eigenvalues.

To find $k$ clusters, one should use $k$ eigenvectors corresponding to $k$ the smallest eigenvalues.
From these vectors, a rectangular matrix $M_ {c \times k}$ containing $c$ rows and $k$ columns is formed.
The columns of the matrix will be suitable eigenvectors, and the rows represent the coordinates of points, each of which corresponds to the further objects $v_i$ $(i = 1, 2, \dots , c)$. By clustering these points using the k-means algorithm, it will be possible to divide the nodes between the respective clusters \cite{VonLuxburg2007}.

For clustering symmetric normalized Laplacian $L_n$ can also be used:
\begin{equation}\label{eq04}
	L_n=D^{-1/2} LD^{-1/2}=I-D^{-1/2} SD^{-1/2}.
\end{equation}
The algorithm of clustering using normalized Laplacian $L_n$ is analogous to the algorithm used for Laplacian matrix $L$.
The only difference is that the length of the row vectors in the rectangular matrix $M_ {c \times k}$ used for k-means clustering are normalized to the unity \cite{VonLuxburg2007} \cite{Ng2002}.

It should be noted that the similarity matrix may also be a matrix of determination coefficients between the random variables.
This matrix, being a generalization of the graph adjacency matrix, can be interpreted as a matrix of a graph whose arcs have been assigned certain non-negative weights.
With this assumption, the method of finding the degree matrix, as well as Laplacian matrix and normalized Laplacian will not change, and as a consequence the algorithm of clustering will not change.

In this paper, when we talk about spectral clustering, the methods that use Laplacian of similarity (or relation) matrix as well as normalized Laplacian of similarity (or relation) matrix will be used.
\subsection{Assumptions for the k-means algorithm}
The k-means algorithm will be used to clustering variables due to the similarity of the primary variables to the principal components. On the other hand, clustering due to the mutual similarity of the primary variables will be carried out using the spectral analysis of Laplacian and normalized Laplacian.
Spectral methods in the final stage of their operation also use the k-means algorithm. In all cases, the k-means algorithm is used with some assumptions, for example, such as the number of clusters or the metrics used (the measure of dissimilarity).
These assumptions will be briefly presented below.
\subsubsection{The number of clusters}
The number of clusters should be determined before clustering.
It is assumed that for clustering, due to similarity of primary variables to principal components, the number of clusters is not less than the number of principal components needed to explain at least half of the variance of the primary variables.
In the case of the similarity relation, the number of clusters will be equal to the number of connected components of the graph. This number may evolve depending on the assumed similarity threshold $\epsilon$.
The value $\epsilon$ should be chosen so that in the case of clustering due to the similarity of primary variables to the principal components, as well as in the case of clustering based on the relation of similarity, the number of clusters is identical.
This will allow to compare the efficiency of clustering methods.
\subsubsection{Measures of dissimilarity}\label{Measures}
The k-means algorithm can use various metrics (or more generally, the measures of dissimilarity).
In this work, the Euclidean metric will be used to measure the distance (dissimilarity) between the points, but also the cosine measure of dissimilarity will also be taken into account.

The distance between two points $a$ and $b$, where $a = (a_1, a_2, \ldots , a_n)$ and $b = (b_1, b_2, \ldots , b_n)$, can be measured using the Euclidean metric:
\begin{equation}\label{eq05}
	d(a,b)= \left\| a-b\right\| _2=\sqrt{\sum_{i=1}^{n}(a_i-b_i)^2}.
\end{equation}
If the point in space is treated as a vector acting in the center of the coordinate system, then to estimate the similarity between the points one can use the cosine of the angle between the vectors representing them:
\begin{equation}\label{eq06}
	cos(a,b)=\frac{a\cdot b}{\left\| a\right\| _2\cdot \left\| b\right\| _2}.
\end{equation}
For random vectors, the correlation coefficient is the same as the cosine of the angle between these random vectors \cite{Gniazdowski2013}.
On the other hand, the square of the correlation coefficient, called the coefficient of determination, measures the common variance of both random vectors and is a measure of their similarity \cite{Gniazdowski2017}.
As the coefficient of determination is a measure of similarity between two random vectors, so the distance between two points in space is a measure of their dissimilarity. Looking for the cosine analogue of dissimilarity, the following measure of the dissimilarity between vectors will be accepted in this work:
\begin{equation}\label{eq07}
	d(a,b)=1-cos^2(a,b).
\end{equation}
For vectors without normalization of their length to unity, as well as after its normalization, the cosine of the angle between them will be identical.
Thus, there is no need to consider a cosine measure (\ref{eq07}) for both cases.
In turn, for normalized and non-normalized points, the Euclidean distance may be different.
Therefore, Euclidean distance can be considered for both non-normalized points and for normalized points.
When clustering using the k-means method due to the similarity of the primary variables to the principal components, the above facts will be taken into account.

On the other hand, spectral algorithms will be used according to their specifications given in section \ref{Luxburg}.
In both cases, the two measures of dissimilarity will be used: both the Euclidean metric and the cosine measure of dissimilarity.

\subsubsection{Diversity of initial points}\label{Entropy}
To find $k$ clusters, it should be chosen $k$ different initial points in this space.
When choosing $k$ initial points, points different in pairs should be selected.
Intuition suggests that if any identical points were chosen among $k$ initial points, the method will not behave as expected.
For the $n$-element set of points, the number of all possible different k-element subsets of points, which can be initial points in the k-means algorithm, can be calculated from the formula:
\begin{equation}\label{eq07a}
	\binom{n}{k}=\frac{n!}{k!(n-k)!}.
\end{equation}
Depending on their cardinality, there may be two strategies for checking the efficiency of clustering.
If the number of all combinations of initial points (\ref{eq07a}) is not too large, then the clustering procedure can be run for all possible subsets of initial points.
The algorithm presented in \cite{Lipski2004} can be used to find these subsets.
For too many of these subsets, the initial points should be random.

The assumption that the initial points will be different in pairs is the weakest assumption.It seems that the points should be maximally diversified.
As a measure of the diversity of points one can accept the entropy of the system that creates the selected points \cite{Grabowski2017}.
To calculate the entropy for a given sampling series, the sum of the distances between them is calculated for all pairs of points:
\begin{equation}\label{eq08}
	S=\sum_{i,j;i>j}d(x_i,y_j).
\end{equation}
For each pair of points the probability is assigned:
\begin{equation}\label{eq09}
	p_{ij}=\frac{d_{ij}}{S}.
\end{equation}
The value of entropy for a given sampling is estimated from the formula:
\begin{equation}\label{eq10}
	E=\sum_{i,j;i>j}(-p_{ij}*log(p_{ij})).
\end{equation}

\subsubsection{The symbols used to describe the clustering options}
In this work clustering of correlated random variables will be described.
For these variables, matrices describing the mutual similarity of the primary variables and the similarity of the primary variables to the principal components will be estimated, as well as matrices explicitly used for spectral clustering.
These matrices will be described using the following symbols:
\begin{itemize}
	\item $C_R$ - a matrix of correlation coefficients;
	\item $S$ - matrix of determination coefficients which describes the mutual similarity of primary variables;
	\item $R_\epsilon$ - relation matrix formed on the basis of the similarity matrix $S$, for the assumed similarity threshold $\epsilon$;
	\item $P$ - matrix containing the coefficients of determination between the primary variables and the principal components;
	\item $L$, $nL$, $L\epsilon$ or $nL\epsilon$ - matrices used by spectral clustering algorithms:
	      \begin{itemize}
		      \item $L$ - the Laplacian of coefficients of determination matrix,
		      \item $nL$ - normalized Laplacian of coefficients of determination matrix,
		      \item $L\epsilon$ - Laplacian of the relation matrix for the similarity threshold $\epsilon$.
		      \item $nL\epsilon$ - normalized Laplacian of the relation matrix for the similarity threshold $\epsilon $.
	      \end{itemize}
\end{itemize}
In addition, it is assumed the following notation:
\begin{itemize}
	\item $k$ - number of clusters;
	\item $E$ or $C$ - the Euclidean metric (\ref{eq05}) or the cosine measure of dissimilarity (\ref{eq07}), respectively;
	\item $m/n$ - $m$ out of $n$ cases, where $n$ is the number of clustered variables, and $m$ is the number of variables correctly classified into clusters.
\end{itemize}
The algorithm of k-means in different variants will be used in thise work.
The following symbols are used to describe these variants:
\begin{itemize}
	\item $kEP$ - $k$ clusters, Euclidean metric, clustering according to similarity of primary variables to principal components;
	\item $kCP$ - $k$ clusters, cosine measure of dissimilarity, clustering according to similarity of primary variables to principal components;
	\item $kEnP$ - $k$ clusters, Euclidean metric, clustering according to similarity of primary variables to principal components, points to clustering normalized to unit length;
	\item $kEL$ - $k$ clusters, Euclidean metric, clustering based on Laplacian of coefficients of determination matrix;
	\item $kCL$ - $k$ clusters, cosine measure of dissimilarity, clustering based on Laplacian of coefficients of determination matrix;
	\item $kEnL$ - $k$ clusters, Euclidean metric, clustering based on the normalized Laplacian of coefficients of determination matrix;
	\item $kCnL$ - $k$ clusters, cosine measure of dissimilarity, clustering based on normalized Laplacian matrix of coefficients of determinantion;
	\item $kEL\epsilon$ - $k$ clusters, Euclidean metric, clustering based on the Laplacian matrix of the relation obtained for the similarity threshold $\epsilon$;
	\item $kCL\epsilon$ - $k$ clusters, cosine measure of dissimilarity, clustering based on the Laplacian matrix of the relation obtained for the similarity threshold $\epsilon$;
	\item $kEnL\epsilon$ - $k$ clusters, Euclidean metric, clustering based on the normalized Laplacian matrix of the relation obtained for the similarity threshold $\epsilon$;
	\item $kCnL\epsilon$ - $k$ clusters, a cosine measure of dissimilarity, clustering based on the normalized Laplacian matrix of the relation obtained for the similarity threshold $\epsilon$;
	\item $kMan$ - containing $k$ clusters arbitrarily accepted pattern for comparing the efficiency of clustering using various methods.
\end{itemize}
\subsection{Other assumptions regarding data analysis}
To answer the questions posed in the introduction, research will be carried out for four different datasets.
For each dataset, clustering algorithms will be run for many different sets of initial points.
In this article, it is assumed that for a number of combinations of initial points not greater than $300$, the clustering procedures will be run for all possible combinations of initial points.
If the number of combinations of initial points would be higher, the initial points will be randomly generated $300$ times.

When using the clustering procedure, a reference set of clusters is needed to compare the clusters received as a result of this procedure.
In this paper, it is assumed that the distribution of nodes obtained as a result of this procedure will be compared with the distribution of nodes between the connected components of the graph of a certain relation, defined for the given similarity threshold $\epsilon$.
The threshold will be set so that the number of connected components of the relation graph is equal to the specified number of clusters $k$.
It should be noted that although the number of connected components is unchanged for a certain range of $\epsilon$ values, within this interval the properties of relation may change.

In this work, the influence of entropy of initial points for the k-means algorithm on the efficiency of clustering will be investigated.
To do this, the clustering efficiency distributions obtained for the studied population of sets containing initial points will be compared with the clustering efficiency distributions for a part of this population.
Approximately one third of the population will be selected for this purpose, containing sets of initial points with the largest entropy.

The article uses the results obtained from numerical calculations.
Therefore, for the presentation of numbers in floating point notation some rules will be applied that will have their justification in the considered context:
\begin{itemize}
	\item Basically, numbers will be presented with three digits after the decimal point. If it could happen that the table or matrix does not fit on the page width, the number of digits after the decimal point can be reduced.
	\item In the case of percentages, if digits after the decimal point are necessary, one digit after the decimal point will be displayed.
	\item Numbers from the range [0,1], depending on the context, will be given in absolute form or in percentages. And so, in the context of a common variance, the coefficients of determination can be given in percent. In turn, in the context of the content of a matrix containing points for clustering with the k-means algorithm, the same coefficients of determination will be given in absolute numbers.
	\item The threshold of the coefficient of determination $\epsilon$, at which the relation between variables takes place, depending on the context can be given in absolute numbers or in percentages.
\end{itemize}

\section{The dataset No. 1 - Iris Data}
Data for analysis will be the classic data of iris flowers, proposed in 1936 by Sir Ronald Fisher \cite{Fisher1936}.
An in-depth analysis of the principal components for this set was carried out in \cite{Gniazdowski2017}.
Here, the table of determination coefficients will be used, describing the similarity of the primary variables to the principal components, as well as the matrix of determination coefficients, which describes the mutual similarity of the primary variables.
\begin{table}
	\centering
	\caption{Coefficients of determination between primary variables and principal components for Iris Data}\label{table01}
	\fontsize{9.5}{13.5}\selectfont{
		\begin{tabular}{c||c|c|c|c} \hline
			    & Sepal Length & Sepal Width & Petal Length & Petal Width \\ \hline \hline
			PC1 & $81.2\%$     & $12.9\%$    & $97.2\%$     & $93.6\%$    \\ \hline
			PC2 & $9.7\%$      & $86.3\%$    & $0.1\%$      & $0.1\%$     \\ \hline
			PC3 & $9.0\%$      & $0.8\%$     & $0.7\%$      & $5.2\%$     \\ \hline
			PC4 & $0.1\%$      & $0.0\%$     & $2.0\%$      & $1.1\%$     \\ \hline
		\end{tabular}}
\end{table}
\subsection{Similarity of primary variables to principal components}
The coefficients of determination between the primary variables and the principal components are shown in Table \ref{table01}.
In the case of clustering due to the similarity of the primary variables to the principal components, the number of clusters can not be less than the number of principal components needed to explain more than half the variance of each primary variable.
Because the two principal components explain more than $90\%$ of the variance of each of the primary variables (Table \ref{table02}), two clusters are enough.

\begin{figure}
	\centering
	\includegraphics[width=6.5cm]{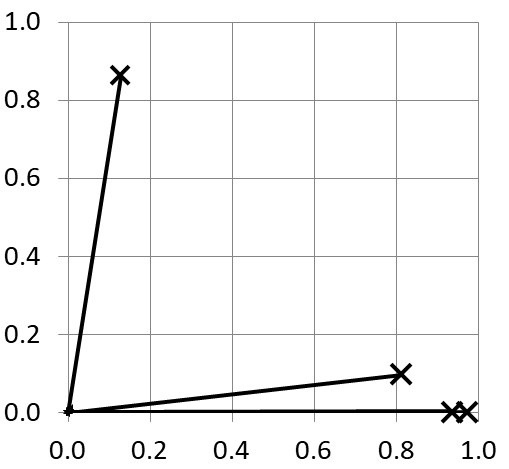}
	\caption{Clustering points (\ref{eq12}) plotted on a plane}\label{figure01}
\end{figure}

\begin{table}
	\centering
	\caption{The level of reconstruction of primary variables for Iris Data}\label{table02}
	\fontsize{9.5}{13.5}\selectfont{
		\begin{tabular}{c||c|c|c|c} \hline
			         & Sepal Length & Sepal Width & Petal Length & Petal Width \\ \hline \hline
			PC1      & $81.2\%$     & $12.9\%$    & $97.2\%$     & $93.6\% $   \\ \hline
			PC2      & $9.7\%$      & $86.3\%$    & $0.1\%$      & $0.1\% $    \\ \hline \hline
			$\Sigma$ & $90.8\%$     & $99.2\%$    & $97.3\%$     & $93.7\% $   \\ \hline
		\end{tabular}}
\end{table}
Clustering can be accomplished with the use of the first two rows of the table containing the coefficients of determination between the primary variables and principal components (Table \ref{table01}).
The matrix $M$ containing the points to be clustered has the form:
\begin{equation}\label{eq12}
	M {=}
	\begin{bmatrix}
		0.812 & 0.097 \\
		0.129 & 0.863 \\
		0.972 & 0.001 \\
		0.936 & 0.001
	\end{bmatrix}.
\end{equation}
The rows in the matrix $M$ are points in space, in which the clusters are located.
Because the space is two-dimensional, these points can be represented on the plane (Figure \ref{figure01}).
The figure shows that they are clustered into two clusters.

\begin{figure}
	\centering
	\includegraphics[width=6.5cm]{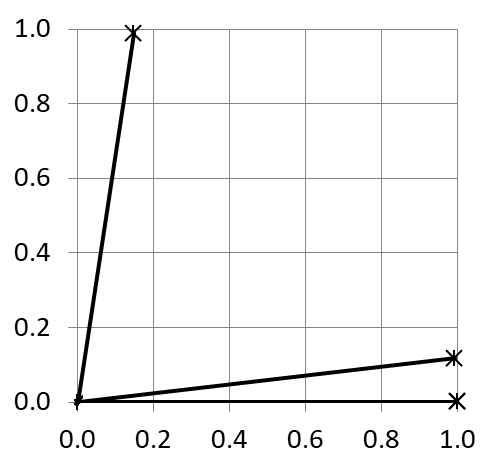}
	\caption{Normalized points for clustering (\ref{eq13}) plotted on a plane}\label{figure02}
\end{figure}

It is also possible to cluster the points, if they are projected onto a circle with the center at point $(0,0)$ and the unitary radius.
Projection is equivalent to normalizing row vectors in the $M$ matrix to the unitary length.
After normalization, the matrix used for clustering takes the form:
\begin{equation}\label{eq13}
	M {=}
	\begin{bmatrix}
		0.993 & 0.118 \\
		0.148 & 0.989 \\
		1.000 & 0.001 \\
		1.000 & 0.001
	\end{bmatrix}.
\end{equation}
Figure \ref{figure02} shows the normalized points on the plane.
All four points are now on the circle and are clustered into two clusters.
\subsection{Mutual similarity of primary variables that is described by the binary relation determined at the level of correlation}
\begin{figure}
	\centering
	{\subfloat[] {\label{figure03a}
			\includegraphics[width=0.30\textwidth]{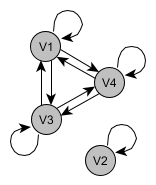}
		}
		\quad
		\quad
		\quad
		\subfloat[] {\label{figure03b}
			\includegraphics[width=0.30\textwidth]{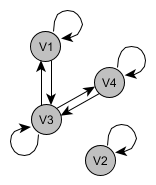}
		}
		\caption{Fragment of the evolution of relation: (a) $\epsilon = 50\%$, equivalence relation (\ref{eq14}) with two equivalence classes; (b) $\epsilon = 70\%$, similarity relation (\ref{eq15}) with two classes of similarity}
		\label{figure03}}
\end{figure}

The matrix of determination coefficients that describes the mutual similarity of primary variables has the form:
\begin{equation}\label{eq11}
	S {=}
	\begin{bmatrix}
		1     & 0.004 & 0.749 & 0.666 \\
		0.004 & 1     & 0.103 & 0.090 \\
		0.749 & 0.103 & 1     & 0.920 \\
		0.666 & 0.090 & 0.920 & 1
	\end{bmatrix}.
\end{equation}
This matrix for variables describing iris flowers allows to define a binary relation.
The threshold for defining this relation is arbitrarily accepted coefficient $\epsilon$.
Depending on the value of $\epsilon$, different relations can be obtained.
For $\epsilon = 50\%$, the relation matrix has the form:
\begin{equation}\label{eq14}
	R_{\epsilon=50\%} {=}
	\begin{bmatrix}
		1 & 0 & 1 & 1 \\
		0 & 1 & 0 & 0 \\
		1 & 0 & 1 & 1 \\
		1 & 0 & 1 & 1\end{bmatrix}.
\end{equation}
For $\epsilon = 70\%$, the relation matrix takes the form:
\begin{equation}\label{eq15}
	R_{\epsilon=70\%} {=}
	\begin{bmatrix}
		1 & 0 & 1 & 0 \\
		0 & 1 & 0 & 0 \\
		1 & 0 & 1 & 1 \\
		0 & 0 & 1 & 1\end{bmatrix}.
\end{equation}
Figure \ref{figure03} shows two graphs\footnote{In this work, the yEd Graph Editor ver. 3.18.1.1 was used for graph presentation.} of the binary relation obtained for two different $\epsilon$ values, which represent a fragment of the evolution of this relationship.
In both cases, the allocation of nodes to the connected components of the graph is identical. This allocation of nodes, marked as 2Man, will be used as a reference allocation to compare the efficiency of clustering using different methods.

Clustering nodes of relations (\ref{eq14}) and (\ref{eq15}) is equivalent to the identification of connected components of graphs shown in Figures \ref{figure03a} and \ref{figure03b}.
They can be done with the use of spectral analysis of the Laplacian $L$ or by using spectral analysis of the normalized Laplacian $L_n$, both found for the adjacency matrix of the relation graphs (\ref{eq14}) or (\ref{eq15}).
Here, the clustering procedures for relation (\ref{eq14}) obtained for the assumed value of $\epsilon = 50\%$ will be presented.

\subsubsection{Laplacian of the relation graph}

For the relation (\ref{eq14}), the Laplacian $L$ was formed:
\begin{equation}\label{eq16}
	L {=}
	\begin{bmatrix}
		2  & 0 & -1 & -1 \\
		0  & 0 & 0  & 0  \\
		-1 & 0 & 2  & -1 \\
		-1 & 0 & -1 & 2\end{bmatrix}.
\end{equation}
Its eigenvalues are equal:
\begin{equation}\label{eq17}
	\lambda {=} \{3,3,0,0\}.
\end{equation}
Two eigenvalues are equal to zero.
This means that the relation graph consists of two connected components.
The corresponding eigenvectors form rows of the following matrix:
\begin{equation}\label{eq18}
	V^T {=}
	\begin{bmatrix}
		-0.408 & 0 & -0.408 & 0.816 \\
		-0.707 & 0 & 0.707  & 0     \\
		0.577  & 0 & 0.577  & 0.577 \\
		0      & 1 & 0      & 0
	\end{bmatrix}.
\end{equation}
To divide the graph into two connected components, one must use eigenvectors, corresponding to zero eigenvalues.
From these vectors, the $M$ matrix is formed:
\begin{equation}\label{eq19}
	M {=}
	\begin{bmatrix}
		0.577 & 0 \\
		0     & 1 \\
		0.577 & 0 \\
		0.577 & 0
	\end{bmatrix}.
\end{equation}

\subsubsection{Normalized Laplacian of the relation graph}
The normalized Laplacian calculated for the relation matrix (\ref{eq14}) has the form:
\begin{equation}\label{eq20}
	L_n {=}
	\begin{bmatrix}
		0.67  & 0 & -0.33 & -0.33 \\
		0     & 0 & 0     & 0     \\
		-0.33 & 0 & 0.67  & -0.33 \\
		-0.33 & 0 & -0.33 & 0.67
	\end{bmatrix}.
\end{equation}
Its eigenvalues are equal:
\begin{equation}\label{eq21}
	\lambda {=} \{1,1,0,0\}.
\end{equation}
Two zero eigenvalues were also obtained here.
The graph of relation has two connected components.
The eigenvectors corresponding to eigenvalues (\ref{eq21}) are the rows of the following matrix:
\begin{equation}\label{eq22}
	V^T {=}
	\begin{bmatrix}
		-0.408 & 0 & -0.408 & 0.816 \\
		-0.707 & 0 & 0.707  & 0     \\
		0      & 1 & 0      & 0     \\
		0.577  & 0 & 0.577  & 0.577
	\end{bmatrix}.
\end{equation}
It can be seen that the obtained matrices (\ref{eq18}) and (\ref{eq22}) are identical, with the accuracy of permutation of eigenvectors corresponding to zero eigenvalues.
The last two eigenvectors, corresponding to zero eigenvalues, should be used for clustering using k-means algorithm.
$M$ matrix is formed from these vectors.
After normalizing its rows to the unit length, matrix $M$ can be used for clustering with the k-means algorithm:

\begin{equation}\label{eq23}
	M {=}
	\begin{bmatrix}
		0 & 1 \\
		1 & 0 \\
		0 & 1 \\
		0 & 1
	\end{bmatrix}.
\end{equation}

\subsection{Mutual similarity of primary variables described by coefficients of determination}
Coefficients of determination describe the similarity of variables.
The matrix containing the determination coefficients can be treated as a generalized graph adjacency matrix, whose arcs have been assigned weighting factors.
For a matrix (\ref{eq11}) containing the determination coefficients, both the Laplacian and the normalized Laplacian can be formed.
By using these matrices, the primary variables can also be clustered.

\subsubsection{Laplacian of determination coefficients}
\begin{figure}
	\centering
	\includegraphics[width=10cm]{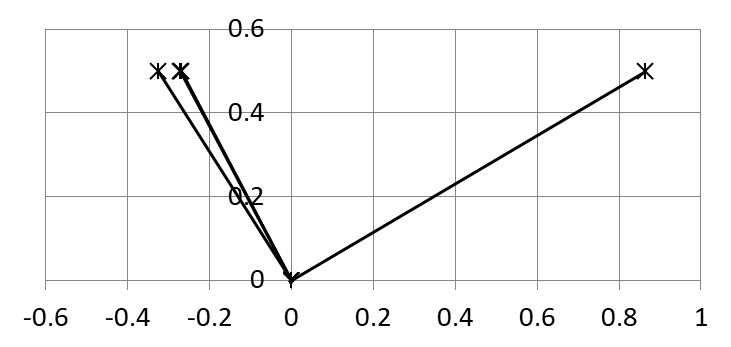}
	\caption{Clustering points (\ref{eq27}) obtained as a result of spectral analysis of Laplacian (\ref{eq24})}\label{figure04}
\end{figure}

Laplacian of determination coefficients has the form:
\begin{equation}\label{eq24}
	L {=}
	\begin{bmatrix}
		1.419  & -0.004 & -0.749 & -0.666 \\
		-0.004 & 0.197  & -0.103 & -0.090 \\
		-0.749 & -0.103 & 1.772  & -0.920 \\
		-0.666 & -0.090 & -0.920 & 1.676
	\end{bmatrix}.
\end{equation}
Its eigenvalues after sorting are equal:
\begin{equation}\label{eq25}
	\lambda {=} \{2.656,2.151,0.259,0\}.
\end{equation}
Eigenvalues correspond to successive eigenvectors, which are the rows of the matrix $V^T$:
\begin{equation}\label{eq26}
	V^T {=}
	\begin{bmatrix}
		-0.121 & -0.009 & 0.764  & -0.634 \\
		0.793  & 0.039  & -0.308 & -0.524 \\
		-0.326 & 0.865  & -0.267 & -0.272 \\
		0.5    & 0.5    & 0.5    & 0.5
	\end{bmatrix}.
\end{equation}
Assuming that the variables will be divided into two clusters, the $M$ matrix for clustering with the k-means algorithm can be formed from the last two rows of the matrix (\ref{eq26}):

\begin{equation}\label{eq27}
	M {=}
	\begin{bmatrix}
		-0.326 & 0.5 \\
		0.865  & 0.5 \\
		-0.267 & 0.5 \\
		-0.272 & 0.5
	\end{bmatrix}.
\end{equation}
The rows of this matrix can be represented as points on the plane (Figure \ref{figure04}).
The figure shows that the points are divided into clusters identically, as in the cases discussed above.

\subsubsection{Normalized Laplacian of determination coefficients}
\begin{figure}
	\centering
	\includegraphics[width=10cm]{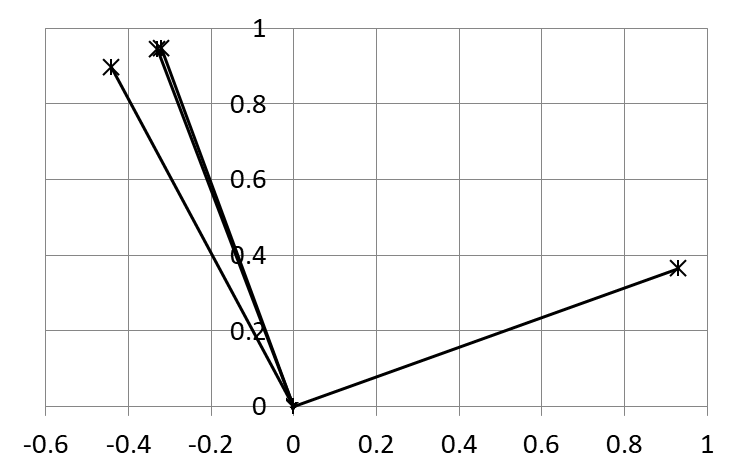}
	\caption{Clustering points (\ref{eq31}) obtained as a result of spectral analysis of normalized Laplacian (\ref{eq28}}\label{figure05}
\end{figure}

For the matrix of coefficients of determination, a normalized Laplacian $L_n$ was also formed:
\begin{equation}\label{eq28}
	L_n {=}
	\begin{bmatrix}
		0.587  & -0.002 & -0.289 & -0.262 \\
		-0.002 & 0.165  & -0.057 & -0.050 \\
		-0.289 & -0.057 & 0.639  & -0.338 \\
		-0.262 & -0.050 & -0.338 & 0.626
	\end{bmatrix}.
\end{equation}
After sorting its eigenvalues, they form a set:
\begin{equation}\label{eq29}
	\lambda {=} \{0.974,0.856,0.187,0\}.
\end{equation}
The corresponding eigenvectors are rows in the matrix $V^T$:
\begin{equation}\label{eq30}
	V^T {=}
	\begin{bmatrix}
		-0.143 & -0.014 & 0.762  & -0.631 \\
		0.805  & 0.058  & -0.279 & -0.520 \\
		-0.254 & 0.930  & -0.187 & -0.189 \\
		0.517  & 0.363  & 0.553  & 0.543
	\end{bmatrix}.
\end{equation}
For clustering into two clusters, the last two rows of the matrix (\ref{eq30}), after normalization to the unitary length form a matrix $M$ whose rows will be used for clustering with the k-means algorithm:
\begin{equation}\label{eq31}
	M {=}
	\begin{bmatrix}
		-0.442 & 0.897 \\
		0.931  & 0.364 \\
		-0.321 & 0.947 \\
		-0.329 & 0.944
	\end{bmatrix}.
\end{equation}
The rows of this matrix describe the points on the plane, located on a circle with a unitary radius (Figure \ref{figure05}).

\begin{figure}
	\centering
	\includegraphics[width=0.75\textwidth]{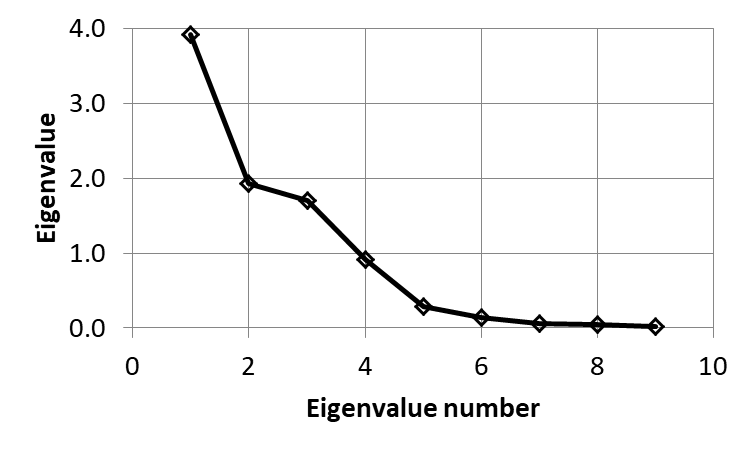}
	\caption{The scree plot for Houses Data}\label{figure06}
\end{figure}

\subsection{Conditions for clustering of primary variables}
Analyzing Figures \ref{figure01}, \ref{figure02}, \ref{figure04} and \ref{figure05}, it can be noticed that due to the mutual distance of points, the Euklidean metric (\ref{eq05}) can be used for k-means clustering.
On the other hand, due to the angular relations between the vectors representing the points, the cosine measure of dissimilarity (\ref{eq07}) can be used.
In this work, both methods of measuring the dissimilarity were used.

In the case of clustering due to the similarity of the primary variables to the principal components, for a cosine measure there is no need to distinguish between the matrix (\ref{eq12}) and the normalized matrix (\ref{eq13}).
In both cases the cosine will be the same (See the penultimate paragraph in subsection \ref{Measures}).
This fact was taken into account in the calculations made for the purpose of this work.

In different variants of the spectral method, the corresponding matrix $M$ is created from eigenvectors corresponding to the smallest eigenvalues.
An analogous matrix $M$ is also created in a method that uses similarity of primary variables to principal components.
In all of these cases, any row in matrix $M$ representing one primary variable has the interpretation of the point in the space in which the k-means algorithm operates.
In the case of clustering four variables describing the iris flowers, to find two clusters, two different initial points should be chosen in this space.

With spectral clustering based on the relation matrix, it can be seen that in the matrix (\ref{eq19}) and (\ref{eq23}) three lines are identical.
This means that in both variants there is only one pair of different initial points.
For matrix (\ref{eq19}), this pair are points with coordinates $(0.577,0)$ and $(0,1)$, while for matrix (\ref{eq23}) they are points $(1,0)$ and $(0,1)$.
This causes both pairs of points to be the final centers of two clusters and means that clustering using a relation matrix becomes a trivial task.

In turn, when clustering due to similarity to the principal components, as well as in the case of spectral clustering with the matrix of determination coefficients as the similarity matrix, for four different points subject to clustering, the number of all pairs of different initial points is equal to $\binom{4}{2}=6$.
Also interesting is the answer to the question of how the algorithm will behave for different pairs of initial points.
For the purpose of this work an algorithm for each of six pairs was started.

\subsection{The results of clustering variables from the Iris Data}
Clustering of variables describing iris flowers has been taken in consideration of all the cases described above.
First, matrices of relations obtained from the matrix of determination coefficients for $\epsilon$ varying from $40\%$ to $70\%$ with a step of $5\%$ were analyzed by spectral methods.
Four different ways of clustering were considered:
\begin{itemize}
	\item $2EL\epsilon$ - Laplacian of relation, Euclidean metric;
	\item $2CL\epsilon$ - Laplacian of relation, cosine measure of dissimilarity;
	\item $2EnL\epsilon$ - normalized Laplacian of relation, Euclidean metric;
	\item $2CnL\epsilon$ - normalized Laplacian of relation, a cosine measure of dissimilarity.
\end{itemize}
The results were compared with the clusters provided in the file containing the graph nodes marked with the $2Man$ symbol, shown in Figures \ref{figure03a} or \ref{figure03b}.
A $100\%$ compatibility was found.
For all values of $\epsilon$, for all clustering methods, the nodes were divided between clusters identical to those in the $2Man$ variant.

Clusters of 2Man variant was also adopted as a reference for the analysis of other ways of clustering:
\begin{enumerate}
	\item For the six different pairs of initial points, primary variables have been clustered due to their similarity to the principal components:
	      \begin{itemize}
		      \item $2EP$ - Euclidean metric;
		      \item $2CP$ - a cosine measure of dissimilarity;
		      \item $2EnP$ - Euclidean metric, normalized points for clustering
	      \end{itemize}
	\item For the matrix of determination coefficients in the role of the similarity matrix, using spectral methods, primary variables were clustered using six different pairs of initial points:
	      \begin{itemize}
		      \item $2EL$ - Laplacian of similarity matrix, Euclidean metric;
		      \item $2CL$ - Laplacian of similarity matrix, the cosine measure of dissimilarity;
		      \item $2EnL$ - normalized Laplacian of similarity matrix, Euclidean metric;
		      \item $2CnL$ - normalized Laplacian of similarity matrix, the cosine measure of dissimilarity.
	      \end{itemize}
\end{enumerate}
In all cases of clustering, both based on the similarity of the primary variables to the principal components, and also with the use of spectral methods, which are based on the matrix of determination coefficients, identical clustering results were obtained, consistent with the 2Man pattern.
Each time, variables 1, 3 and 4 formed one cluster, and the variable number 2 was in the second cluster.
The similarity of primary variables to the principal components was identical with the mutual similarity of variables described both by the relation as well as the coefficients of determination.
\section{The dataset No. 2 - Houses Data}
As set number 2, a dataset known as "Houses Data" was used, which was also used in \cite{Larose2006} and \cite{Pace1997}. In \cite{Larose2006} there is also a link to its location \cite{Pace1999}.
The set describes nine variables. Each variable was measured 41280 times.
The first variable is the response variable.
The remaining eight variables are predictor variables.
In \cite{Larose2006} the analysis of the principal components is carried out only for eight predictor variables.
Since all the variables are correlated, in this study the analysis is carried out for all nine variables.

The correlation coefficients were calculated for Houses Data:
\begin{equation}\label{eq32}
	C_R {=}
	\begin{bmatrix}
		1     & 0.69  & 0.11  & 0.13  & 0.05  & -0.02 & 0.07  & -0.14 & -0.05 \\
		0.69  & 1     & -0.12 & 0.20  & -0.01 & 0.00  & 0.01  & -0.08 & -0.02 \\
		0.11  & -0.12 & 1     & -0.36 & -0.32 & -0.30 & -0.30 & 0.01  & -0.11 \\
		0.13  & 0.20  & -0.36 & 1     & 0.93  & 0.86  & 0.92  & -0.04 & 0.04  \\
		0.05  & -0.01 & -0.32 & 0.93  & 1     & 0.88  & 0.98  & -0.07 & 0.07  \\
		-0.02 & 0.00  & -0.30 & 0.86  & 0.88  & 1     & 0.91  & -0.11 & 0.10  \\
		0.07  & 0.01  & -0.30 & 0.92  & 0.98  & 0.91  & 1     & -0.07 & 0.06  \\
		-0.14 & -0.08 & 0.01  & -0.04 & -0.07 & -0.11 & -0.07 & 1     & -0.92 \\
		-0.05 & -0.02 & -0.11 & 0.04  & 0.07  & 0.10  & 0.06  & -0.92 & 1
	\end{bmatrix}.
\end{equation}
For this correlation matrix, the eigenproblem was solved, and then principal components were found.
Table \ref{table03} presents an analysis of the cumulative variance of the principal components. A scree plot was also found (Figure \ref{figure06}).
From Table \ref{table03}, it can be read that the three principal components would explain more than $83\%$ of the average variance of the primary variables.
On the other hand, on the scree plot in Figure \ref{figure06}, there are $4$ variables on the scree.

\begin{table}
	\centering
	\caption{The percentage of the variance explained by the succesive principal components estimated for Houses Data}\label{table03}
	\fontsize{9.5}{13.5}\selectfont{
		\begin{tabular}{c|c|c|c|c} \hline
			\multirow{2}{*}{No.} & \multirow{2}{*}{Eigenvalue} & Cumulative  & Percentage of variance & Cumulative             \\
			                     &                             & eigenvalues & explained by each PC   & percentage of variance \\ \hline \hline
			$1$                  & $3.912$                     & $3.912$     & $43.5\%$               & $43.5\%$               \\ \hline
			$2$                  & $1.923$                     & $5.835$     & $21.4\%$               & $64.8\%$               \\ \hline
			$3$                  & $1.697$                     & $7.532$     & $18.9\%$               & $83.7\%$               \\ \hline
			$4$                  & $0.910$                     & $8.442$     & $10.1\%$               & $93.8\%$               \\ \hline
			$5$                  & $0.293$                     & $8.736$     & $3.3\%$                & $97.1\%$               \\ \hline
			$6$                  & $0.143$                     & $8.878$     & $1.6\%$                & $98.6\%$               \\ \hline
			$7$                  & $0.063$                     & $8.941$     & $0.7\%$                & $99.3\%$               \\ \hline
			$8$                  & $0.045$                     & $8.985$     & $0.5\%$                & $99.8\%$               \\ \hline
			$9$                  & $0.015$                     & $9$         & $0.2\%$                & $100\%$                \\ \hline
		\end{tabular}}
\end{table}

\subsection{Similarity of primary variables to the principal components}
Having principal components, the coefficients of determination between primary variables and principal components were estimated (Table \ref{table04}).
\begin{table}
	\centering
	\caption{Determination coefficients between primary variables and principal components for Houses Data}\label{table04}
	\fontsize{9.5}{13.5}\selectfont{
		\begin{tabular}{c||c|c|c|c|c|c|c|c|c} \hline
			      & $V1$     & $V2$     & $V3$     & $V4$     & $V5$     & $V6$     & $V7$     & $V8$     & $V9 $     \\ \hline \hline
			$PC1$ & $0.8\%$  & $1.2\%$  & $18.3\%$ & $91.8\%$ & $93.8\%$ & $86.5\%$ & $94.3\%$ & $2.2\%$  & $2.2\% $  \\ \hline
			$PC2$ & $7.0\%$  & $6.1\%$  & $0.0\%$  & $0.7\%$  & $1.0\%$  & $0.4\%$  & $1.0\%$  & $91.7\%$ & $84.4\% $ \\ \hline
			$PC3$ & $77.0\%$ & $76.5\%$ & $0.4\%$  & $1.2\%$  & $0.3\%$  & $1.1\%$  & $0.1\%$  & $2.7\%$  & $10.3\% $ \\ \hline
			$PC4$ & $2.5\%$  & $4.0\%$  & $79.0\%$ & $0.1\%$  & $1.4\%$  & $1.3\%$  & $1.9\%$  & $0.7\%$  & $0.1\% $  \\ \hline
			$PC5$ & $12.1\%$ & $11.8\%$ & $2.2\%$  & $0.8\%$  & $0.6\%$  & $1.2\%$  & $0.3\%$  & $0.2\%$  & $0.3\% $  \\ \hline
			$PC6$ & $0.2\%$  & $0.0\%$  & $0.1\%$  & $2.3\%$  & $1.5\%$  & $9.2\%$  & $0.1\%$  & $0.2\%$  & $0.5\% $  \\ \hline
			$PC7$ & $0.3\%$  & $0.4\%$  & $0.0\%$  & $2.3\%$  & $0.6\%$  & $0.3\%$  & $1.4\%$  & $0.5\%$  & $0.5\% $  \\ \hline
			$PC8$ & $0.1\%$  & $0.0\%$  & $0.0\%$  & $0.8\%$  & $0.1\%$  & $0.0\%$  & $0.1\%$  & $1.7\%$  & $1.6\% $  \\ \hline
			$PC9$ & $0.0\%$  & $0.0\%$  & $0.0\%$  & $0.0\%$  & $0.7\%$  & $0.0\%$  & $0.7\%$  & $0.0\%$  & $0.0\% $  \\ \hline
		\end{tabular}}
\end{table}
\begin{table}
	\centering
	\caption{The level of explanation of the variance of primary variables by the four principal components}\label{table05}
	\fontsize{9.5}{13.5}\selectfont{
		\begin{tabular}{c||c|c|c|c|c|c|c|c|c} \hline
			         & $V1$     & $V2$     & $V3$     & $V4$     & $V5$     & $V6$     & $V7$     & $V8$     & $V9 $    \\ \hline \hline
			$PC1$    & $0.8\%$  & $1.2\%$  & $18.3\%$ & $91.8\%$ & $93.8\%$ & $86.5\%$ & $94.3\%$ & $2.2\%$  & $2.2\%$  \\ \hline
			$PC2$    & $7.0\%$  & $6.1\%$  & $0.0\%$  & $0.7\%$  & $1.0\%$  & $0.4\%$  & $1.0\%$  & $91.7\%$ & $84.4\%$ \\ \hline
			$PC3$    & $77.0\%$ & $76.5\%$ & $0.4\%$  & $1.2\%$  & $0.3\%$  & $1.1\%$  & $0.1\%$  & $2.7\%$  & $10.3\%$ \\ \hline
			$PC4$    & $2.5\%$  & $4.0\%$  & $79.0\%$ & $0.1\%$  & $1.4\%$  & $1.3\%$  & $1.9\%$  & $0.7\%$  & $0.1\%$  \\ \hline\hline
			$\Sigma$ & $87.4\%$ & $87.8\%$ & $97.7\%$ & $93.8\%$ & $96.5\%$ & $89.3\%$ & $97.3\%$ & $97.4\%$ & $97.1\%$ \\ \hline
		\end{tabular}}
\end{table}

Due to the necessity to explain the variance of $V3$ variable, you can not use only three principal components.
With the three principal components, the level of variance representation of the variable $V3$ would be lower than $19\%$.
Only the four principal components will allow to represent the variable $V3$ in a satisfactory way.
Table \ref{table05} shows the level of representation of the variance of all primary variables using four principal components:
\begin{itemize}
	\item For the first and second variable, the four principal components will explain more than $87\%$ of their variances. From this, the third principal component explains more than $77\%$ of the variance of the first variable and over $76\%$ of the variance of the second variable.
	\item The third variable is explained in more than $97\%$. Of this, almost $79\%$ falls on the fourth principal component.
	\item The level of explanation of primary variables from the fourth to the seventh by the four principal components ranges from over $89\%$ (variable six), to over $97\%$ (variable seventh). In this, the first principal component explains more than $86\%$ of the variance of the sixth variable, and more than $94\%$ of the variance of the seventh variable.
	\item The eighth and ninth primary variables are explained by the four principal components in more than $97\%$. From this, the second principal component explains more than $91\%$ of the eighth variable and more than $84\%$ of the ninth variable.
\end{itemize}
The assumption of the necessity of the four principal components suggests the division of the set of primary variables into four clusters.
In each cluster there will be primary variables similar to those principal components.
Clustering can be performed using the first four rows of the table containing the coefficients of determination between the primary variables and the principal components (Table \ref{table04}). The transposition of matrix $M$ used for clustering takes the following form:

\begin{equation}\label{eq33}
	M^T {=}
	\begin{bmatrix}
		0.008 & 0.012 & 0.183 & 0.918 & 0.938 & 0.865 & 0.943 & 0.022 & 0.022 \\
		0.070 & 0.061 & 0.000 & 0.007 & 0.010 & 0.004 & 0.010 & 0.917 & 0.844 \\
		0.770 & 0.765 & 0.004 & 0.012 & 0.003 & 0.011 & 0.001 & 0.027 & 0.103 \\
		0.025 & 0.040 & 0.790 & 0.001 & 0.014 & 0.013 & 0.019 & 0.007 & 0.001
	\end{bmatrix}.
\end{equation}

\begin{figure}
	\centering
	{\subfloat[] {\label{figure07a}
			\includegraphics[width=0.35\textwidth]{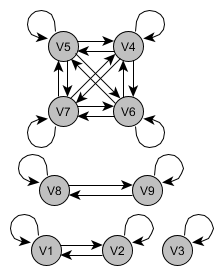}
		}
		\quad
		\quad
		\quad
		\subfloat[] {\label{figure07b}
			\includegraphics[width=0.35\textwidth]{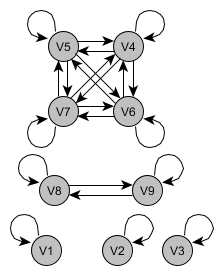}
		}
		\caption{Fragment of the evolution of the relation: (a) $\epsilon=45\%$, equivalence relation (\ref{eq35}), four equivalence classes; (b) $\epsilon=50\%$, equivalence relation, five equivalence classes}
		\label{figure07}}
\end{figure}

\subsection{The relation established on the level of similarity}
The matrix of determination coefficients describes the mutual similarity of primary variables:
\begin{equation}\label{eq34}
	S {=}
	\begin{bmatrix}
		1     & 0.473 & 0.011 & 0.018 & 0.003 & 0.001 & 0.004 & 0.021 & 0.002 \\
		0.473 & 1     & 0.014 & 0.039 & 0.000 & 0.000 & 0.000 & 0.006 & 0.000 \\
		0.011 & 0.014 & 1     & 0.131 & 0.103 & 0.088 & 0.092 & 0.000 & 0.012 \\
		0.018 & 0.039 & 0.131 & 1     & 0.865 & 0.735 & 0.844 & 0.001 & 0.002 \\
		0.003 & 0.000 & 0.103 & 0.865 & 1     & 0.771 & 0.960 & 0.004 & 0.005 \\
		0.001 & 0.000 & 0.088 & 0.735 & 0.771 & 1     & 0.823 & 0.012 & 0.010 \\
		0.004 & 0.000 & 0.092 & 0.844 & 0.960 & 0.823 & 1     & 0.005 & 0.003 \\
		0.021 & 0.006 & 0.000 & 0.001 & 0.004 & 0.012 & 0.005 & 1     & 0.855 \\
		0.002 & 0.000 & 0.012 & 0.002 & 0.005 & 0.010 & 0.003 & 0.855 & 1
	\end{bmatrix}.
\end{equation}
Based on the matrix (\ref{eq34}), different relations were tested, depending on the $\epsilon$ value.
For $\epsilon = 45\%$, a relation graph with four connected components was obtained.
For $\epsilon = 50\%$, the graph has five connected components.
Figure \ref{figure07} shows a fragment of the relation evolution for $\epsilon = 45\%$ and $\epsilon = 50\%$.

In order to analyze the partition into four clusters, the threshold value of $\epsilon=45\%$ is assumed for the tests, at which the relation graph also has 4 connected components.
The relation matrix for $\epsilon=45\%$ has the form:
\begin{equation}\label{eq35}
	R_{\epsilon=45\%} {=}
	\begin{bmatrix}
		$1$ & $1$ & $0$ & $0$ & $0$ & $0$ & $0$ & $0$ & $0$ \\
		$1$ & $1$ & $0$ & $0$ & $0$ & $0$ & $0$ & $0$ & $0$ \\
		$0$ & $0$ & $1$ & $0$ & $0$ & $0$ & $0$ & $0$ & $0$ \\
		$0$ & $0$ & $0$ & $1$ & $1$ & $1$ & $1$ & $0$ & $0$ \\
		$0$ & $0$ & $0$ & $1$ & $1$ & $1$ & $1$ & $0$ & $0$ \\
		$0$ & $0$ & $0$ & $1$ & $1$ & $1$ & $1$ & $0$ & $0$ \\
		$0$ & $0$ & $0$ & $1$ & $1$ & $1$ & $1$ & $0$ & $0$ \\
		$0$ & $0$ & $0$ & $0$ & $0$ & $0$ & $0$ & $1$ & $1$ \\
		$0$ & $0$ & $0$ & $0$ & $0$ & $0$ & $0$ & $1$ & $1$
	\end{bmatrix}.
\end{equation}

\subsubsection{Laplacian matrix of the relation graph}
For the relation (\ref{eq35}), the Laplacian $L$ was established:
\begin{equation}\label{eq36}
	L {=}
	\begin{bmatrix}
		$1$  & $-1$ & $0$ & $0$  & $0$  & $0$  & $0$  & $0$  & $0$  \\
		$-1$ & $1$  & $0$ & $0$  & $0$  & $0$  & $0$  & $0$  & $0$  \\
		$0$  & $0$  & $0$ & $0$  & $0$  & $0$  & $0$  & $0$  & $0$  \\
		$0$  & $0$  & $0$ & $3$  & $-1$ & $-1$ & $-1$ & $0$  & $0$  \\
		$0$  & $0$  & $0$ & $-1$ & $3$  & $-1$ & $-1$ & $0$  & $0$  \\
		$0$  & $0$  & $0$ & $-1$ & $-1$ & $3$  & $-1$ & $0$  & $0$  \\
		$0$  & $0$  & $0$ & $-1$ & $-1$ & $-1$ & $3$  & $0$  & $0$  \\
		$0$  & $0$  & $0$ & $0$  & $0$  & $0$  & $0$  & $1$  & $-1$ \\
		$0$  & $0$  & $0$ & $0$  & $0$  & $0$  & $0$  & $-1$ & $1$
	\end{bmatrix}.
\end{equation}
For Laplacian (\ref{eq36}), eigenvalues were calculated:
\begin{equation}\label{eq37}
	\lambda {=} \{4,4,4,2,2,0,0,0,0\}.
\end{equation}
Four zero eigenvalues show that the relation graph has four connected components.
This is also confirmed in Fig. \ref{figure07a}.
The corresponding eigenvectors are the rows of the following matrix $V^T$:
\begin{equation}\label{eq38}
	V^T {=}
	\begin{bmatrix}
		0      & 0     & 0 & -0.408 & -0.408 & 0.816  & 0     & 0      & 0     \\
		0      & 0     & 0 & -0.707 & 0.707  & 0      & 0     & 0      & 0     \\
		0      & 0     & 0 & -0.289 & -0.289 & -0.289 & 0.866 & 0      & 0     \\
		-0.707 & 0.707 & 0 & 0      & 0      & 0      & 0     & 0      & 0     \\
		0      & 0     & 0 & 0      & 0      & 0      & 0     & -0.707 & 0.707 \\
		0      & 0     & 0 & 0.5    & 0.5    & 0.5    & 0.5   & 0      & 0     \\
		0      & 0     & 1 & 0      & 0      & 0      & 0     & 0      & 0     \\
		0.707  & 0.707 & 0 & 0      & 0      & 0      & 0     & 0      & 0     \\
		0      & 0     & 0 & 0      & 0      & 0      & 0     & 0.707  & 0.707
	\end{bmatrix}.
\end{equation}
The last four eigenvectors are used for clustering into four clusters.
These vectors form columns in the matrix $M$. Its transposition has the form:
\begin{equation}\label{eq39}
	M^T {=}
	\begin{bmatrix}
		0     & 0     & 0 & 0.5 & 0.5 & 0.5 & 0.5 & 0     & 0     \\
		0     & 0     & 1 & 0   & 0   & 0   & 0   & 0     & 0     \\
		0.707 & 0.707 & 0 & 0   & 0   & 0   & 0   & 0     & 0     \\
		0     & 0     & 0 & 0   & 0   & 0   & 0   & 0.707 & 0.707
	\end{bmatrix}.
\end{equation}

\subsubsection{Normalized Laplacian of the relation graph}
For the relation (\ref{eq35}), the normalized Laplacian $L_n$ was also estimated:
\begin{equation}\label{eq40}
	L_n {=}
	\begin{bmatrix}
		$0.5$  & $-0.5$ & $0$ & $0$     & $0$     & $0$     & $0$     & $0$    & $0$    \\
		$-0.5$ & $0.5$  & $0$ & $0$     & $0$     & $0$     & $0$     & $0$    & $0$    \\
		$0$    & $0$    & $0$ & $0$     & $0$     & $0$     & $0$     & $0$    & $0$    \\
		$0$    & $0$    & $0$ & $0.70$  & $-0.25$ & $-0.25$ & $-0.25$ & $0$    & $0$    \\
		$0$    & $0$    & $0$ & $-0.25$ & $0.75$  & $-0.25$ & $-0.25$ & $0$    & $0$    \\
		$0$    & $0$    & $0$ & $-0.25$ & $-0.25$ & $0.75$  & $-0.25$ & $0$    & $0$    \\
		$0$    & $0$    & $0$ & $-0.25$ & $-0.25$ & $-0.25$ & $0.75$  & $0$    & $0$    \\
		$0$    & $0$    & $0$ & $0$     & $0$     & $0$     & $0$     & $0.5$  & $-0.5$ \\
		$0$    & $0$    & $0$ & $0$     & $0$     & $0$     & $0$     & $-0.5$ & $0.5$
	\end{bmatrix}.
\end{equation}
For the matrix (\ref{eq40}), the eigenvalues were also estimated:
\begin{equation}\label{eq41}
	\lambda {=} \{1,1,1,1,1,0,0,0,0\}.
\end{equation}
The four zero eigenvalues also confirm the fact that the relation graph has four connected components (Figure \ref{figure07a}).
The corresponding eigenvectors form the rows of the following matrix $V^T$:

\begin{equation}\label{eq42}
	V^T {=}
	\begin{bmatrix}
		0      & 0     & 0 & -0.408 & -0.408 & 0.816  & 0     & 0      & 0     \\
		0      & 0     & 0 & -0.707 & 0.707  & 0      & 0     & 0      & 0     \\
		-0.707 & 0.707 & 0 & 0      & 0      & 0      & 0     & 0      & 0     \\
		0      & 0     & 0 & -0.289 & -0.289 & -0.289 & 0.866 & 0      & 0     \\
		0      & 0     & 0 & 0      & 0      & 0      & 0     & -0.707 & 0.707 \\
		0      & 0     & 0 & 0.5    & 0.5    & 0.5    & 0.5   & 0      & 0     \\
		0      & 0     & 1 & 0      & 0      & 0      & 0     & 0      & 0     \\
		0.707  & 0.707 & 0 & 0      & 0      & 0      & 0     & 0      & 0     \\
		0      & 0     & 0 & 0      & 0      & 0      & 0     & 0.707  & 0.707
	\end{bmatrix}.
\end{equation}
It should be noted that the eigenvectors (\ref{eq42}) of normalized Laplacian $L_n$ (\ref{eq40}) are identical to eivenvectors (\ref{eq38}) of Laplacian $L$ (\ref{eq36}), with accuracy to the permutation of eigenvectors.

As above, the last four eigenvectors should be used for clustering into four clusters.
After their transposition, a rectangular matrix is formed, the rows of which, after normalizing to the unitary length, form the matrix $M$, used by the k-means algorithm. The transposition of matrix $M$ takes the form:
\begin{equation}\label{eq43}
	M^T {=}
	\begin{bmatrix}
		$0$ & $0$ & $0$ & $1$ & $1$ & $1$ & $1$ & $0$ & $0$ \\
		$0$ & $0$ & $1$ & $0$ & $0$ & $0$ & $0$ & $0$ & $0$ \\
		$1$ & $1$ & $0$ & $0$ & $0$ & $0$ & $0$ & $0$ & $0$ \\
		$0$ & $0$ & $0$ & $0$ & $0$ & $0$ & $0$ & $1$ & $1$
	\end{bmatrix}.
\end{equation}
\subsection{Spectral clustering based on the matrix of coefficients of determination}
Matrix (\ref{eq34}) containing the determination coefficients describes the mutual similarity of the primary variables.
This matrix can also be treated as a generalized graph adjacency matrix, for which both Laplacian $L$ and normalized Laplacian $L_n$ were formed.
Both were used for clustering of primary variables.
\subsubsection{Laplacian of determination coefficients matrix}
Laplacian $L$ formed for the matrix of coefficients of determination (\ref{eq34}) has the form:

\begin{equation}\label{eq44}
	L {=}
	\begin{bmatrix}
		0.53  & -0.47 & -0.01 & -0.02 & 0.00  & 0.00  & 0.00  & -0.02 & 0.00  \\
		-0.47 & 0.53  & -0.01 & -0.04 & 0.00  & 0.00  & 0.00  & -0.01 & 0.00  \\
		-0.01 & -0.01 & 0.45  & -0.13 & -0.10 & -0.09 & -0.09 & 0.00  & -0.01 \\
		-0.02 & -0.04 & -0.13 & 2.63  & -0.86 & -0.73 & -0.84 & 0.00  & 0.00  \\
		0.00  & 0.00  & -0.10 & -0.86 & 2.71  & -0.77 & -0.96 & 0.00  & 0.00  \\
		0.00  & 0.00  & -0.09 & -0.73 & -0.77 & 2.44  & -0.82 & -0.01 & -0.01 \\
		0.00  & 0.00  & -0.09 & -0.84 & -0.96 & -0.82 & 2.73  & -0.01 & 0.00  \\
		-0.02 & -0.01 & 0.00  & 0.00  & 0.00  & -0.01 & -0.01 & 0.90  & -0.86 \\
		0.00  & 0.00  & -0.01 & 0.00  & 0.00  & -0.01 & 0.00  & -0.86 & 0.89
	\end{bmatrix}.
\end{equation}
Its eigenvalues after being sorted are equal:
\begin{equation}\label{eq45}
	\lambda {=} \{3.684,3.488,3.213,1.752,1.007,0.550,0.077,0.053,0.000\}.
\end{equation}
Eigenvalues correspond to successive eigenvectors, which are the rows of the matrix $V^T$:
\begin{equation}\label{eq46}
	V^T {=}
	\begin{bmatrix}
		$0.00$  & $0.00$  & $0.00$  & $0.04$  & $-0.69$ & $-0.07$ & $0.72$  & $0.00$  & $0.00$  \\
		$0.00$  & $-0.01$ & $-0.01$ & $0.77$  & $-0.44$ & $0.13$  & $-0.44$ & $0.00$  & $0.00$  \\
		$0.00$  & $0.01$  & $0.01$  & $-0.41$ & $-0.29$ & $0.85$  & $-0.17$ & $0.00$  & $0.00$  \\
		$-0.01$ & $0.00$  & $0.01$  & $0.00$  & $0.00$  & $0.00$  & $0.00$  & $0.71$  & $-0.70$ \\
		$-0.71$ & $0.71$  & $0.00$  & $0.00$  & $0.00$  & $0.00$  & $0.00$  & $0.00$  & $0.01$  \\
		$-0.01$ & $-0.01$ & $0.90$  & $-0.21$ & $-0.22$ & $-0.23$ & $-0.22$ & $-0.01$ & $0.00$  \\
		$0.62$  & $0.62$  & $-0.19$ & $-0.20$ & $-0.21$ & $-0.22$ & $-0.21$ & $-0.09$ & $-0.10$ \\
		$-0.08$ & $-0.09$ & $-0.21$ & $-0.21$ & $-0.22$ & $-0.21$ & $-0.22$ & $0.61$  & $0.62$  \\
		$0.33$  & $0.33$  & $0.33$  & $0.33$  & $0.33$  & $0.33$  & $0.33$  & $0.33$  & $0.33$
	\end{bmatrix}.
\end{equation}
Assuming that the variables are clustered into four clusters, from the last four rows of this matrix, an $M$ matrix is created for clustering with the use of k-means algorithm. The transposition of matrix $M$ takes the form::
\begin{equation}\label{eq47}
	M^T {=}
	\begin{bmatrix}
		$-0.01$ & $-0.01$ & $0.90$  & $-0.21$ & $-0.22$ & $-0.23$ & $-0.22$ & $-0.01$ & $0.00$  \\
		$0.62$  & $0.62$  & $-0.19$ & $-0.20$ & $-0.21$ & $-0.22$ & $-0.21$ & $-0.09$ & $-0.10$ \\
		$-0.08$ & $-0.09$ & $-0.21$ & $-0.21$ & $-0.22$ & $-0.21$ & $-0.22$ & $0.61$  & $0.62$  \\
		$0.33$  & $0.33$  & $0.33$  & $0.33$  & $0.33$  & $0.33$  & $0.33$  & $0.33$  & $0.33$
	\end{bmatrix}.
\end{equation}
\subsubsection{Normalized Laplacian of determination coefficients matrix}
For the matrix of determination coefficients (\ref{eq34}), a standardized Laplacian $L_n$ was also formed:
\begin{equation}\label{eq48}
	L_n {=}
	\begin{bmatrix}
		0.35  & -0.31 & -0.01 & -0.01 & 0.00  & 0.00  & 0.00  & -0.01 & 0.00  \\
		-0.31 & 0.35  & -0.01 & -0.02 & 0.00  & 0.00  & 0.00  & 0.00  & 0.00  \\
		-0.01 & -0.01 & 0.31  & -0.06 & -0.04 & -0.04 & -0.04 & 0.00  & -0.01 \\
		-0.01 & -0.02 & -0.06 & 0.72  & -0.24 & -0.21 & -0.23 & 0.00  & 0.00  \\
		0.00  & 0.00  & -0.04 & -0.24 & 0.73  & -0.22 & -0.26 & 0.00  & 0.00  \\
		0.00  & 0.00  & -0.04 & -0.21 & -0.22 & 0.71  & -0.23 & 0.00  & 0.00  \\
		0.00  & 0.00  & -0.04 & -0.23 & -0.26 & -0.23 & 0.73  & 0.00  & 0.00  \\
		-0.01 & 0.00  & 0.00  & 0.00  & 0.00  & 0.00  & 0.00  & 0.48  & -0.45 \\
		0.00  & 0.00  & -0.01 & 0.00  & 0.00  & 0.00  & 0.00  & -0.45 & 0.47
	\end{bmatrix}.
\end{equation}
Its eigenvalues after sorting will form a set:
\begin{equation}\label{eq49}
	\lambda {=} \{0.99,0.95,0.92,0.92,0.66,0.34,0.04,0.02,0\}.
\end{equation}
The corresponding eigenvectors are the rows of the following matrix $V^T$:
\begin{equation}\label{eq50}
	V^T {=}
	\begin{bmatrix}
		0.00  & 0.00  & 0.00  & 0.09  & -0.68 & -0.12 & 0.71  & 0.00  & 0.00  \\
		0.00  & -0.02 & -0.02 & 0.75  & -0.42 & 0.18  & -0.47 & 0.00  & 0.00  \\
		-0.02 & 0.00  & 0.01  & 0.03  & 0.02  & -0.07 & 0.01  & 0.71  & -0.70 \\
		0.00  & 0.01  & 0.01  & -0.43 & -0.30 & 0.84  & -0.10 & 0.05  & -0.06 \\
		-0.71 & 0.71  & 0.00  & 0.02  & -0.01 & -0.01 & -0.01 & 0.00  & 0.02  \\
		-0.02 & -0.02 & 0.96  & -0.12 & -0.15 & -0.15 & -0.16 & -0.01 & 0.00  \\
		0.65  & 0.65  & -0.05 & -0.12 & -0.14 & -0.14 & -0.14 & -0.18 & -0.19 \\
		0.08  & 0.07  & -0.12 & -0.23 & -0.23 & -0.22 & -0.23 & 0.62  & 0.62  \\
		0.26  & 0.26  & 0.25  & 0.40  & 0.40  & 0.39  & 0.40  & 0.29  & 0.29
	\end{bmatrix}.
\end{equation}
For clustering nine variables into four clusters, the last four rows of the matrix (\ref{eq50}) form a matrix whose row vectors after normalization to a unit length form an $M$ matrix for k-means clustering. Its transposition takes the form:
\begin{equation}\label{eq51}
	M^T {=}
	\begin{bmatrix}
		-0.02 & -0.02 & 0.96  & -0.12 & -0.15 & -0.15 & -0.16 & -0.01 & 0.00  \\
		0.65  & 0.65  & -0.05 & -0.12 & -0.14 & -0.14 & -0.14 & -0.18 & -0.19 \\
		0.08  & 0.07  & -0.12 & -0.23 & -0.23 & -0.22 & -0.23 & 0.62  & 0.62  \\
		0.26  & 0.26  & 0.25  & 0.40  & 0.40  & 0.39  & 0.40  & 0.29  & 0.29
	\end{bmatrix}.
\end{equation}

\subsection{The results of clustering variables from the Houses Data}
As in the case of the Iris Dataset, all of the clustering options discussed earlier were examined.
First, clustering of primary variables into four clusters was investigated.
Then, the primary variables were clustered in the same way into five clusters.

\subsubsection{Clustering of primary variables into four clusters}
At the beginning, the relation matrix (\ref{eq35}) obtained for the assumed value of $\epsilon = 45\%$ was analyzed.
Four versions of the spectral clustering algorithm were considered:
\begin{itemize}
	\item $4EL45\%$ - Laplacian of relation, Euclidean metric;
	\item $4CL45\%$ - Laplacian of relation, cosine measure of dissimilarity;
	\item $4EnL45\%$ - normalized Laplacian of relation, Euclidean metric;
	\item $4CnL45\%$ - normalized Laplacian of relation, cosine measure of dissimilarity.
\end{itemize}
Because in matrices used for clustering with the k-means algorithm (see their transpositions (\ref{eq39}), and (\ref{eq43})) only four different rows exist, they will be the only initial points for clustering.
At the same time, these quartets are in themselves cluster centers.
Already on the occasion of clustering based on the relation for the Iris Data, it was noticed that clustering is $100\%$ effective.
The same thing is with the collection of Houses Data.
The results of the four versions of the algorithm presented above were compared with the distribution of the graph nodes shown in Figure 7a, labeled as 4Man.
Full compatibility was obtained for the four tested versions of the algorithm and for the 4Man set.

\begin{figure}
	\centering
	{\subfloat[] {\label{figure08a}
			\includegraphics[width=0.75\textwidth]{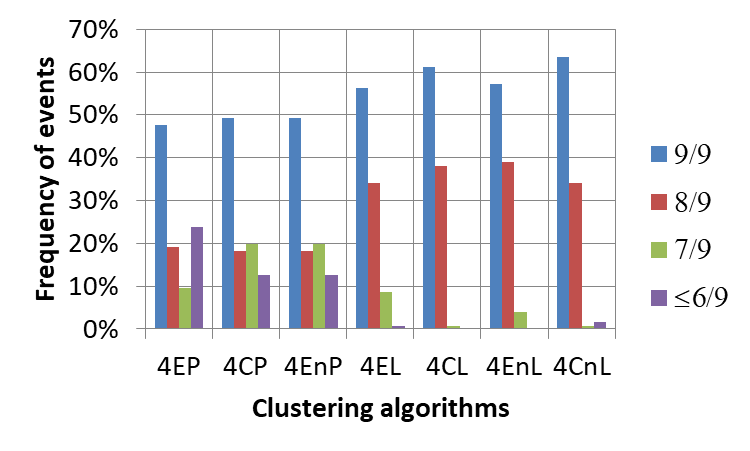}
		}
		\quad
		\subfloat[] {\label{figure08b}
			\includegraphics[width=0.75\textwidth]{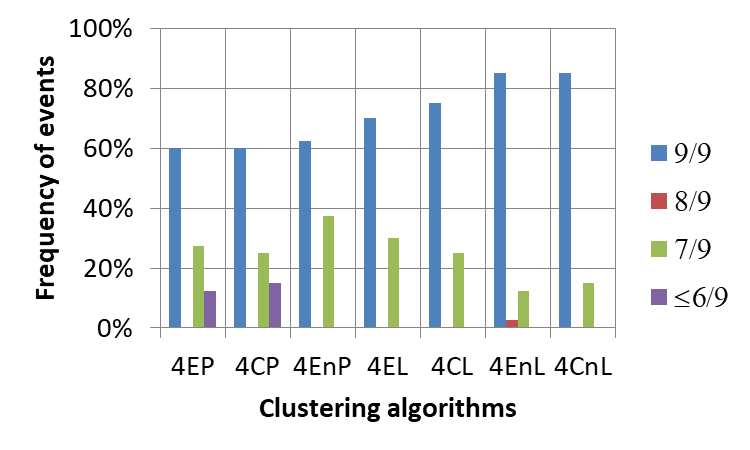}
		}
		\caption{Distributions of the efficiency of the clustering algorithms - Houses Data, four clusters: (a) All possible $126$ combinations of initial points; (b) Forty quartets with the highest entropy}\label{figure08}}
\end{figure}

When clustering due to similarity of primary variables to the principal components, as well as in the case of spectral clustering with the matrix of determination coefficients as the similarity matrix, for nine variables clustered into four clusters, the number of all quartets of different initial points is equal to $\binom{9}{4}=126$.
The clustering algorithm was launched for each of the $126$ quartets of initial points.
The following versions of clustering were examined:
\begin{enumerate}
	\item For 126 different quartets of initial points, variables have been clustered due to their similarity to the principal components:
	      \begin{itemize}
		      \item $4EP$ - Euclidean metric;
		      \item $4CP$ - cosine measure of dissimilarity;
		      \item $4EnP$ - Euclidean metric, normalized points for clustering.
	      \end{itemize}
	\item 	For 126 different quartets of initial points, variables were clustered using spectral methods, treating the matrix of determination coefficients as a similarity matrix:
	      \begin{itemize}
		      \item $4EL$ - Laplacian of similarity matrix, Euclidean metric;
		      \item $4CL$ - Laplacian of similarity matrix, cosine measure of dissimilarity;
		      \item $4EnL$ - normalized Laplacian of similarity matrix, Euclidean metric;
		      \item $4CnL$ - normalized Laplacian of similarity matrix, cosine measure of dissimilarity.
	      \end{itemize}
\end{enumerate}
Obtained results of all algorithms were compared with the distribution of nodes in the $4Man$ pattern.
The efficiency of clustering depended on both the version of the algorithm and the selected quartet of initial points.
Table \ref{table06} presents the basic statistics of all clustering procedures.
Table \ref{table07} shows the distribution of clustering efficiency for different clustering algorithms.
The results from Table \ref{table07} were transferred to the bar chart (Figure \ref{figure08}).
For the nine variables different levels of clustering validity were obtained: from four variables classified correctly $(4/9)$, up to nine $(9/9)$.
\begin{table}
	\centering
	\caption{Basic statistics of clustering procedures - Houses Data, four clusters, 126 estimates}\label{table06}
	\fontsize{9.5}{13.5}\selectfont{
		\begin{tabular}{c||c|c|c|c|c|c|c} \hline
			Statistics         & 4EP      & 4CP      & 4EnP     & 4EL      & 4CL      & 4EnL     & 4CnL     \\ \hline \hline
			Average efficiency & $81.6\%$ & $83.7\%$ & $83.7\%$ & $89.0\%$ & $91.2\%$ & $89.3\%$ & $91.1\%$ \\ \hline
			Median             & $7/9$    & $7/9$    & $7/9$    & $9/9$    & $9/9$    & $9/9$    & $9/9$    \\ \hline
			Mode               & $9/9$    & $9/9$    & $9/9$    & $9/9$    & $9/9$    & $9/9$    & $9/9$    \\ \hline
			Minimal efficiency & $4/9$    & $5/9$    & $5/9$    & $4/9$    & $6/9$    & $6/9$    & $4/9$    \\ \hline
			Maximal efficiency & $9/9$    & $9/9$    & $9/9$    & $9/9$    & $9/9$    & $9/9$    & $9/9$    \\ \hline
		\end{tabular}}
\end{table}

The results in Tables \ref{table06} and \ref{table07}, as well as in Figure \ref{figure08a}, describe the statistics obtained for the efficiency of clustering for all possible combinations of initial points.
Because it was assumed that the efficiency of clustering may depend on the diversity of initial points, therefore the entropy for all subsets of initial points was estimated.
Fig. \ref{figure08b} shows the distribution of clustering results for forty sets of initial points with the highest entropy.
When comparing the clustering efficiency distributions shown in Figures \ref{figure08a} and \ref{figure08b}, it is possible to observe greater clustering efficiency for sets of initial points with greater entropy.

\begin{table}
	\centering
	\caption{Distributions of the efficiency of clustering algorithms - Houses Data, four clusters, 126 estimates}\label{table07}
	\fontsize{9.5}{13.5}\selectfont{
		\begin{tabular}{c||c|c|c|c|c|c|c} \hline
			Performance levels & 4EP      & 4CP      & 4EnP     & 4EL      & 4CL      & 4EnL     & 4CnL     \\ \hline \hline
			$9/9$              & $47.6\%$ & $49.2\%$ & $49.2\%$ & $56.3\%$ & $61.1\%$ & $57.1\%$ & $63.5\%$ \\ \hline
			$8/9$              & $19.0\%$ & $18.3\%$ & $18.3\%$ & $34.1\%$ & $38.1\%$ & $38.9\%$ & $34.1\%$ \\ \hline
			$7/9$              & $9.5\%$  & $19.8\%$ & $19.8\%$ & $8.7\%$  & $0.8\%$  & $4.0\%$  & $0.8\%$  \\ \hline
			$\le 6/9$          & $23.8\%$ & $12.7\%$ & $12.7\%$ & $0.8\%$  & $0\%$    & $0\%$    & $1.6\%$  \\ \hline
		\end{tabular}}
\end{table}

\subsubsection{Clustering of primary variables into five clusters}
With a small change in $\epsilon$ from $45\%$ to $50\%$, the relation obtains an additional, fifth connected component (Figure \ref{figure07b}).
Therefore, clustering into five clusters was also performed.
Clustering into five clusters has been analyzed both because of the similarity of the primary variables to the principal components, and because of the mutual similarity of the primary variables first described by the matrix of relation, and then by the matrix of determination coefficients.

When clustering relation nodes into five clusters, the k-means method used by various spectral algorithms always had exactly one quintet of different initial points.
As could be expected, the compatibility of spectral methods with 5Man clusters (Figure \ref{figure07b}) was full.

\begin{figure}
	\centering
	{\subfloat[] {\label{figure09a}
			\includegraphics[width=0.75\textwidth]{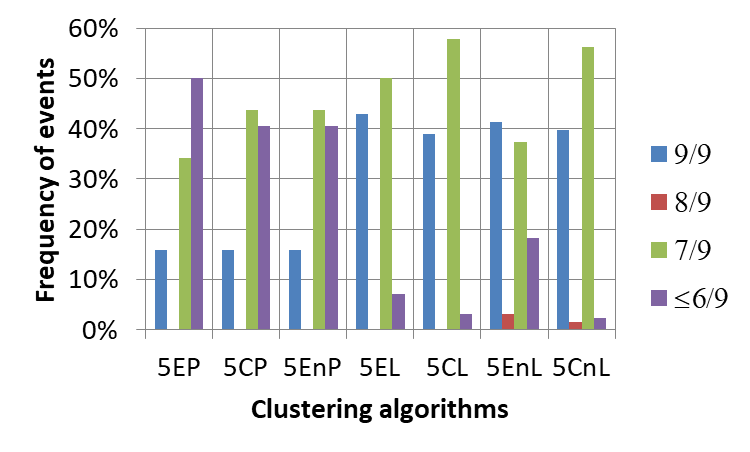}
		}
		\quad
		\subfloat[] {\label{figure09b}
			\includegraphics[width=0.75\textwidth]{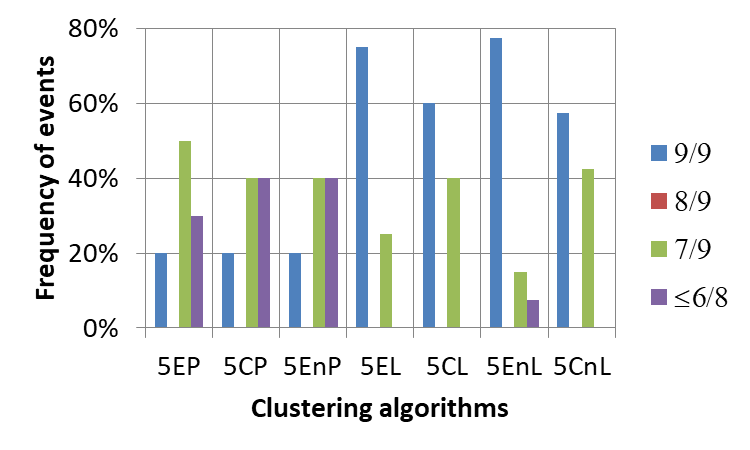}
		}
		\caption{Distributions of the efficiency of the clustering algorithms - Houses Data, five clusters: (a) All possible $126$ combinations of initial points; (b) Forty quintets with the highest entropy}\label{figure09}}
\end{figure}

On the other hand, for clustering according to the similarity of the primary variables to the principal components, the number of possible different initial quintets was greater.
Clustering a set of nine elements into five clusters requires five initial points.
The number of all possible different quintets of initial points are equal to $\binom{9}{5}=126$.
The clustering was performed for all possible combinations of initial points.
Table \ref{table08} shows statistics of clustering procedures. Table \ref{table09} as well as Figure \ref{figure09} show the distributions of the efficiency of the clustering algorithms.

Also, when clustering into five clusters, entropy was estimated for all initial quintets.
Comparing the results presented in Figures \ref{figure09a} and \ref{figure09b}, it can be seen that for points with greater entropy the efficiency of clustering is greater.

\begin{table}
	\centering
	\caption{Basic statistics of clustering procedures - Houses Data, five clusters, 126 estimates}\label{table08}
	\fontsize{9.5}{13.5}\selectfont{
		\begin{tabular}{c||c|c|c|c|c|c|c} \hline
			Statistics         & 5EP      & 5CP    & 5EnP   & 5EL      & 5CL      & 5EnL     & 5CnL     \\ \hline \hline
			Average efficiency & $73.7\%$ & $75\%$ & $75\%$ & $85.7\%$ & $85.7\%$ & $84.7\%$ & $86.2\%$ \\ \hline
			Median             & $7/9$    & $7/9$  & $7/9$  & $7/9$    & $7/9$    & $7/9$    & $7/9$    \\ \hline
			Mode               & $7/9$    & $7/9$  & $7/9$  & $7/9$    & $7/9$    & $9/9$    & $7/9$    \\ \hline
			Minimal efficiency & $3/9$    & $5/9$  & $5/9$  & $5/9$    & $5/9$    & $5/9$    & $5/9$    \\ \hline
			Maximal efficiency & $9/9$    & $9/9$  & $9/9$  & $9/9$    & $9/9$    & $9/9$    & $9/9$    \\ \hline
		\end{tabular}}
\end{table}

\begin{table}
	\centering
	\caption{Distributions of the efficiency of the clustering algorithms - Houses Data, five clusters, 126 estimates}\label{table09}
	\fontsize{9.5}{13.5}\selectfont{
		\begin{tabular}{c||c|c|c|c|c|c|c} \hline
			Performance levels & 5EP      & 5CP      & 5EnP     & 5EL      & 5CL      & 5EnL     & 5CnL     \\ \hline \hline
			$9/9$              & $15.9\%$ & $15.9\%$ & $15.9\%$ & $42.9\%$ & $38.9\%$ & $44.4\%$ & $41.3\%$ \\ \hline
			$8/9$              & $34.1\%$ & $43.7\%$ & $43.7\%$ & $50.0\%$ & $57.9\%$ & $37.3\%$ & $56.3\%$ \\ \hline
			$7/9$              & $33.3\%$ & $23.8\%$ & $23.8\%$ & $0\%$    & $0\%$    & $12.7\%$ & $0\%$    \\ \hline
			$\le 6/9$          & $16.7\%$ & $16.7\%$ & $16.7\%$ & $7.1\%$  & $3.2\%$  & $5.6\%$  & $2.4\%$  \\ \hline
		\end{tabular}}
\end{table}

\begin{figure}
	\centering
	\includegraphics[width=0.75\textwidth]{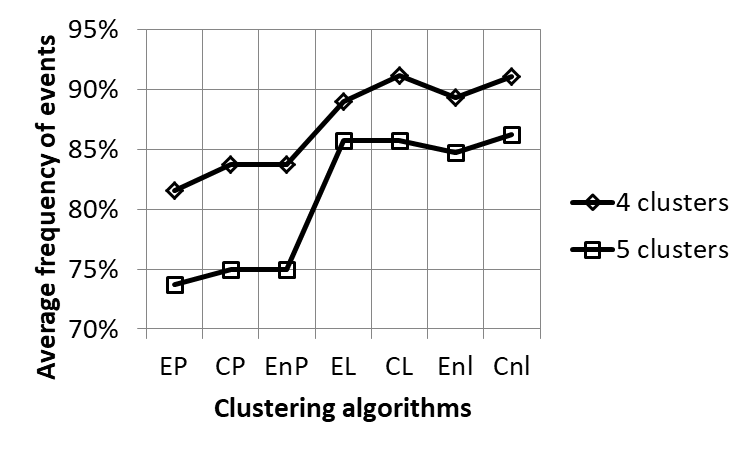}
	\caption{Average clustering efficiency for Houses Data}\label{figure_a2}
\end{figure}

\begin{figure}
	\centering
	\includegraphics[width=0.75\textwidth]{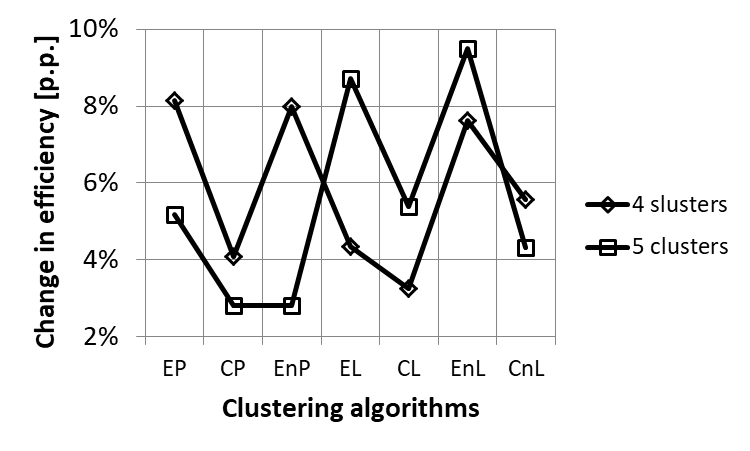}
	\caption{Change in the average efficiency of clustering (in percentage points) for initial points with the largest entropy for Houses Data}\label{CE_ii}
\end{figure}

\subsubsection{Efficiency of clustering the Houses Data}
The analysis of Tables \ref{table06} and \ref{table07}, as well as Figure \ref{figure08}, will assess the efficiency of clustering into four clusters.
In the same way, the analysis of Tables \ref{table08} and \ref{table09} and Figure \ref{figure09} will give the opportunity to evaluate the efficiency of clustering into five clusters.
Figure \ref{figure_a2} shows the average clustering efficiency for both of the above cases.

The efficiency of clustering into four clusters is greater than the efficiency of clustering into five clusters.
In turn, in both cases, clustering using spectral methods is more effective.

Another conclusion concerns the impact of the entropy of sets of initial points in the k-means algorithm on the efficiency of clustering.
To estimate this impact, the average clustering efficiency for all the initial point combinations was calculated first.
Then the average clustering efficiency was calculated for about $30\%$ of the sets of initial points with the highest entropy.
In Fig. \ref{CE_ii}, the differences of both results are given in percentage points.
The above difference shows how the efficiency of clustering changes with the increase in entropy of initial points in the k-means algorithm, for cases of clustering into four and five clusters.

\section{The dataset No. 3}
\begin{figure}
	\centering
	\includegraphics[width=0.75\textwidth]{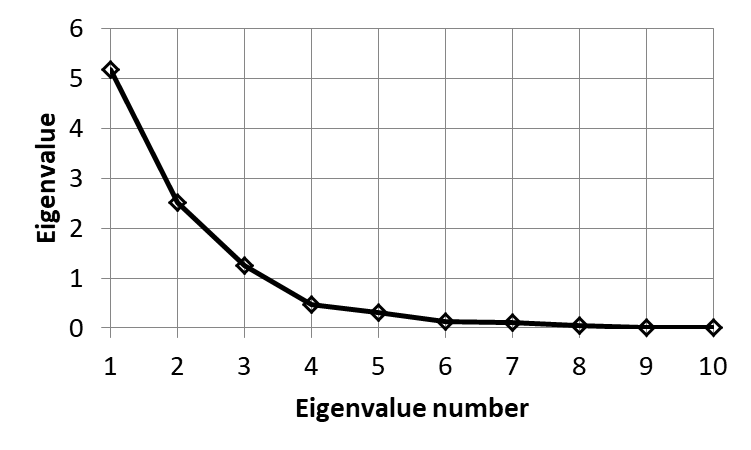}
	\caption{The scree plot for dataset No. 3}\label{figure10}
\end{figure}

As the third set of data, the data collected for the purposes of the master's thesis \cite{Kaliszewski2012} was used for the analysis.
The set contains ten random variables.
Each variable was measured 308 times.
Correlation coefficients have been estimated for this dataset:
\begin{equation}\label{eq52}
	C_R {=}
	\begin{bmatrix}
		1     & 0.81  & 0.78  & 0.39  & 0.87  & 0.78  & 0.44  & -0.33 & 0.08  & 0.53  \\
		0.81  & 1     & 0.87  & 0.78  & 0.90  & 0.75  & 0.34  & 0.11  & -0.28 & 0.47  \\
		0.78  & 0.87  & 1     & 0.62  & 0.72  & 0.64  & 0.48  & -0.05 & -0.29 & 0.78  \\
		0.39  & 0.78  & 0.62  & 1     & 0.45  & 0.42  & 0.06  & 0.62  & -0.65 & 0.25  \\
		0.87  & 0.90  & 0.72  & 0.45  & 1     & 0.82  & 0.34  & -0.21 & 0.07  & 0.34  \\
		0.78  & 0.75  & 0.64  & 0.42  & 0.82  & 1     & 0.24  & -0.10 & -0.04 & 0.33  \\
		0.44  & 0.34  & 0.48  & 0.06  & 0.34  & 0.24  & 1     & -0.36 & 0.02  & 0.55  \\
		-0.33 & 0.11  & -0.05 & 0.62  & -0.21 & -0.10 & -0.36 & 1     & -0.85 & -0.17 \\
		0.08  & -0.28 & -0.29 & -0.65 & 0.07  & -0.04 & 0.02  & -0.85 & 1     & -0.34 \\
		0.53  & 0.47  & 0.78  & 0.25  & 0.34  & 0.33  & 0.55  & -0.17 & -0.34 & 1
	\end{bmatrix}.
\end{equation}
Using matrix (\ref{eq52}), principal components were found.
Table \ref{table10} shows the cumulative variance of the principal components, measured by the sum of the successive eigenvalues.
The two principal components explain over $76\%$ of the average variance of the primary variables, the three principal components explain over $89\%$ of this variance.
Figure \ref{figure10} shows a scree plot.
There are three variables on the scree.

\begin{table}
	\centering
	\caption{The percentage of variance explained by the successive principal components for dataset No. 3}\label{table10}
	\fontsize{9.5}{13.5}\selectfont{
		\begin{tabular}{c|c|c|c|c} \hline
			\multirow{2}{*}{No.} & \multirow{2}{*}{Eigenvalue} & Cumulative  & Percentage of variance & Cumulative             \\
			                     &                             & eigenvalues & explained by each PC   & percentage of variance \\ \hline \hline
			$1$                  & $5.164$                     & $5.164$     & $51.6\%$               & $51.6\%$               \\ \hline
			$2$                  & $2.507$                     & $7.671$     & $25.1\%$               & $76.7\%$               \\ \hline
			$3$                  & $1.248$                     & $8.919$     & $12.5\%$               & $89.2\%$               \\ \hline
			$4$                  & $0.461$                     & $9.380$     & $4.6\%$                & $93.8\%$               \\ \hline
			$5$                  & $0.320$                     & $9.700$     & $3.2\%$                & $97.0\%$               \\ \hline
			$6$                  & $0.126$                     & $9.826$     & $1.3\%$                & $98.3\%$               \\ \hline
			$7$                  & $0.116$                     & $9.942$     & $1.2\%$                & $99.4\%$               \\ \hline
			$8$                  & $0.040$                     & $9.982$     & $0.4\%$                & $99.8\%$               \\ \hline
			$9$                  & $0.017$                     & $10.000$    & $0.2\%$                & $100.0\%$              \\ \hline
			$10$                 & $0.000$                     & $10$        & $0$                    & $100.0\%$              \\ \hline
		\end{tabular}}
\end{table}

\begin{table}
	\centering
	\caption{Determination coefficients between primary variables and principal components}\label{table11}
	\fontsize{8}{14}\selectfont{
		\begin{tabular}{c||c|c|c|c|c|c|c|c|c|c} \hline
			       & $V1$     & $V2$     & $V3$     & $V4$     & $V5$     & $V6$     & $V7$     & $V8$     & $V9$     & $V10 $    \\ \hline \hline
			$PC1$  & $79.2\%$ & $91.0\%$ & $87.7\%$ & $42.9\%$ & $76.5\%$ & $65.1\%$ & $24.6\%$ & $0.2\%$  & $6.1\%$  & $43.2\% $ \\ \hline
			$PC2$  & $9.7\%$  & $2.0\%$  & $0.1\%$  & $44.6\%$ & $4.7\%$  & $1.6\%$  & $13.1\%$ & $95.7\%$ & $79.0\%$ & $0.3\%$   \\ \hline
			$PC3$  & $1.7\%$  & $3.6\%$  & $3.8\%$  & $1.9\%$  & $12.5\%$ & $12.8\%$ & $34.3\%$ & $0.7\%$  & $9.8\%$  & $43.6\%$  \\ \hline
			$PC4$  & $0.4\%$  & $0.6\%$  & $1.9\%$  & $2.1\%$  & $0.5\%$  & $0.5\%$  & $26.9\%$ & $1.0\%$  & $0.2\%$  & $12.0\%$  \\ \hline
			$PC5$  & $0.0\%$  & $1.8\%$  & $2.5\%$  & $4.9\%$  & $0.0\%$  & $17.5\%$ & $1.1\%$  & $1.1\%$  & $3.1\%$  & $0.0\%$   \\ \hline
			$PC6$  & $2.2\%$  & $0.1\%$  & $1.7\%$  & $1.8\%$  & $2.8\%$  & $2.5\%$  & $0\%$    & $0.3\%$  & $1.1\%$  & $0.0\%$   \\ \hline
			$PC7$  & $6.6\%$  & $0.7\%$  & $0.4\%$  & $0.9\%$  & $2.9\%$  & $0\%$    & $0\%$    & $0.0\%$  & $0.0\%$  & $0.0\%$   \\ \hline
			$PC8$  & $0.2\%$  & $0.0\%$  & $1.8\%$  & $0.9\%$  & $0.1\%$  & $0\%$    & $0\%$    & $0.5\%$  & $0.0\%$  & $0.6\%$   \\ \hline
			$PC9$  & $0\%$    & $0.1\%$  & $0\%$    & $0\%$    & $0.1\%$  & $0\%$    & $0\%$    & $0.7\%$  & $0.6\%$  & $0.3\%$   \\ \hline
			$PC10$ & $0\%$    & $0\%$    & $0\%$    & $0\%$    & $0\%$    & $0\%$    & $0\%$    & $0\%$    & $0\%$    & $0\%$     \\ \hline
		\end{tabular}}
\end{table}

\subsection{Similarity of the primary variables to the principal components}
Having the principal components, the coefficients of determination between primary variables and principal components were found (Table \ref{table11}).
Calculated coefficients of determination show that variables with numbers $1$, $2$ and $3$, as well as $5$ and $6$ have more than $65\%$ of the common variance with the first principal component.
This means that all of these variables are significantly similar to the first principal component.
In the same way, variables with numbers $8$ and $9$ are significantly similar to the second principal component.
On the other hand, for variables with numbers $4$, $7$ and $10$ there is no dominant principal component that could represent these variables significantly.
To represent more than half of the variance of these variables, at least a third principal component is required.
Assuming that three principal components are enough to represent the primary variables, it can also be assumed that these variables can be clustered into three clusters.
Assuming that the set of primary variables will be clustered into three clusters, transposition of the matrix $M$ takes the form:
\begin{equation}\label{eq53}
	M^T {=}
	\begin{bmatrix}
		0.79 & 0.91 & 0.88 & 0.43 & 0.76 & 0.65 & 0.25 & 0.00 & 0.06 & 0.43 \\
		0.10 & 0.02 & 0.00 & 0.45 & 0.05 & 0.02 & 0.13 & 0.96 & 0.79 & 0.00 \\
		0.02 & 0.04 & 0.04 & 0.02 & 0.13 & 0.13 & 0.34 & 0.01 & 0.10 & 0.44
	\end{bmatrix}.
\end{equation}

\begin{table}
	\centering
	\caption{The level of explanation of the variance of primary variables by the four principal components}\label{table12}
	\fontsize{8}{14}\selectfont{
		\begin{tabular}{c||c|c|c|c|c|c|c|c|c|c} \hline
			         & $V1$     & $V2$     & $V3$     & $V4$     & $V5$     & $V6$     & $V7$     & $V8$     & $V9$     & $V10 $   \\ \hline \hline
			$PC1$    & $79.2\%$ & $91.0\%$ & $87.7\%$ & $42.9\%$ & $76.5\%$ & $65.1\%$ & $24.6\%$ & $0.2\%$  & $6.1\%$  & $43.2\%$ \\ \hline
			$PC2$    & $9.7\%$  & $2.0\%$  & $0.1\%$  & $44.6\%$ & $4.7\%$  & $1.6\%$  & $13.1\%$ & $95.7\%$ & $79.0\%$ & $0.3\%$  \\ \hline
			$PC3$    & $1.7\%$  & $3.6\%$  & $3.8\%$  & $1.9\%$  & $12.5\%$ & $12.8\%$ & $34.3\%$ & $0.7\%$  & $9.8\%$  & $43.6\%$ \\ \hline
			$\Sigma$ & $90.6\%$ & $96.6\%$ & $91.6\%$ & $89.4\%$ & $93.7\%$ & $79.5\%$ & $71.9\%$ & $96.5\%$ & $94.9\%$ & $87.1\%$ \\ \hline
		\end{tabular}}
\end{table}

\begin{figure}
	\centering
	{\subfloat[] {\label{figure11a}
			\includegraphics[width=0.29\textwidth]{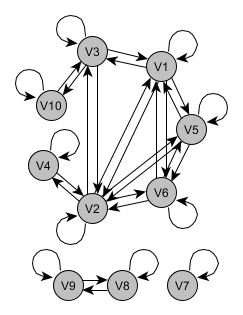}
		}
		\quad
		\subfloat[] {\label{figure11b}
			\includegraphics[width=0.29\textwidth]{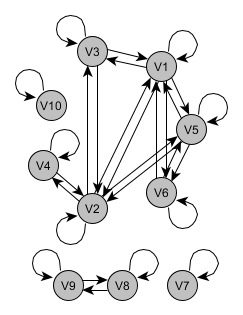}
		}
		\quad
		\subfloat[] {\label{figure11c}
			\includegraphics[width=0.29\textwidth]{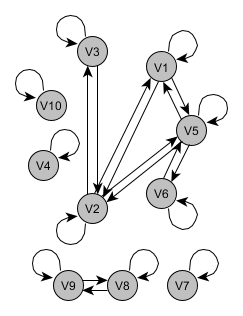}
		}
		\caption{Fragment of the evolution of the relation for dataset No. 3: (a) $\epsilon=55\%$, relation of similarity with three classes of similarity; (b) $\epsilon=60.1\%$, similarity relation with four similarity classes; (c) $\epsilon=65\%$, similarity relation with five similarity classes}	\label{figure11}}
\end{figure}

\subsection{The relation established on the level of similarity}
Based on the matrix of correlation coefficients, the matrix of determination coefficients was found:
\begin{equation}\label{eq54}
	S {=}
	\begin{bmatrix}
		1    & 0.66 & 0.61 & 0.15 & 0.75 & 0.61 & 0.20 & 0.11 & 0.01 & 0.28 \\
		0.66 & 1    & 0.76 & 0.60 & 0.80 & 0.57 & 0.11 & 0.01 & 0.08 & 0.22 \\
		0.61 & 0.76 & 1    & 0.39 & 0.52 & 0.41 & 0.23 & 0.00 & 0.08 & 0.60 \\
		0.15 & 0.60 & 0.39 & 1    & 0.20 & 0.17 & 0.00 & 0.38 & 0.42 & 0.06 \\
		0.75 & 0.80 & 0.52 & 0.20 & 1    & 0.68 & 0.11 & 0.04 & 0.01 & 0.11 \\
		0.61 & 0.57 & 0.41 & 0.17 & 0.68 & 1    & 0.06 & 0.01 & 0.00 & 0.11 \\
		0.20 & 0.11 & 0.23 & 0.00 & 0.11 & 0.06 & 1    & 0.13 & 0.00 & 0.31 \\
		0.11 & 0.01 & 0.00 & 0.38 & 0.04 & 0.01 & 0.13 & 1    & 0.71 & 0.03 \\
		0.01 & 0.08 & 0.08 & 0.42 & 0.01 & 0.00 & 0.00 & 0.71 & 1    & 0.12 \\
		0.28 & 0.22 & 0.60 & 0.06 & 0.11 & 0.11 & 0.31 & 0.03 & 0.12 & 1
	\end{bmatrix}.
\end{equation}
For $\epsilon$ in the range of $40\%$ to $70\%$, relation matrices were built up every $5\%$.
It was noted that in the case of $\epsilon$ in the range from $45\%$ to $60\%$, the relation graph has three connected components, while for $65\%$ the graph had five connected components.
The conclusion is that in the range of $60\% < \epsilon < 65\%$, the relation has four connected components.
After a thorough examination of this interval with a step of $0.1\%$, it turned out that for $\epsilon = 60.1\%$ and $60.2\%$ the relation graph has four connected components.
Figure \ref{figure11} shows a details of the evolution of the relation for $\epsilon = 55\%$, $\epsilon = 60.1\%$ and $\epsilon = 65\%$.
For $\epsilon = 55\%$, the following relation matrix was obtained:
\begin{equation}\label{eq55}
	R_{\epsilon=55\%} {=}
	\begin{bmatrix}
		1 & 1 & 1 & 0 & 1 & 1 & 0 & 0 & 0 & 0 \\
		1 & 1 & 1 & 1 & 1 & 1 & 0 & 0 & 0 & 0 \\
		1 & 1 & 1 & 0 & 0 & 0 & 0 & 0 & 0 & 1 \\
		0 & 1 & 0 & 1 & 0 & 0 & 0 & 0 & 0 & 0 \\
		1 & 1 & 0 & 0 & 1 & 1 & 0 & 0 & 0 & 0 \\
		1 & 1 & 0 & 0 & 1 & 1 & 0 & 0 & 0 & 0 \\
		0 & 0 & 0 & 0 & 0 & 0 & 1 & 0 & 0 & 0 \\
		0 & 0 & 0 & 0 & 0 & 0 & 0 & 1 & 1 & 0 \\
		0 & 0 & 0 & 0 & 0 & 0 & 0 & 1 & 1 & 0 \\
		0 & 0 & 1 & 0 & 0 & 0 & 0 & 0 & 0 & 1
	\end{bmatrix}.
\end{equation}
The graph of this relation is presented in Figure \ref{figure11a}.
\subsubsection{Laplacian of the relation graph}
For the relation (\ref{eq55}) Laplacian $L$ was estimated:
\begin{equation}\label{eq56}
	L {=}
	\begin{bmatrix}
		4  & -1 & -1 & 0  & -1 & -1 & 0 & 0  & 0  \\
		-1 & 5  & -1 & -1 & -1 & -1 & 0 & 0  & 0  \\
		-1 & -1 & 3  & 0  & 0  & 0  & 0 & 0  & 0  \\
		0  & -1 & 0  & 1  & 0  & 0  & 0 & 0  & 0  \\
		-1 & -1 & 0  & 0  & 3  & -1 & 0 & 0  & 0  \\
		-1 & -1 & 0  & 0  & -1 & 3  & 0 & 0  & 0  \\
		0  & 0  & 0  & 0  & 0  & 0  & 0 & 0  & 0  \\
		0  & 0  & 0  & 0  & 0  & 0  & 0 & 1  & -1 \\
		0  & 0  & 0  & 0  & 0  & 0  & 0 & -1 & 1
	\end{bmatrix}.
\end{equation}
For Laplacian (\ref{eq56}) eigenvalues were calculated:
\begin{equation}\label{eq57}
	\lambda = \{6.060,5.142,4,3.034,2,1.075,0.689,0,0,0\}
\end{equation}
Three zero eigenvalues indicate that the relation graph has three connected components.
This is confirmed by Figure \ref{figure11a}.
The eigenvectors corresponding to the above eigenvalues form the rows of the following matrix $V^T$:

\begin{equation}\label{eq58}
	V^T {=}
	\begin{bmatrix}
		-0.12 & 0.90  & -0.27 & -0.18 & -0.19 & -0.19 & 0 & 0     & 0    & 0.05  \\
		0.85  & -0.11 & -0.39 & 0.03  & -0.23 & -0.23 & 0 & 0     & 0    & 0.09  \\
		0     & 0     & 0     & 0     & -0.71 & 0.71  & 0 & 0     & 0    & 0     \\
		0.22  & 0.13  & 0.76  & -0.06 & -0.33 & -0.33 & 0 & 0     & 0    & -0.37 \\
		0     & 0     & 0     & 0     & 0     & 0     & 0 & -0.71 & 0.71 & 0     \\
		-0.27 & -0.06 & -0.02 & 0.77  & -0.35 & -0.35 & 0 & 0     & 0    & 0.29  \\
		-0.07 & -0.15 & 0.25  & -0.48 & -0.17 & -0.17 & 0 & 0     & 0    & 0.79  \\
		0     & 0     & 0     & 0     & 0     & 0     & 1 & 0     & 0    & 0     \\
		0     & 0     & 0     & 0     & 0     & 0     & 0 & 0.71  & 0.71 & 0     \\
		0.38  & 0.38  & 0.38  & 0.38  & 0.38  & 0.38  & 0 & 0     & 0    & 0.38
	\end{bmatrix}.
\end{equation}
The last three vectors are used for clustering into three clusters.
These vectors form the matrix $M$ used by the k-means algorithm. Its transposition has the form:
\begin{equation}\label{eq59}
	M^T {=}
	\begin{bmatrix}
		0    & 0    & 0    & 0    & 0    & 0    & 1 & 0    & 0    & 0    \\
		0    & 0    & 0    & 0    & 0    & 0    & 0 & 0.71 & 0.71 & 0    \\
		0.38 & 0.38 & 0.38 & 0.38 & 0.38 & 0.38 & 0 & 0    & 0    & 0.38
	\end{bmatrix}.
\end{equation}
\subsubsection{Normalized Laplacian of the relation graph}
For the relation (\ref{eq55}), the normalized Laplacian $L_n$ was also established:
\begin{equation}\label{eq60}
	L_n {=}
	\begin{bmatrix}
		0.8   & -0.18 & -0.22 & 0     & -0.22 & -0.22 & 0 & 0    & 0    & 0     \\
		-0.18 & 0.83  & -0.20 & -0.29 & -0.20 & -0.20 & 0 & 0    & 0    & 0     \\
		-0.22 & -0.20 & 0.75  & 0     & 0     & 0     & 0 & 0    & 0    & -0.35 \\
		0     & -0.29 & 0     & 0.5   & 0     & 0     & 0 & 0    & 0    & 0     \\
		-0.22 & -0.20 & 0     & 0     & 0.75  & -0.25 & 0 & 0    & 0    & 0     \\
		-0.22 & -0.20 & 0     & 0     & -0.25 & 0.75  & 0 & 0    & 0    & 0     \\
		0     & 0     & 0     & 0     & 0     & 0     & 0 & 0    & 0    & 0     \\
		0     & 0     & 0     & 0     & 0     & 0     & 0 & 0.5  & -0.5 & 0     \\
		0     & 0     & 0     & 0     & 0     & 0     & 0 & -0.5 & 0.5  & 0     \\
		0     & 0     & -0.35 & 0     & 0     & 0     & 0 & 0    & 0    & 0.5
	\end{bmatrix}.
\end{equation}
Solving the eigenproblem for the normalized Laplacian (\ref{eq60}), the eigenvalues were estimated:
\begin{equation}\label{eq61}
	\lambda = \{1.24,1.07,1,1,0.85,0.43,0.29,0,0,0\}.
\end{equation}
Three zero eigenvalues confirm the fact that the relation graph has three connected components (Figure \ref{figure11a}).
The corresponding eigenvectors form rows of the following matrix $V^T$:
\begin{equation}\label{eq62}
	V^T {=}
	\begin{bmatrix}
		-0.32 & -0.56 & 0.58 & 0.22  & 0.25  & 0.25  & 0 & 0     & 0    & -0.28 \\
		-0.67 & 0.62  & 0.23 & -0.31 & 0.04  & 0.04  & 0 & 0     & 0    & -0.14 \\
		0     & 0     & 0    & 0     & -0.71 & 0.71  & 0 & 0     & 0    & 0     \\
		0     & 0     & 0    & 0     & 0     & 0     & 0 & -0.71 & 0.71 & 0     \\
		0.43  & 0.15  & 0.50 & -0.12 & -0.36 & -0.36 & 0 & 0     & 0    & -0.51 \\
		-0.26 & 0.19  & 0.01 & 0.84  & -0.30 & -0.30 & 0 & 0     & 0    & 0.07  \\
		-0.09 & -0.17 & 0.46 & -0.23 & -0.26 & -0.26 & 0 & 0     & 0    & 0.75  \\
		0.43  & 0.47  & 0.38 & 0.27  & 0.38  & 0.38  & 0 & 0     & 0    & 0.27  \\
		0     & 0     & 0    & 0     & 0     & 0     & 1 & 0     & 0    & 0     \\
		0     & 0     & 0    & 0     & 0     & 0     & 0 & 0.71  & 0.71 & 0
	\end{bmatrix}.
\end{equation}
The last three eigenvectors should be used for clustering into three clusters.
After their transposition, a rectangular matrix is obtained, the rows of which are normalized to the unit length. The matrix obtained in this way, designated as $M$, is used by the k-means algorithm.
Its transposition has the form:
\begin{equation}\label{eq63}
	M^T {=}
	\begin{bmatrix}
		1 & 1 & 1 & 1 & 1 & 1 & 0 & 0 & 0 & 1 \\
		0 & 0 & 0 & 0 & 0 & 0 & 1 & 0 & 0 & 0 \\
		0 & 0 & 0 & 0 & 0 & 0 & 0 & 1 & 1 & 0
	\end{bmatrix}.
\end{equation}
\subsection{Spectral clustering based on the matrix of coefficients of determination}
The matrix of determination coefficients (\ref{eq54}) is a similarity matrix for the set of primary variables.
Treating this matrix as a generalized graph adjacency matrix, both Laplacian $L$ and the normalized Laplacian $L_n$ were formed for it.
\subsubsection{Laplacian of determination coefficients matrix}
The Laplacian $L$ formed for the matrix (\ref{eq54}) has the form:
\begin{equation}\label{eq64}
	L {=}
	\begin{bmatrix}
		3.37  & -0.66 & -0.61 & -0.15 & -0.75 & -0.61 & -0.20 & -0.11 & -0.01 & -0.28 \\
		-0.66 & 3.81  & -0.76 & -0.60 & -0.80 & -0.57 & -0.11 & -0.01 & -0.08 & -0.22 \\
		-0.61 & -0.76 & 3.60  & -0.39 & -0.52 & -0.41 & -0.23 & 0.00  & -0.08 & -0.60 \\
		-0.15 & -0.60 & -0.39 & 2.39  & -0.20 & -0.17 & 0.00  & -0.38 & -0.42 & -0.06 \\
		-0.75 & -0.80 & -0.52 & -0.20 & 3.23  & -0.68 & -0.11 & -0.04 & -0.01 & -0.11 \\
		-0.61 & -0.57 & -0.41 & -0.17 & -0.68 & 2.62  & -0.06 & -0.01 & 0.00  & -0.11 \\
		-0.20 & -0.11 & -0.23 & 0.00  & -0.11 & -0.06 & 1.15  & -0.13 & 0.00  & -0.31 \\
		-0.11 & -0.01 & 0.00  & -0.38 & -0.04 & -0.01 & -0.13 & 1.42  & -0.71 & -0.03 \\
		-0.01 & -0.08 & -0.08 & -0.42 & -0.01 & 0.00  & 0.00  & -0.71 & 1.43  & -0.12 \\
		-0.28 & -0.22 & -0.60 & -0.06 & -0.11 & -0.11 & -0.31 & -0.03 & -0.12 & 1.84
	\end{bmatrix}.
\end{equation}
Its eigenvalues after sorting are equal:
\begin{equation}\label{eq65}
	\lambda = \{4.60,4.13,4.03,3.39,2.62,2.18,1.95,1.23,0.73,0\}.
\end{equation}
Successive eigenvalues correspond to successive eigenvectors, which are the rows of the following matrix $V^T$:
\begin{equation}\label{eq66}
	V^T {=}
	\begin{bmatrix}
		-0.05 & 0.84  & -0.44 & -0.13 & -0.27 & -0.03 & 0.01  & 0.02  & 0.00  & 0.05  \\
		-0.70 & 0.27  & 0.63  & -0.18 & 0.01  & 0.03  & -0.01 & 0.05  & -0.01 & -0.10 \\
		0.45  & 0.02  & 0.46  & -0.07 & -0.75 & 0.04  & -0.02 & 0.00  & 0.00  & -0.14 \\
		-0.33 & -0.20 & -0.23 & 0.21  & -0.35 & 0.78  & 0.04  & -0.01 & -0.03 & 0.14  \\
		-0.17 & 0.08  & 0.06  & 0.85  & -0.17 & -0.36 & 0.04  & -0.10 & -0.25 & 0.03  \\
		-0.02 & 0.06  & -0.01 & 0.13  & 0.03  & 0.05  & 0.19  & -0.66 & 0.61  & -0.37 \\
		-0.10 & -0.09 & 0.07  & -0.11 & -0.18 & -0.22 & -0.21 & -0.29 & 0.30  & 0.82  \\
		-0.15 & -0.19 & -0.11 & -0.18 & -0.21 & -0.26 & 0.87  & 0.13  & 0.01  & 0.09  \\
		-0.18 & -0.15 & -0.17 & 0.13  & -0.19 & -0.21 & -0.24 & 0.59  & 0.61  & -0.19 \\
		0.32  & 0.32  & 0.32  & 0.32  & 0.32  & 0.32  & 0.32  & 0.32  & 0.32  & 0.32
	\end{bmatrix}.
\end{equation}
Assuming that the primary variables are clustered into three clusters, the last three rows of the matrix (\ref{eq66}) form the $M$ matrix for clustering with the k-means algorithm. The transposition of matrix $M$ is as follows:
\begin{equation}\label{eq67}
	M^T {=}
	\begin{bmatrix}
		-0.15 & -0.19 & -0.11 & -0.18 & -0.21 & -0.26 & 0.87  & 0.13 & 0.01 & 0.09  \\
		-0.18 & -0.15 & -0.17 & 0.13  & -0.19 & -0.21 & -0.24 & 0.59 & 0.61 & -0.19 \\
		0.32  & 0.32  & 0.32  & 0.32  & 0.32  & 0.32  & 0.32  & 0.32 & 0.32 & 0.32
	\end{bmatrix}.
\end{equation}
\subsubsection{Normalized Laplacian of determination coefficients matrix}
For the matrix of coefficients of determination (\ref{eq54}), the normalized Laplacian $L_n$ was also formed:
\begin{equation}\label{eq68}
	L_n {=}
	\begin{bmatrix}
		0.77  & -0.14 & -0.14 & -0.04 & -0.18 & -0.15 & -0.06 & -0.03 & 0.00  & -0.08 \\
		-0.14 & 0.79  & -0.16 & -0.15 & -0.18 & -0.14 & -0.03 & 0.00  & -0.02 & -0.06 \\
		-0.14 & -0.16 & 0.78  & -0.10 & -0.12 & -0.10 & -0.07 & 0.00  & -0.02 & -0.17 \\
		-0.04 & -0.15 & -0.10 & 0.70  & -0.05 & -0.05 & 0.00  & -0.13 & -0.15 & -0.02 \\
		-0.18 & -0.18 & -0.12 & -0.05 & 0.76  & -0.17 & -0.04 & -0.01 & 0.00  & -0.03 \\
		-0.15 & -0.14 & -0.10 & -0.05 & -0.17 & 0.72  & -0.02 & 0.00  & 0.00  & -0.03 \\
		-0.06 & -0.03 & -0.07 & 0.00  & -0.04 & -0.02 & 0.54  & -0.06 & 0.00  & -0.12 \\
		-0.03 & 0.00  & 0.00  & -0.13 & -0.01 & 0.00  & -0.06 & 0.59  & -0.29 & -0.01 \\
		0.00  & -0.02 & -0.02 & -0.15 & 0.00  & 0.00  & 0.00  & -0.29 & 0.59  & -0.04 \\
		-0.08 & -0.06 & -0.17 & -0.02 & -0.03 & -0.03 & -0.12 & -0.01 & -0.04 & 0.65
	\end{bmatrix}.
\end{equation}
Non-ascending sorted eigenvalues form the following sequence:
\begin{equation}\label{eq69}
	\lambda = \{0.99,0.95,0.95,0.90,0.88,0.77,0.69,0.50,0.27,0\}.
\end{equation}
The corresponding eigenvectors are rows in the matrix $V^T$:
\begin{equation}\label{eq70}
	V^T {=}
	\begin{bmatrix}
		0.01  & 0.79  & -0.29 & -0.28 & -0.44 & 0.04  & -0.01 & 0.10  & 0.00  & 0.06  \\
		-0.51 & 0.12  & 0.70  & -0.33 & -0.03 & 0.09  & -0.02 & 0.23  & -0.07 & -0.24 \\
		0.56  & -0.19 & 0.37  & 0.02  & -0.63 & 0.09  & 0.02  & -0.12 & 0.12  & -0.28 \\
		-0.15 & -0.20 & -0.11 & 0.30  & -0.31 & 0.54  & -0.07 & 0.38  & -0.50 & 0.22  \\
		-0.38 & -0.06 & -0.12 & 0.02  & -0.11 & 0.58  & 0.15  & -0.48 & 0.49  & 0.00  \\
		-0.27 & 0.28  & 0.10  & 0.71  & -0.15 & -0.29 & 0.25  & -0.24 & -0.18 & -0.27 \\
		-0.14 & 0.02  & 0.30  & 0.13  & -0.23 & -0.24 & -0.47 & -0.23 & 0.11  & 0.69  \\
		-0.09 & -0.21 & 0.03  & -0.21 & -0.22 & -0.25 & 0.79  & 0.05  & -0.03 & 0.42  \\
		-0.20 & -0.16 & -0.17 & 0.22  & -0.22 & -0.22 & -0.08 & 0.60  & 0.61  & -0.11 \\
		0.35  & 0.37  & 0.36  & 0.31  & 0.35  & 0.32  & 0.25  & 0.26  & 0.26  & 0.29
	\end{bmatrix}.
\end{equation}
The last three eigenvectors become columns in the matrix $M$, the transposition of which is presented below:
\begin{equation}\label{eq71}
	M^T {=}
	\begin{bmatrix}
		-0.22 & -0.45 & 0.08  & -0.48 & -0.47 & -0.53 & 0.95  & 0.07 & -0.05 & 0.81  \\
		-0.48 & -0.35 & -0.42 & 0.51  & -0.47 & -0.48 & -0.10 & 0.91 & 0.92  & -0.21 \\
		0.85  & 0.82  & 0.90  & 0.71  & 0.75  & 0.70  & 0.30  & 0.40 & 0.39  & 0.55
	\end{bmatrix}.
\end{equation}
\subsection{Results of clustering of primary variables}
As before, all of the discussed clustering options were checked.
First, the relation matrix (\ref{eq55}), obtained from the matrix of determination coefficients for the assumed value of $\epsilon$ equal to $55\%$, was analyzed by spectral methods.
For the matrix of relation (\ref{eq55}), four versions of the algorithm were considered:
\begin{itemize}
	\item $3L55\%$ - Laplacian of relation matrix, Euclidean metric;
	\item $3CL55\%$ - Laplacian of relation matrix, cosine measure of dissimilarity;
	\item $3EnL55\%$ - normalized Laplacian of relation matrix, Euclidean metric;
	\item $3CnL55\%$ - normalized Laplacian of relation matrix, cosine measure of dissimilarity.
\end{itemize}
Because in the matrix $M$ in variants (\ref{eq59}) and (\ref{eq63}) there are only three different rows, this trio is the initial points for clustering and at the same time it creates the final centers of clusters. In this case, clustering is absolutely certain.
The above conclusion is confirmed by observation: the results obtained for the above variants of the algorithm were compared with the distribution of nodes in the 3Man graph, presented in Figure \ref{figure11a}.
For all variants, a full match of the results with the 3Man graph was obtained.

\begin{figure}
	\centering
	{\subfloat[] {\label{figure12a}
			\includegraphics[width=0.75\textwidth]{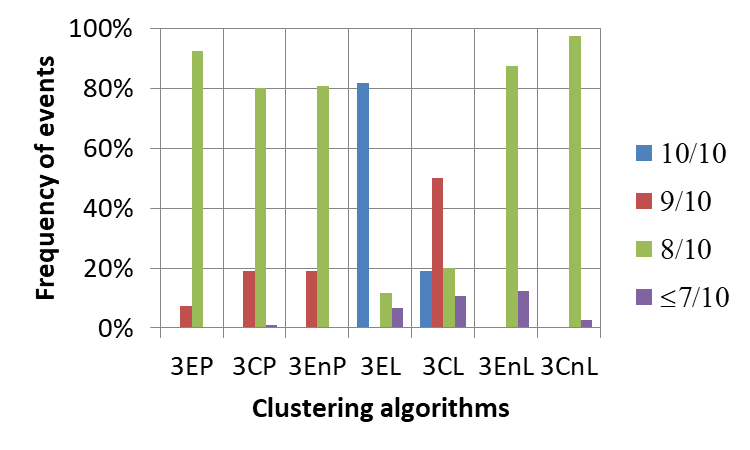}
		}
		\quad
		\subfloat[] {\label{figure12b}
			\includegraphics[width=0.75\textwidth]{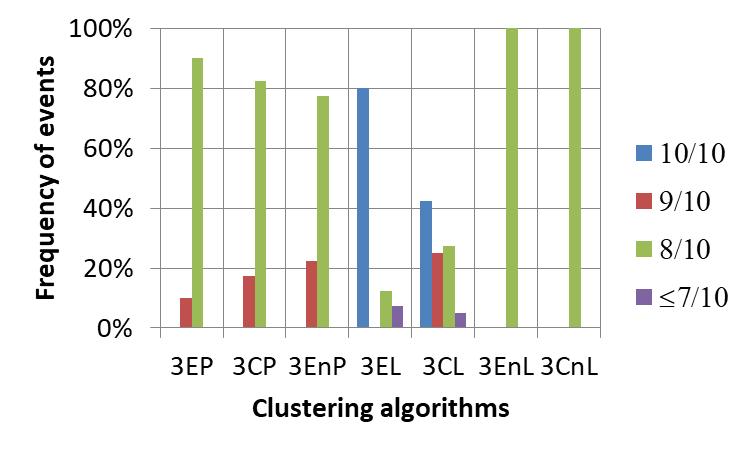}
		}
		\caption{Distributions of the efficiency of the clustering algorithms - Dataset No. 3, three clusters: (a) All possible $120$ combinations of initial points; (b) Forty initial trios with the highest entropy}\label{figure12}}
\end{figure}

While clustering because of the similarity to the principal components, as well as in the case of spectral clustering with the matrix of coefficients of determination as a similarity matrix, for ten variables clustered into three clusters, the number of all trios of different initial points is equal to $\binom{10}{3}=120$.
Clustering algorithms have been run for each of the $120$ trios.
The following versions of clustering algorithms were tested:
\begin{enumerate}
	\item For 120 different trios of initial points, primary variables have been clustered due to their similarity to the principal components:
	      \begin{itemize}
		      \item $3EP$ - Euclidean metric;
		      \item $3CP$ - a cosine measure of dissimilarity;
		      \item $3EnP$ - Euclidean metric, normalized clustering points;
	      \end{itemize}
	\item For 120 different initial trios, the primary variables were clustered using spectral methods, assuming a matrix of coefficients of determination as a similarity matrix:
	      \begin{itemize}
		      \item $ 3EL $ - Laplacian of similarity matrix, Euclidean metric;
		      \item $ 3CL $ - Laplacian of similarity matrix, cosine measure of dissimilarity;
		      \item $ 3EnL $ - normalized Laplacian of similarity matrix, Euclidean metric;
		      \item $ 3CnL $ - normalized Laplacian of similarity matrix, cosine measure of dissimilarity;
	      \end{itemize}
\end{enumerate}
Obtained results of all algorithms were compared with the distribution of nodes in the pattern marked as 3Man.
The efficiency of the algorithms studied depended on both the version of the algorithm as well as the selected initial trio.
Table \ref{table06} shows basic statistics of clustering algorithms.
Table \ref{table07} shows the distribution of clustering efficiency for different clustering algorithms.
The results from Table \ref{table07} were transferred to the bar chart (Figure \ref{figure12a}).

As before, the effect of entropy of initial trios on the efficiency of clustering was also examined here.
It can be seen that for forty initial points with the highest entropy, the efficiency of clustering improves.

\begin{table}
	\centering
	\caption{Basic statistics of clustering procedures - Dataset No. 3, three clusters}\label{table13}
	\fontsize{9.5}{13.5}\selectfont{
		\begin{tabular}{c||c|c|c|c|c|c|c} \hline
			Statistics         & $3EP$    & $3CP$    & $3EnP$   & $3EL$    & $3CL$    & $3EnL$   & $3CnL$   \\ \hline \hline
			Average efficiency & $80.7\%$ & $81.8\%$ & $81.9\%$ & $95.7\%$ & $87.0\%$ & $77.4\%$ & $79.7\%$ \\ \hline
			Median             & $8/10$   & $8/10$   & $8/10$   & $10/10$  & $9/10$   & $8/10$   & $8/10$   \\ \hline
			Mode               & $8/10$   & $8/10$   & $8/10$   & $10/10$  & $9/10$   & $8/10$   & $8/10$   \\ \hline
			Minimal efficiency & $8/10$   & $7/10$   & $8/10$   & $7/10$   & $6/10$   & $5/10$   & $7/10$   \\ \hline
			Maximal efficiency & $9/10$   & $9/10$   & $9/10$   & $10/10$  & $10/10$  & $8/10$   & $8/10$   \\ \hline
		\end{tabular}}
\end{table}

\begin{table}
	\centering
	\caption{Distributions of the efficiency of the clustering algorithms - Dataset No. 3, three clusters}\label{table14}
	\fontsize{9.5}{13.5}\selectfont{
		\begin{tabular}{c||c|c|c|c|c|c|c} \hline
			Performance levels & $3EP$    & $3CP$    & $3EnP$   & $3EL$    & $3CL$    & $3EnL$   & $3CnL$    \\ \hline \hline
			$10/10$            & $0\%$    & $0\%$    & $0\%$    & $81.7\%$ & $19.2\%$ & $0\%$    & $0\% $    \\ \hline
			$9/10$             & $7.5\%$  & $19.2\%$ & $19.2\%$ & $0\%$    & $50\%$   & $0\%$    & $0\% $    \\ \hline
			$8/10$             & $92.5\%$ & $80\%$   & $80.8\%$ & $11.7\%$ & $20\%$   & $87.5\%$ & $97.5\% $ \\ \hline
			$\le 7/10$         & $0\%$    & $0.8\%$  & $0\%$    & $6.7\%$  & $10.8\%$ & $12.5\%$ & $2.5\% $  \\ \hline
		\end{tabular}}
\end{table}

\begin{table}
	\centering
	\caption{Basic statistics of clustering procedures - Dataset No. 3, four clusters}\label{table15}
	\fontsize{9.5}{13.5}\selectfont{
		\begin{tabular}{c||c|c|c|c|c|c|c} \hline
			Statistics         & $4EP$    & $4CP$    & $4EnP$   & $4EL$    & $4CL$    & $4EnL$   & $4CnL$   \\ \hline \hline
			Average efficiency & $72.7\%$ & $74.0\%$ & $74.0\%$ & $86.8\%$ & $79.2\%$ & $68.3\%$ & $72.2\%$ \\ \hline
			Median             & $8/10$   & $8/10$   & $8/10$   & $9/10$   & $8/10$   & $7/10$   & $8/10$   \\ \hline
			Mode               & $8/10$   & $8/10$   & $8/10$   & $10/10$  & $8/10$   & $8/10$   & $8/10$   \\ \hline
			Minimal efficiency & $6/10$   & $6/10$   & $5/10$   & $6/10$   & $6/10$   & $4/10$   & $5/10$   \\ \hline
			Maximal efficiency & $10/10$  & $10/10$  & $10/10$  & $10/10$  & $10/10$  & $8/10$   & $8/10$   \\ \hline
		\end{tabular}}
\end{table}

\begin{table}
	\centering
	\caption{Distributions of the efficiency of the clustering algorithms - Dataset No. 3, four clusters}\label{table16}
	\fontsize{9.5}{13.5}\selectfont{
		\begin{tabular}{c||c|c|c|c|c|c|c} \hline
			Performance levels & $4EP$    & $4CP$    & $4EnP$   & $4EL$    & $4CL$    & $4EnL$   & $4CnL$   \\ \hline \hline
			$10/10$            & $2.4\%$  & $1.4\%$  & $1.4\%$  & $46.2\%$ & $16.7\%$ & $0\%$    & $0\%$    \\ \hline
			$9/10$             & $4.8\%$  & $5.7\%$  & $5.7\%$  & $4.8\%$  & $15.7\%$ & $0\%$    & $0\%$    \\ \hline
			$8/10$             & $47.6\%$ & $54.8\%$ & $55.2\%$ & $21.4\%$ & $26.2\%$ & $43.8\%$ & $55.2\%$ \\ \hline
			$\le 7/10$         & $45.2\%$ & $38.1\%$ & $37.6\%$ & $27.6\%$ & $41.4\%$ & $56.2\%$ & $44.8\%$ \\ \hline
		\end{tabular}}
\end{table}

\begin{figure}
	\centering
	{\subfloat[] {\label{figure13a}
			\includegraphics[width=0.75\textwidth]{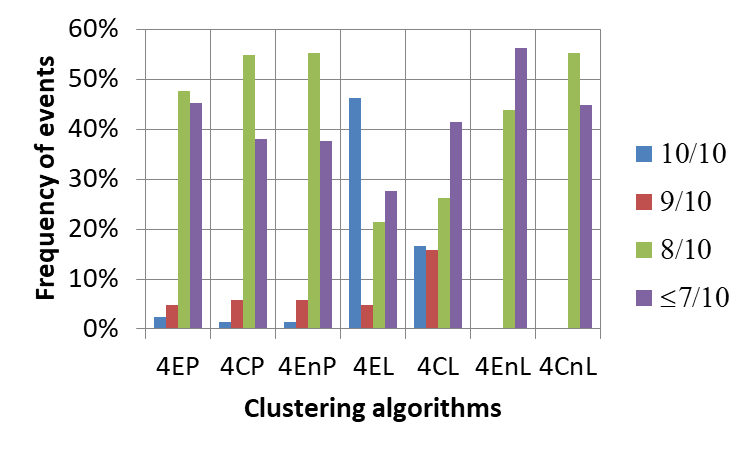}
		}
		\quad
		\subfloat[] {\label{figure13b}
			\includegraphics[width=0.75\textwidth]{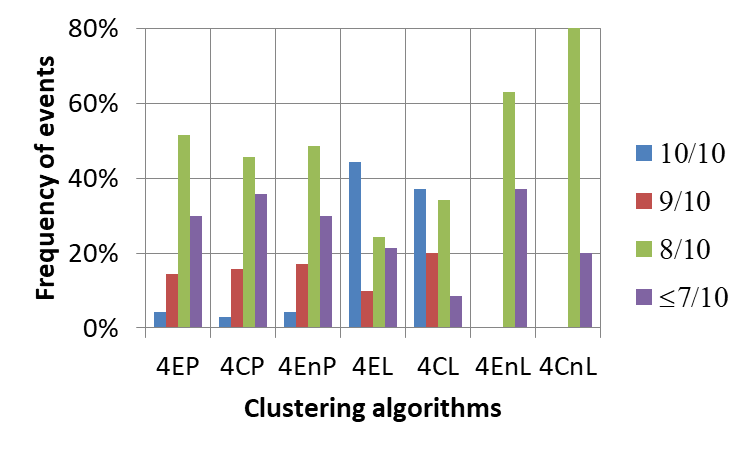}
		}
		\caption{Distributions of the efficiency of the clustering algorithms - Dataset No. 3, four clusters: (a) All possible $210$ combinations of initial points; (b) Seventy quartets with the highest entropy}\label{figure13}}
\end{figure}

At the value of $\epsilon = 60.1\%$, a relation was obtained whose graph has four connected components (Figure \ref{figure11b}).
When clustering due to similarity of primary variables to principal components, and also due to the mutual similarity of primary variables, the number of different initial quartets in the k-means algorithm is equal to $\binom{10}{4}=210$.
For all $210$ initial quartets, clustering was performed.
Descriptive statistics are presented in Table \ref{table15}.
Table \ref{table16} and Figure \ref{figure13a} show the distribution of results.

As before, an attempt was made to check the impact of entropy of initial points for the k-means algorithm on the efficiency of clustering.
In the case presented here, the bar chart of the clustering efficiency shown in Fig. \ref{figure13b}, obtained for seventy sets of initial points with the highest entropy shows that for the initial quartets with the highest entropy, clustering is more effective.

\begin{figure}
	\centering
	{\subfloat[] {\label{figure14a}
			\includegraphics[width=0.75\textwidth]{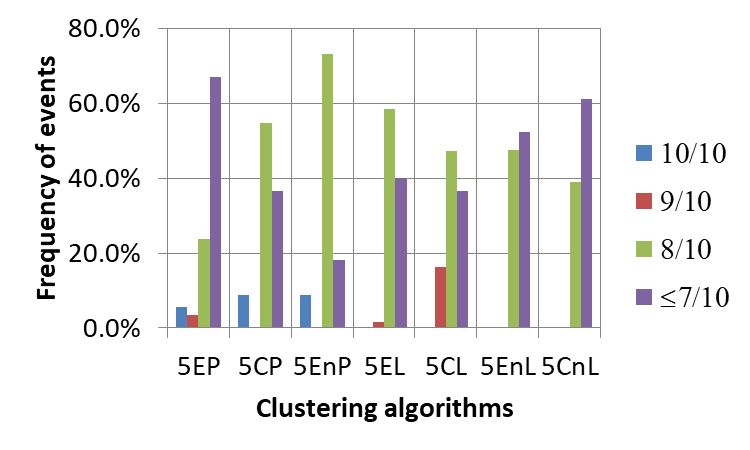}
		}
		\quad
		\subfloat[] {\label{figure14b}
			\includegraphics[width=0.75\textwidth]{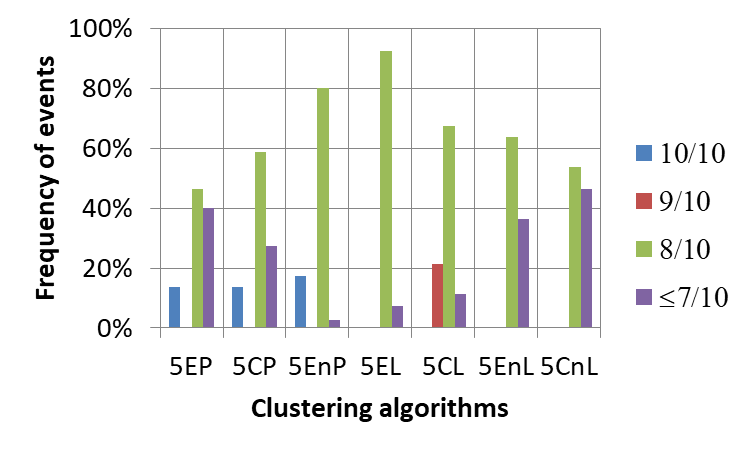}
		}
		\caption{Distributions of the efficiency of the clustering algorithms - Dataset No. 3, five clusters: (a) All possible 252 combinations of initial points; (b) Eighty initial quintets with the highest entropy}\label{figure14}}
\end{figure}

Clustering of variables into five clusters was also examined. For the assumed value of $\epsilon = 65\%$, the relation graph has five connected components.
All the various five-element subsets of the ten-element set are $\binom{10}{5}=252$.
This means that each of the above $252$ quintets can be used as initial points in the k-means algorithm.
Therefore, in clustering due to the similarity of the primary variables to the principal components, the clustering algorithms were run $252$ times for each initial quintet.
Also in clustering based on the mutual similarity of the primary variables, the clustering algorithms were run $252$ times.
The obtained results of clustering are presented in Tables \ref{table17} and \ref{table18}, as well as in Figure \ref{figure14a}.

For the case of clustering into five clusters, entropy of initial quintets was also estimated.
Fig. \ref{figure14b} shows the efficiency of clustering for seventy initial points with the highest entropy.
This bar chart shows that the efficiency of clustering also increases with the increase in entropy.
\begin{table}
	\centering
	\caption{Basic statistics of clustering procedures - Dataset No. 3, five clusters}\label{table17}
	\fontsize{9.5}{13.5}\selectfont{
		\begin{tabular}{c||c|c|c|c|c|c|c} \hline
			Statistics         & $5EP$    & $5CP$    & $5EnP$   & $5EL$    & $5CL$    & $5EnL$   & $5CnL$    \\ \hline \hline
			Average efficiency & $71.2\%$ & $75.4\%$ & $77.2\%$ & $72.7\%$ & $75.9\%$ & $70.1\%$ & $68.5\% $ \\ \hline
			Median             & $7/10$   & $8/10$   & $8/10$   & $8/10$   & $8/10$   & $7/10$   & $7/10 $   \\ \hline
			Mode               & $7/10$   & $8/10$   & $8/10$   & $8/10$   & $8/10$   & $8/10$   & $8/10 $   \\ \hline
			Minimal efficiency & $5/10$   & $5/10$   & $5/10$   & $5/10$   & $4/10$   & $4/10$   & $4/10 $   \\ \hline
			Maximal efficiency & $10/10$  & $10/10$  & $10/10$  & $9/10$   & $9/10$   & $8/10$   & $8/10 $   \\ \hline
		\end{tabular}}
\end{table}

\begin{table}
	\centering
	\caption{Distributions of the efficiency of the clustering algorithms - Dataset No. 3, five clusters}\label{table18}
	\fontsize{9.5}{13.5}\selectfont{
		\begin{tabular}{c||c|c|c|c|c|c|c} \hline
			Performance levels & $5EP$    & $5CP$    & $5EnP$   & $5EL$    & $5CL$    & $5EnL$   & $5CnL$    \\ \hline \hline
			$10/10$            & $5.6\%$  & $8.7\%$  & $8.7\%$  & $0\%$    & $0\%$    & $0\%$    & $0\% $    \\ \hline
			$9/10$             & $3.6\%$  & $0\%$    & $0\%$    & $1.6\%$  & $16.3\%$ & $0\%$    & $0\% $    \\ \hline
			$8/10$             & $23.8\%$ & $54.8\%$ & $73.0\%$ & $58.3\%$ & $47.2\%$ & $47.6\%$ & $38.9\% $ \\ \hline
			$\le 7/10$         & $67.1\%$ & $36.5\%$ & $18.3\%$ & $40.1\%$ & $36.5\%$ & $52.4\%$ & $61.1\% $ \\ \hline
		\end{tabular}}
\end{table}

\begin{figure}
	\centering
	\includegraphics[width=0.75\textwidth]{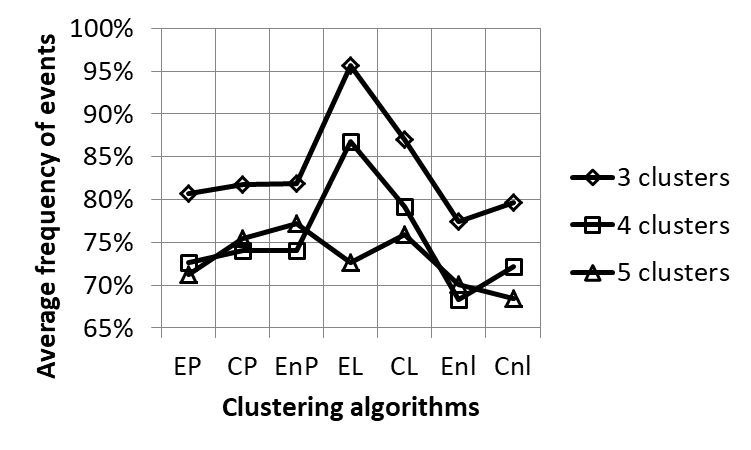}
	\caption{Average clustering efficiency for dataset No. 3}\label{figure_a3}
\end{figure}

\subsubsection{Efficiency of clustering the dataset No. 3}
Figures \ref{figure12}-\ref{figure14}, and Tables \ref{table13}-\ref{table18} show the efficiency of clustering for dataset No. 3, first for three clusters, and then for four and five clusters.
Figure \ref{figure_a3} shows the average clustering efficiency for this dataset.

Clustering of this set looks quite difficult.
When clustering into three clusters (Figure \ref{figure12}), $100\%$ efficiency is not achievable for most algorithms.
For the $3EnL$ and $3CnL$ algorithms, $90\%$ efficiency is also not achievable.
The best result is given by the $3E$L algorithm, and slightly weaker results are given by the $3CL$ algorithm.
When clustering into four clusters, the results are even lower.
There are a lot of results with an efficiency not exceeding $70\%$.
The $4EL$ spectral algorithm is the best again.
When clustering into five clusters dominate the results with clustering efficiency not exceeding $80\%$.
In contrast to the previous two cases (clustering into $3$ and $4$ clusters), when clustering due to the similarity of the primary variables to the principal components, rare cases of $100\%$ efficiency occur.

Analysis of Figure \ref{figure_a3} shows that clustering the set of primary variables into three clusters is more effective than clustering into four and five clusters.

\begin{figure}
	\centering
	\includegraphics[width=0.75\textwidth]{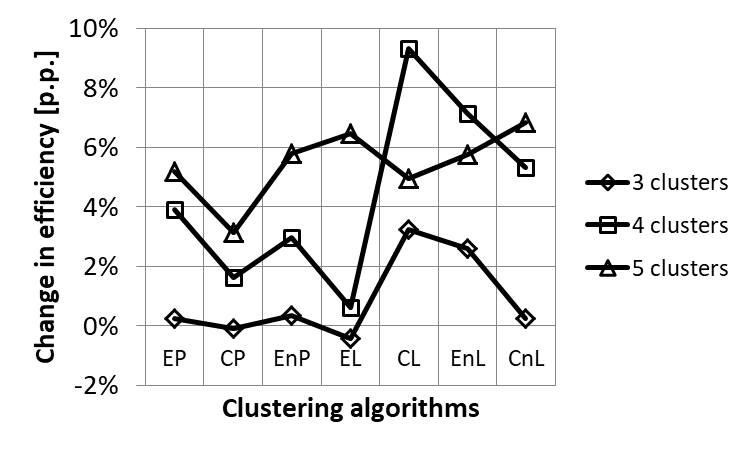}
	\caption{Change in the average efficiency of clustering (in percentage points) for initial points with the highest entropy for dataset No. 3}\label{CE_iii}
\end{figure}

Also the increase in the entropy of initial points for the k-means algorithm has a positive effect on the efficiency of clustering.
Figure \ref{CE_iii} shows how the average efficiency of clustering changes with the increase in entropy. The results are given in percentage points.
It can be noticed that for $EP$, $CP$, $EnP$, $EL$ and $CnL$ methods the average clustering efficiency into three clusters, with the increase of entropy changes very slightly (Figure \ref{figure12}).
In turn, when clustering into four or five clusters (Figures \ref{figure13} and \ref{figure14}), the number of cases with the smallest clustering efficiency decreases with the increase in entropy.

\section{The dataset No. 4}
As a last example, the dataset described in \cite{Poelstra2015} was used.
The content of this dataset is available in \cite{Poelstra2015a}.
The set contains $123$ correlated random variables.
Each variable was measured $14504$ times.

Before describing the tests performed, it should be noted that due to the large number of variables, the article will not be able to present complete tables characterizing the data.
In particular, the matrix of correlation coefficients, similarity matrix, relation matrices, Laplace matrices as well as eigenvalues and eigenvectors will not be presented.
For the PCA method, for a table describing the percentage of variation explained by the principal components, only a few rows of it will be presented.
It will also be shown only the fragment of the scree plot.

\begin{table}
	\centering
	\caption{The percentage of variance explained by the successive principal components for dataset No. 4}\label{table19}
	\fontsize{9.5}{13.5}\selectfont{
		\begin{tabular}{c|c|c|c|c} \hline
			\multirow{2}{*}{No.} & \multirow{2}{*}{Eigenvalue} & Cumulative  & Percentage of variance & Cumulative             \\
			                     &                             & eigenvalues & explained by each PC   & percentage of variance \\ \hline \hline
			$1$                  & $58.252$                    & $58.252$    & $47.4\%$               & $47.4\%$               \\ \hline
			$2$                  & $19.518$                    & $77.770$    & $15.9\%$               & $63.2\%$               \\ \hline
			$3$                  & $17.933$                    & $95.703$    & $14.6\%$               & $77.8\%$               \\ \hline
			$4$                  & $8.511$                     & $104.214$   & $6.9\%$                & $84.7\%$               \\ \hline
			$5$                  & $7.320$                     & $111.534$   & $6.0\%$                & $90.7\%$               \\ \hline
			$6$                  & $4.164$                     & $115.698$   & $3.4\%$                & $94.1\%$               \\ \hline
			$7$                  & $2.238$                     & $117.936$   & $1.8\%$                & $95.9\%$               \\ \hline
			$8$                  & $0.751$                     & $118.688$   & $0.6\%$                & $96.5\%$               \\ \hline
			$9$                  & $0.507$                     & $119.195$   & $0.4\%$                & $96.9\%$               \\ \hline
			$10$                 & $0.486$                     & $119.68$    & $0.4\%$                & $97.3\%$               \\ \hline
			$\ldots$             & $\ldots$                    & $\ldots$    & $\ldots$               & $\ldots$               \\ \hline
		\end{tabular}}
\end{table}

\begin{figure}
	\centering
	\includegraphics[width=0.75\textwidth]{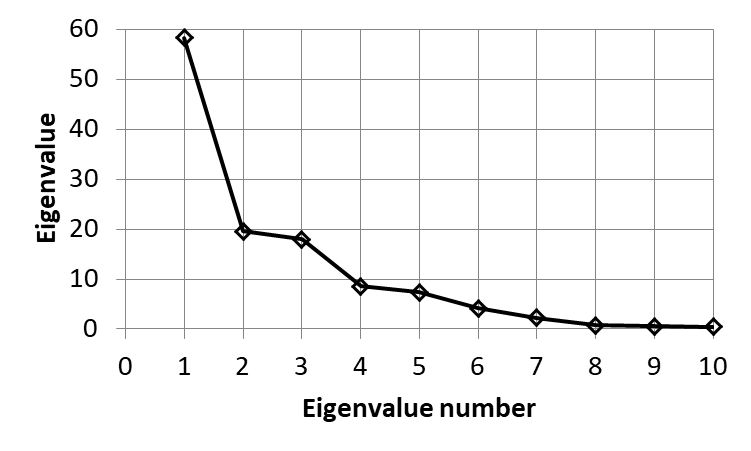}
	\caption{Fragment of a scree plot for Dataset No. 4}\label{figure15}
\end{figure}

Correlation coefficients were calculated for a set of primary variables, and then the principal components were estimated.
Table \ref{table19} shows the cumulative variance for the ten principal components.
Figure \ref{figure15} shows a scree plot for ten eigenvalues.
From Table \ref{table19}, it can be read that the five principal components explain over $90\%$ of the variance of the primary variables.
Figure \ref{figure15} shows that there are seven variables on the scree.

\subsection{Similarity of primary variables to principal components}

\begin{figure}
	\centering
	{\subfloat[] {\label{figure16a}
			\includegraphics[width=0.75\textwidth]{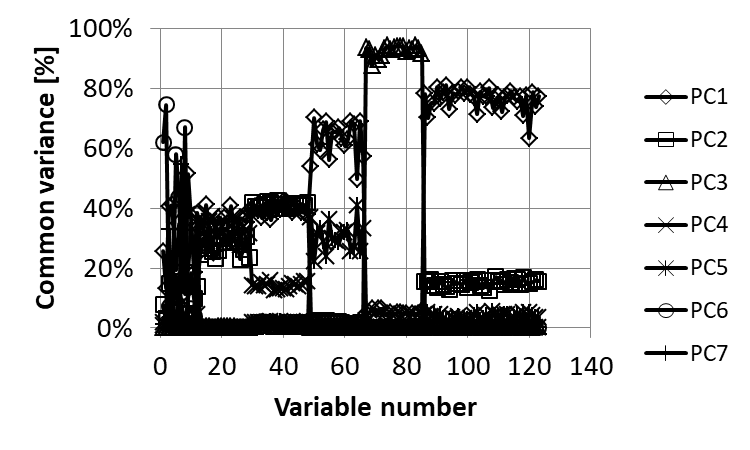}
		}
		\quad
		\subfloat[] {\label{figure16b}
			\includegraphics[width=0.75\textwidth]{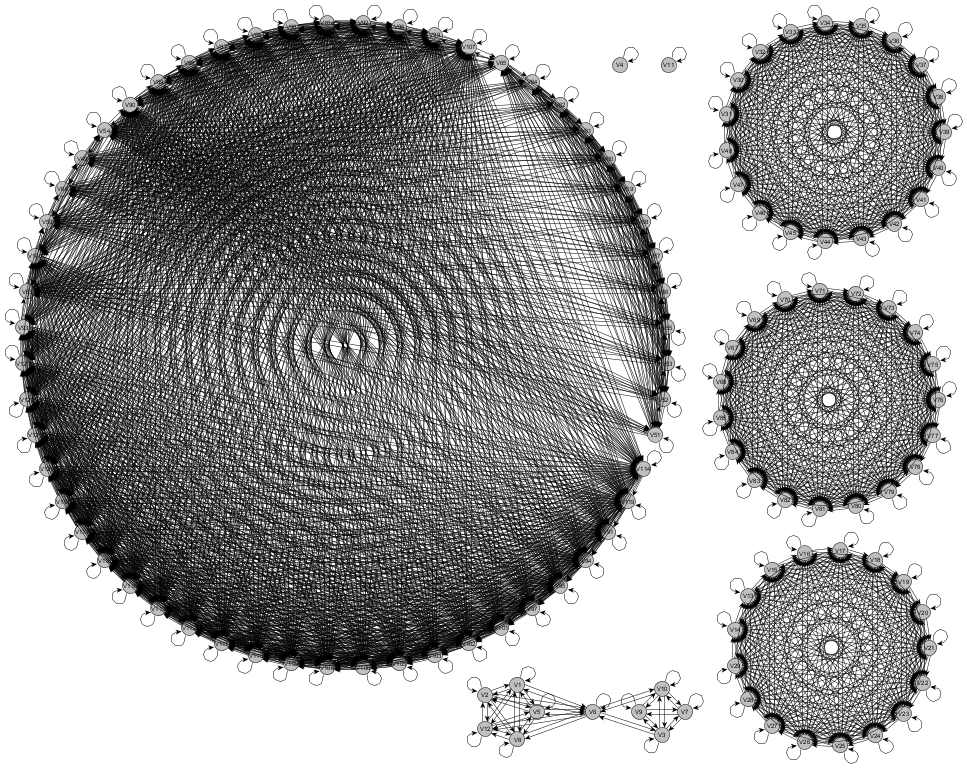}
		}
		\caption{Characterization of Dataset No. 4: (a) The similarity of the primary variables for the first seven principal components; (b) Relation graph for $\epsilon = 50\%$}\label{figure16}}
\end{figure}

The coefficients of determination between primary variables and principal components were estimated.
These coefficients measure a common variance that describes the similarity between random variables.
Fig. \ref{figure16a} presents the distributions of similarity between primary variables and the seven principal components.
In the plot it can be seen at least six zones dividing the primary variables, because of their similarity to the principal components.
In five zones (counting from the end of the chart), the determination coefficients they have values contained in narrow ranges.
On the other hand, at the beginning of the graph for the first twelve primary variables, the coefficients of determination are distributed quite chaotically.
The above observation means that due to the similarity to the principal components, the primary variables can be divided into at least six clusters.
Clustering should use at least six rows from the matrix containing the coefficients of determination between the principal components and the primary variables.
\subsection{The relation established on the level of similarity}
Based on the matrix of determination coefficients for primary variables, relations were studied.
To find relation matrices, the threshold values $\epsilon = 50\%$ and $\epsilon = 55\%$ were adopted.
For both values of $\epsilon$, a different number of connected components of the graph was obtained.
\subsubsection{The threshold value $\epsilon = 50\%$}
\begin{figure}
	\centering
	{\subfloat[] {\label{figure17a}
			\includegraphics[width=0.75\textwidth]{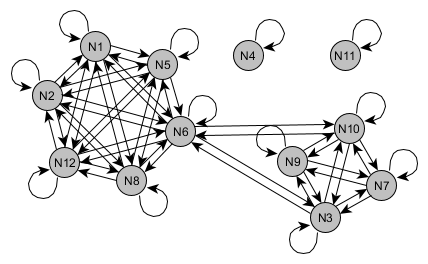}
		}
		\quad
		\subfloat[] {\label{figure17b}
			\includegraphics[width=0.75\textwidth]{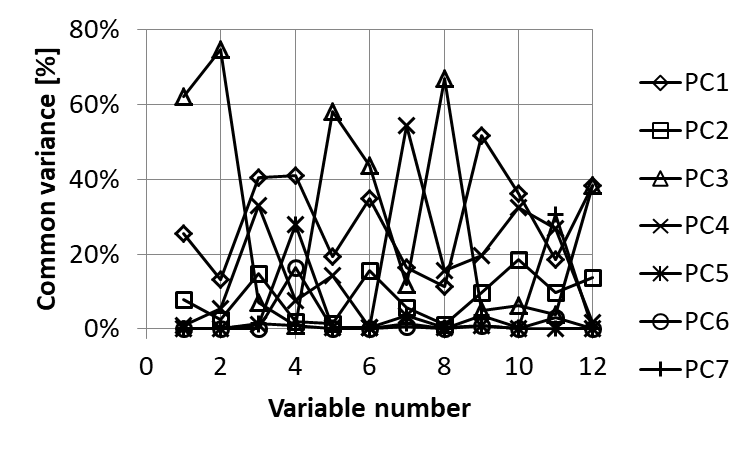}
		}
		\caption{Dataset No. 4 - variables with numbers from $1$ to $12$: (a) Subgraph for $\epsilon = 50\%$; (b) The similarity of the primary variables numbered from 1 to 12 to the principal components $PC1-PC7$}\label{figure17}}
\end{figure}

\begin{figure}
	\centering
	{\subfloat[] {\label{figure18a}
			\includegraphics[width=0.70\textwidth]{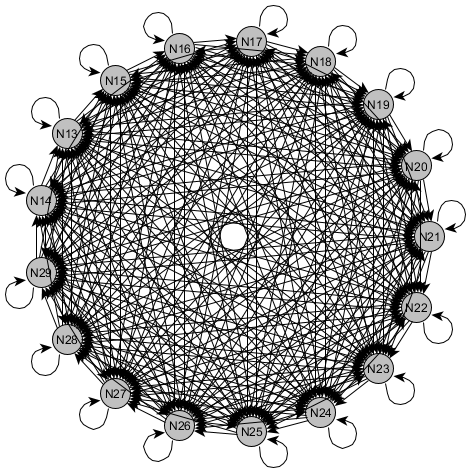}
		}
		\quad
		\subfloat[] {\label{figure18b}
			\includegraphics[width=0.75\textwidth]{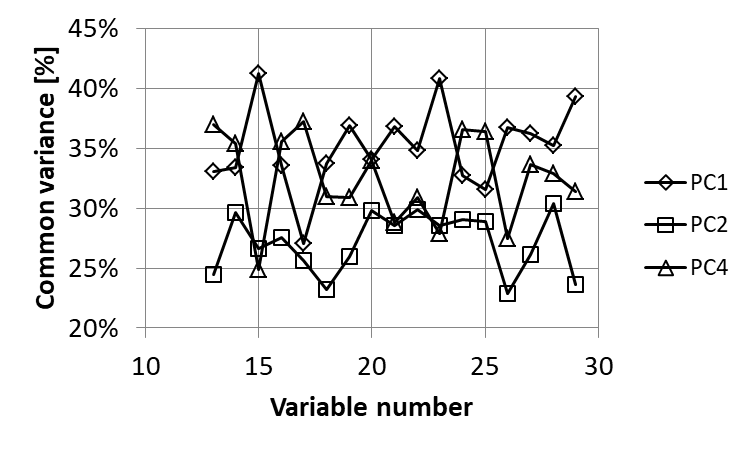}
		}
		\caption{Dataset No. 4 - variables with numbers from $13$ to $29$: (a) Subgraph for $\epsilon = 50\%$; (b) The similarity of the primary variables numbered from $13$ to $29$ to the principal components $PC1$, $PC2$ and $PC4$}\label{figure18}}
\end{figure}

\begin{figure}
	\centering
	{\subfloat[] {\label{figure19a}
			\includegraphics[width=0.70\textwidth]{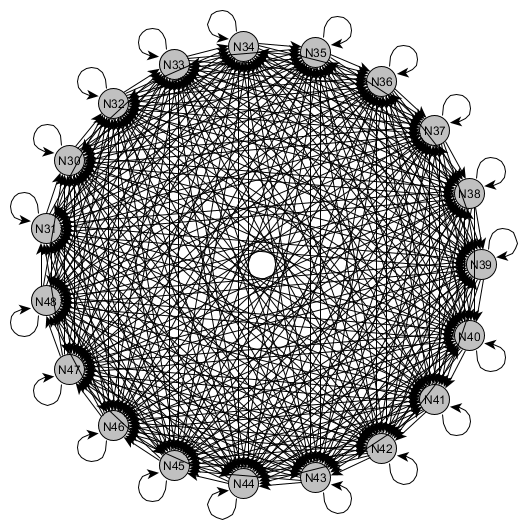}
		}
		\quad
		\subfloat[] {\label{figure19b}
			\includegraphics[width=0.75\textwidth]{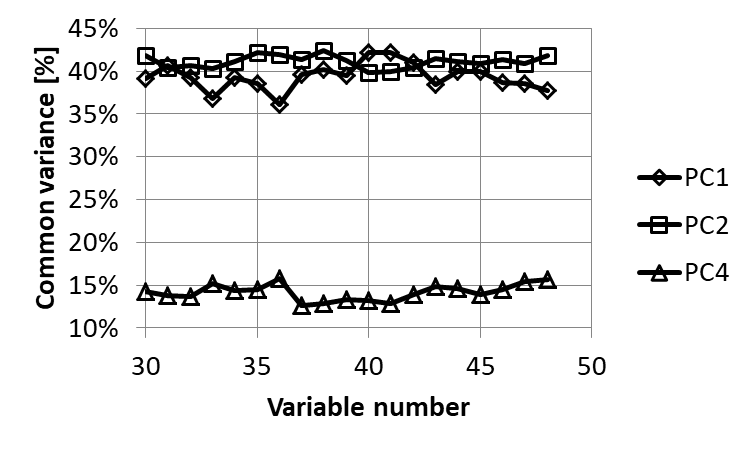}
		}
		\caption{Dataset No. 4 - variables with numbers from $30$ to $48$: (a) Subgraph for $\epsilon = 50\%$; (b) The similarity of the primary variables numbered from $30$ to $48$ to the principal components $PC1$, $PC2$ and $PC4$}\label{figure19}}
\end{figure}

\begin{figure}
	\centering
	{\subfloat[] {\label{figure20a}
			\includegraphics[width=0.70\textwidth]{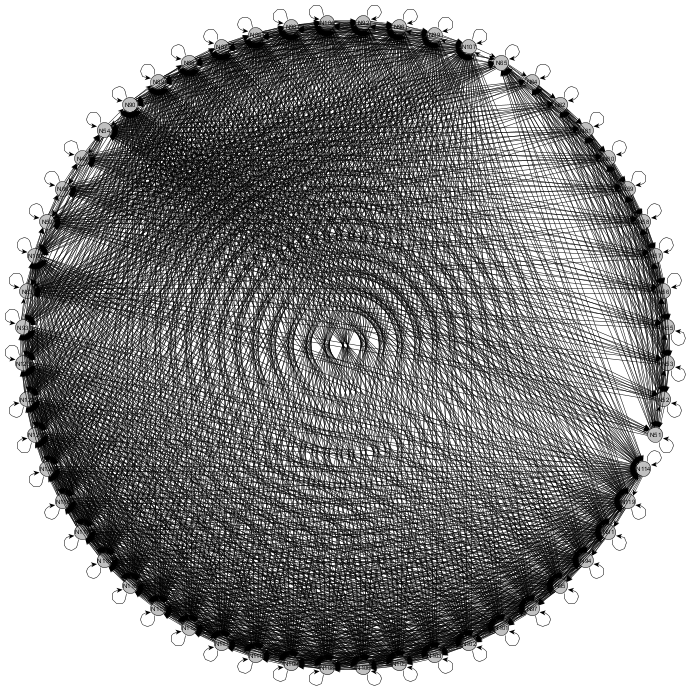}
		}
		\quad
		\subfloat[] {\label{figure20b}
			\includegraphics[width=0.75\textwidth]{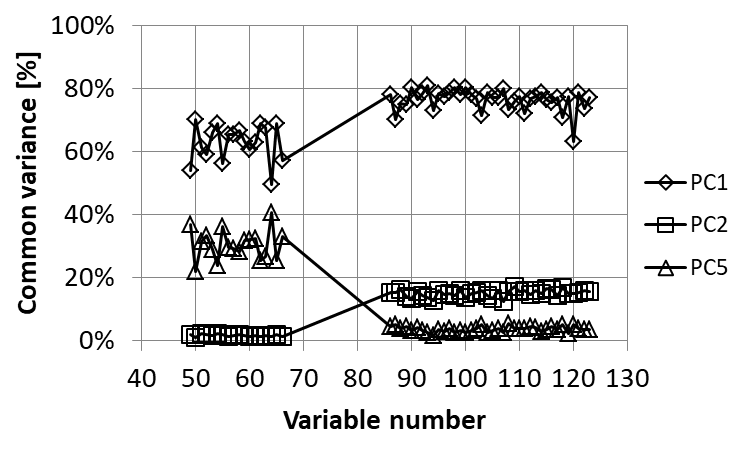}
		}
		\caption{Dataset No. 4 - variables with numbers $49-66$ and $86-123$: (a) Subgraph for $\epsilon = 50\%$; (b) The similarity of the primary variables with numbers $49-66$ and $86-123$ to the principal components $PC1$, $PC2$ and $PC5$}\label{figure20}}
\end{figure}

\begin{figure}
	\centering
	{\subfloat[] {\label{figure21a}
			\includegraphics[width=0.70\textwidth]{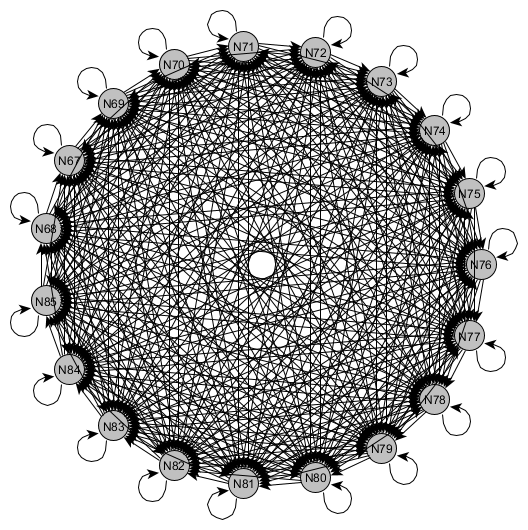}
		}
		\quad
		\subfloat[] {\label{figure21b}
			\includegraphics[width=0.75\textwidth]{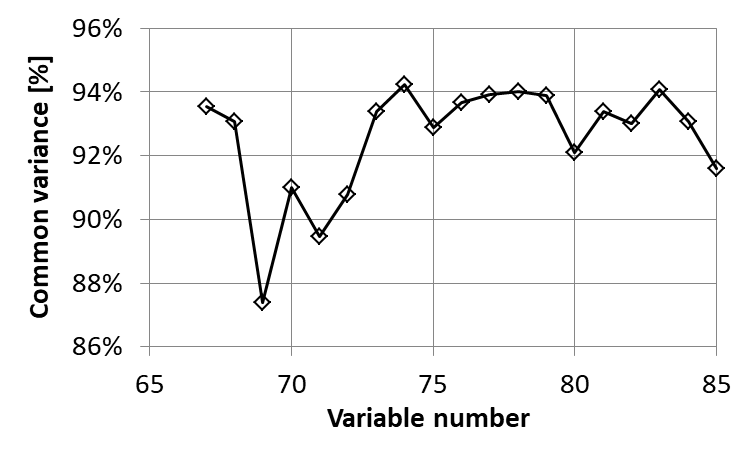}
		}
		\caption{Dataset No. 4 - variables with numbers $67-85$: (a) Subgraph for $\epsilon = 50\%$; (b) The similarity of the primary variables with numbers $67-85$ to the principal component $PC3$}\label{figure21}}
\end{figure}

For $\epsilon = 50\%$, a relation graph with seven connected components was obtained.
The graph for $\epsilon = 50\%$ is shown in Figure \ref{figure16b}.
Its connected components representing groups of mutually correlated primary variables almost perfectly represent the zones from Figure \ref{figure16a}:
\begin{itemize}
	\item The first zone in Figure \ref{figure16a} corresponds to the graph in Figure \ref{figure17a}.
	      This zone represents the similarity between variables with numbers $1-12$ and the principal components $PC1-PC7$ (Figure \ref{figure17b}).
	\item The second zone from Figure \ref{figure16a} corresponds to the similarity graph between variables with numbers $13-29$ (Figure \ref{figure18a}). Common variance of variables, which are numbered $13-29$ with the principal components $PC1$, $PC2$ as well as $PC4$ is in the range of over $20\%$ to slightly over $40\%$ (Figure \ref{figure18b}).
	\item The third zone in Figure \ref{figure16a} corresponds to the graph in Figure \ref{figure19a}, representing the similarity of the primary variables numbered $30-48$. These variables have a common variance with the principal components $PC1$ and $PC2$ at the level of approx. $40\%$, and with the principal component $PC4$ at the level of approx. $15\%$ (\ref{figure19b}).
	\item The fourth and sixth zones in Figure \ref{figure16a} corresponds to the graph representing the similarity between primary variables with numbers $49-66$ and $86-123$ (Figure \ref{figure20a}). These variables have a significant shared variance with the principal components $PC1$, $PC2$ and $PC5$ (Figure \ref{figure20b}).
	\item The fifth zone in Fig. \ref{figure16a} corresponds to the graph in Fig. \ref{figure21a} showing the mutual similarity of variables with numbers $67-85$. Primary variables have a large common variance with the $PC3$ principal component, exceeding $87\%$ (Figure \ref{figure21b}).
\end{itemize}

\subsubsection{The threshold value $\epsilon = 55\%$}
\begin{figure}
	\centering
	\includegraphics[width=8cm]{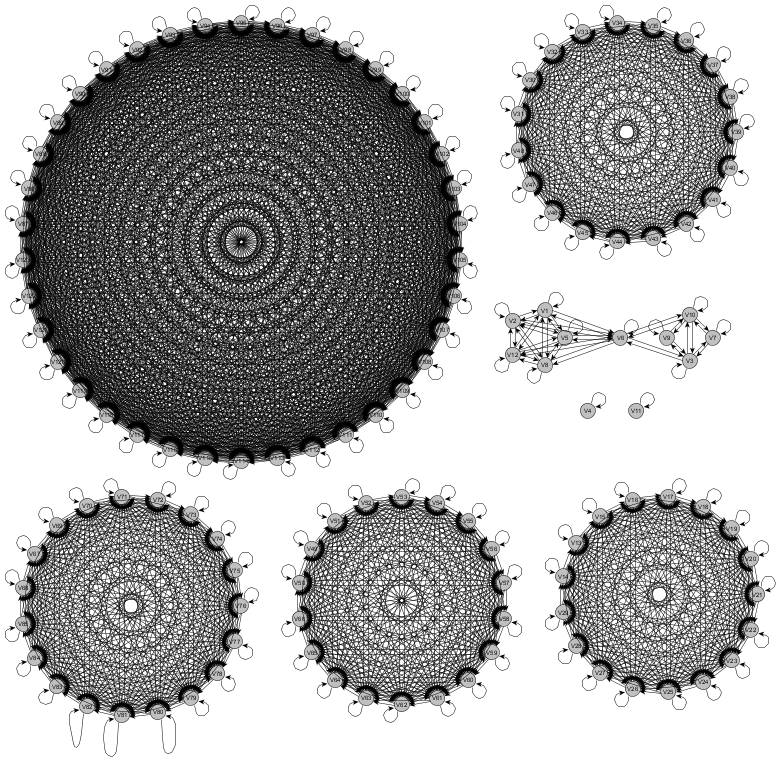}
	\caption{Dataset No. 4 - relation graph for $\epsilon = 55\%$}\label{figure22}
\end{figure}

\begin{figure}
	\centering
	{\subfloat[] {\label{figure23a}
			\includegraphics[width=0.70\textwidth]{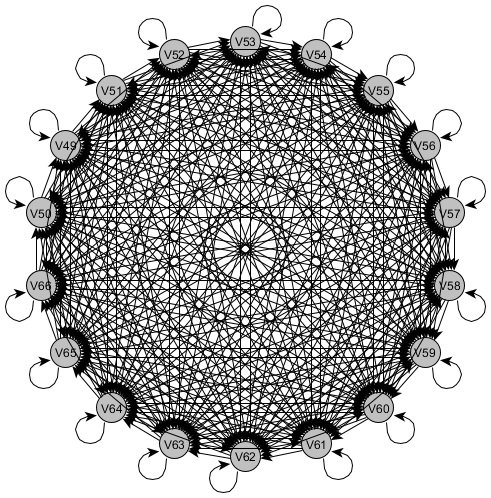}
		}
		\quad
		\subfloat[] {\label{figure23b}
			\includegraphics[width=0.75\textwidth]{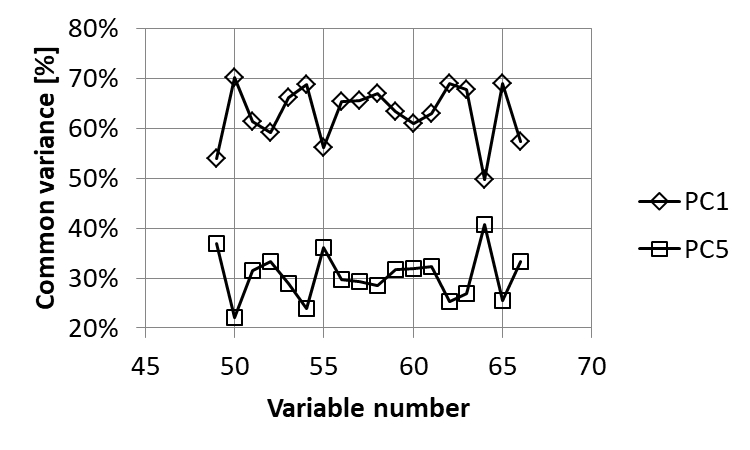}
		}
		\caption{Dataset No. 4 - variables with numbers $49-66$: (a) Subgraph for $\epsilon = 55\%$; (b) The similarity of the primary variables with numbers $49-66$ to the principal components $PC1$ and $PC5$}\label{figure23}}
\end{figure}

\begin{figure}
	\centering
	{\subfloat[] {\label{figure24a}
			\includegraphics[width=0.70\textwidth]{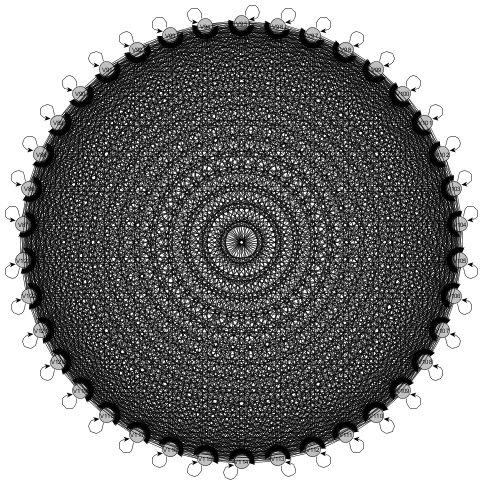}
		}
		\quad
		\subfloat[] {\label{figure24b}
			\includegraphics[width=0.75\textwidth]{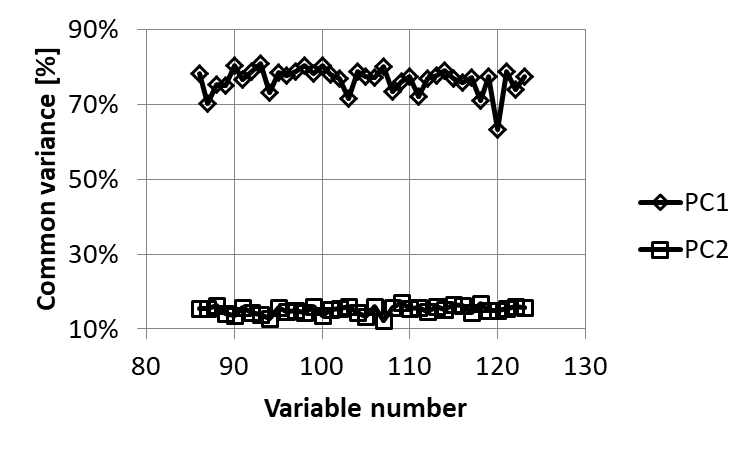}
		}
		\caption{Dataset No. 4 - variables with numbers $86-123$: (a) Subgraph for $\epsilon = 55\%$; (b) The similarity of the primary variables with numbers $86-123$ to the principal component $PC1$ and $PC2$}\label{figure24}}
\end{figure}

For $\epsilon = 55\%$, a graph of relation with eight connected components was obtained. This graph is shown in Fig. \ref{figure22}.
After increasing the value of $\epsilon$, there was a division of the connected component shown in Fig. \ref{figure20a} into two connected components, which are shown in Fig. \ref{figure23a} and Fig. \ref{figure24a}.
The first new connected component representing the mutual similarity of variables with numbers $49-66$ (Figure \ref{figure23b}) corresponds to the fourth zone in Figure \ref{figure16a}.
Primary variables with numbers $49-66$ have a significant similarity in this zone with the principal components PC1 and PC5.
The similarity to the PC1 principal component exceeds $50\%$, and the similarity to the PC5 principal component exceeds $20\%$.
On the other hand, the second connected component (Figure \ref{figure24a}) corresponds to the sixth zone in Figure \ref{figure16a}.
In this zone, the primary variables with the numbers $86-123$ have more than $70\%$ of the common variance with the principal component $PC1$, and more than $10\%$ of the common variance with the $PC2$ principal component.

\subsection{Results of clustering of primary variables}
For dataset No. 4, all of the above clustering algorithms were checked.
First, the matrix of relatios obtained from the matrix of determination coefficients with the assumed value of $\epsilon = 50\%$ was analyzed using spectral methods.
Four versions of the algorithm were considered:
\begin{itemize}
	\item $7L50\%$ - Laplacian of relation matrix, Euclidean metrics;
	\item $7CL50\%$ - Laplacian of relation matrix, cosine measure of dissimilarity;
	\item $7EnL50\%$ - Normalized Laplacian of relation matrix, Euclidean metric;
	\item $7CnL50\%$ - Normalized Laplacian of relation matrix, cosine measure of dissimilarity.
\end{itemize}
The results of clustering using the above presented versions of the algorithm were compared with the distribution of nodes in the $7Man$ graph, presented in Figure \ref{figure16b}.
Obviously, full agreement was obtained for the analyzed algorithms and the $7Man$ set.

\begin{figure}
	\centering
	{\subfloat[] {\label{figure25a}
			\includegraphics[width=0.75\textwidth]{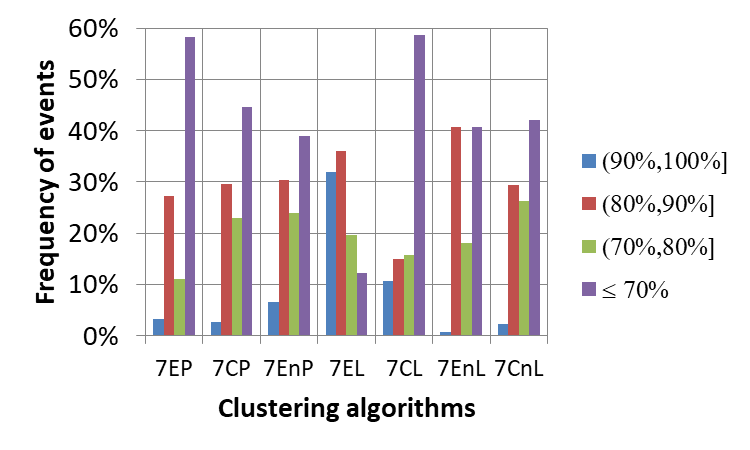}
		}
		\quad
		\subfloat[] {\label{figure25b}
			\includegraphics[width=0.75\textwidth]{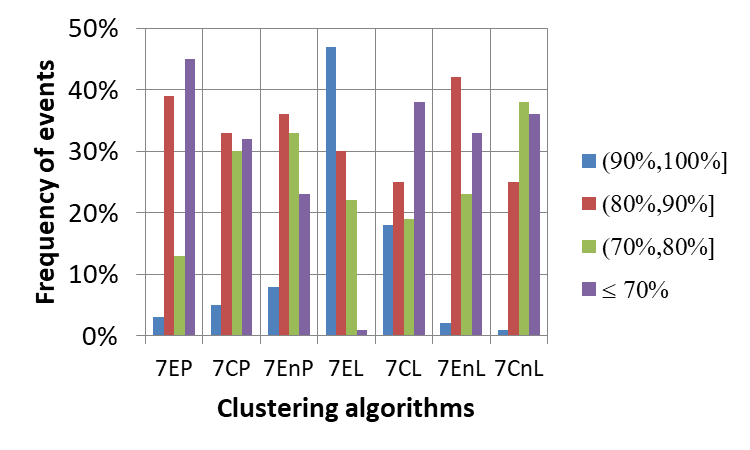}
		}
		\caption{Distributions of the efficiency of the clustering algorithms - Dataset No. 4, seven clusters: (a) Three hundred combinations of initial points; (b) One hundred initial septets with the highest entropy}\label{figure25}}
\end{figure}

When clustering due to similarity of primary variables to the principal components, and also in the case of spectral clustering with the matrix of determination coefficients in the role of similarity matrix, for $123$ variables clustered into $7$ clusters, the number of all seven-member subsets of different initial points in the k-means algorithm is is equal to $\binom{123}{7}$ and exceeds $7\cdot10^{10}$.
Therefore, unlike for fewer sets of variables, it was not possible to check the results of clustering for all possible combinations of seven-element sets of initial points.
The number of tests had to be acceptable from the point of view of the time of computation.
It was assumed that clustering would be performed for $300$ different seven-element sets of initial points.
Within a given septet of initial points, none of the seven points could be repeated.
The following versions of clustering were examined:
\begin{enumerate}
	\item For 300 different septets of initial points, variables have been clustered due to their similarity to the principal components:
	      \begin{itemize}
		      \item 7EP - Euclidean metric;
		      \item 7CP - cosine measure of dissimilarity;
		      \item 7EnP - Euclidean metric, normalized points for clustering;
	      \end{itemize}
	\item 	For 300 different septets of initial points, variables were clustered using spectral methods, treating the matrix of coefficients of determination as a similarity matrix:
	      \begin{itemize}
		      \item 7EL - Laplacian of similarity matrix, Euclidean metric;
		      \item 7CL - Laplacian of similarity matrix, cosine measure of dissimilarity;
		      \item 7EnL - normalized Laplacian of similarity matrix, Euclidean metric;
		      \item 7CnL - normalized Laplacian of similarity matrix, cosine measure of dissimilarity;
	      \end{itemize}
\end{enumerate}
Obtained results of all clustering algorithms were compared with the model marked as 7Man (Figure \ref{figure16b}).
The efficiency of the algorithms depended on both the algorithm version and the selected initial point septets.
Table \ref{table20} presents the statistics of the efficiency of all the used clustering algorithms.
For $123$ clustered variables different levels of compliance were obtained.
Table \ref{table21} shows the clustering compliance distributions for each version of the algorithm.
The results from Table \ref{table21} are presented in the bar chart in Figure \ref{figure25a}.

For the last dataset, the impact of entropy of initial points for the k-means algorithm on the efficiency of clustering was also examined.
The distribution of clustering efficiency was found for $100$ septets with the highest entropy (Fig. \ref{figure25b}).
It can be seen that in this case the efficiency of clustering was greater than for all $300$ randomly selected septets.
This means that the impact of entropy of initial points for the k-means algorithm on the results of clustering can not be ignored.

\begin{table}
	\centering
	\caption{Basic statistics of clustering procedures - Dataset No. 4, seven clusters}\label{table20}
	\fontsize{9.5}{13.5}\selectfont{
		\begin{tabular}{c||c|c|c|c|c|c|c} \hline
			Statistics         & 7EP      & 7CP      & 7EnP     & 7EL      & 7CL      & 7EnL     & 7CnL     \\ \hline \hline
			Average efficiency & $71.2\%$ & $73.9\%$ & $75.4\%$ & $82.1\%$ & $71.8\%$ & $73.3\%$ & $72.1\%$ \\ \hline
			Median             & $67.5\%$ & $74.0\%$ & $76.8\%$ & $81.3\%$ & $69.9\%$ & $76.4\%$ & $72.4\%$ \\ \hline
			Mode               & $61.0\%$ & $81.3\%$ & $81.3\%$ & $81.3\%$ & $69.9\%$ & $81.3\%$ & $83.7\%$ \\ \hline
			Minimal efficiency & $44.7\%$ & $36.6\%$ & $52.0\%$ & $59.3\%$ & $45.5\%$ & $39.8\%$ & $43.1\%$ \\ \hline
			Maximal efficiency & $95.9\%$ & $96.7\%$ & $96.7\%$ & $93.5\%$ & $95.9\%$ & $93.5\%$ & $98.4\%$ \\ \hline
		\end{tabular}}
\end{table}

\begin{table}
	\centering
	\caption{Distributions of the efficiency of the clustering algorithms - Dataset No. 4, seven clusters}\label{table21}
	\fontsize{9.5}{13.5}\selectfont{
		\begin{tabular}{c||c|c|c|c|c|c|c} \hline
			Performance levels & 7EP      & 7CP      & 7EnP     & 7EL      & 7CL      & 7EnL     & 7CnL     \\ \hline \hline
			$(90\%,100\%]$     & $3.3\%$  & $2.7\%$  & $6.7\%$  & $32.0\%$ & $10.7\%$ & $0.7\%$  & $2.3\%$  \\ \hline
			$(80\%,90\%]$      & $27.3\%$ & $29.7\%$ & $30.3\%$ & $36.0\%$ & $15.0\%$ & $40.7\%$ & $29.3\%$ \\ \hline
			$(70\%,80\%]$      & $11.0\%$ & $23.0\%$ & $24.0\%$ & $19.7\%$ & $15.7\%$ & $18.0\%$ & $26.3\%$ \\ \hline
			$\le 70\%$         & $58.3\%$ & $44.7\%$ & $39.0\%$ & $12.3\%$ & $58.7\%$ & $40.7\%$ & $42.0\%$ \\ \hline
		\end{tabular}}
\end{table}

Assuming $\epsilon = 55\%$, a relation has been obtained from the matrix of determination coefficients, whose graph has eight connected components (Figure \ref{figure22}).
As before, four versions of the algorithm were considered for the obtained relation matrix:
\begin{itemize}
	\item $8L55\%$ - Laplacian of relation matrix, Euclidean metric;
	\item $8CL55\%$ - Laplacian of relation matrix, cosine measure of dissimilarity;
	\item $8EnL55\%$ - normalized Laplacian of relation matrix, Euclidean metric;
	\item $8CnL55\%$ - normalized Laplacian of relation matrix, cosine measure of dissimilarity;
\end{itemize}
All results of the four versions of the algorithm presented above were compared with the distribution of nodes of reference graph named 8Man, shown in Figure \ref{figure22}.
Full compatibility was obtained for all four analyzed algorithms and the 8Man set.

\begin{figure}
	\centering
	{\subfloat[] {\label{figure26a}
			\includegraphics[width=0.75\textwidth]{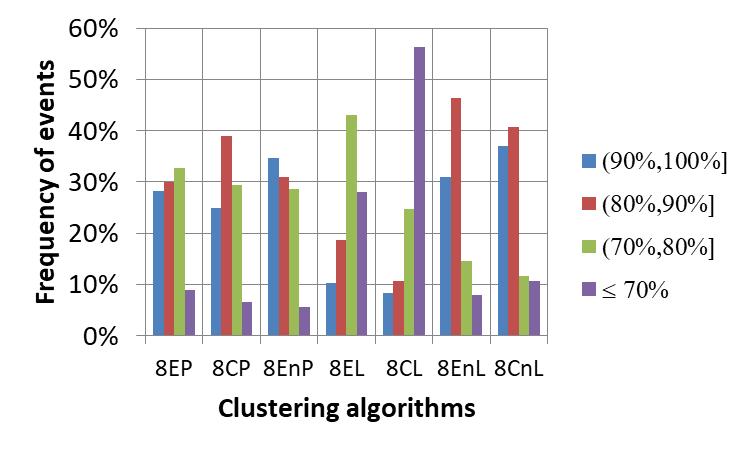}
		}
		\quad
		\subfloat[] {\label{figure26b}
			\includegraphics[width=0.75\textwidth]{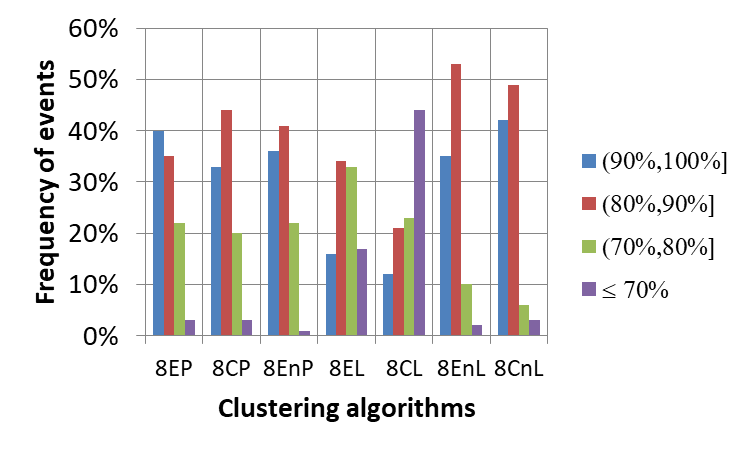}
		}
		\caption{Distributions of the efficiency of the clustering algorithms - Dataset No. 4, eight clusters: (a) Three hundred combinations of initial points; (b) One hundred initial octets with the highest entropy}\label{figure26}}
\end{figure}

When clustering due to the similarity of primary variables to principal components, and also in the case of spectral clustering with the matrix of determination coefficients in the role of similarity matrix, for $123$ variables clustered into eight clusters, the number of all eight-element subsets of different initial points is equal to $\binom{123}{8}$ and exceeds $10^{12}$.
In this case, clustering was also performed for 300 different sets of points.
The uniqueness of the initial sets as well as the uniqueness of the points in the initial sets were also assumed here.
Statistics on the efficiency of clustering are presented in Table \ref{table22}.
For the $123$ clustered variables different levels of compliance were obtained.
Table \ref{table23} shows the distribution of clustering results for all versions of the algorithm.
The results from this table are presented in a bar chart (Figure \ref{figure26a}).
\begin{table}
	\centering
	\caption{Basic statistics of clustering procedures - Dataset No. 4, eight clusters}\label{table22}
	\fontsize{9.5}{13.5}\selectfont{
		\begin{tabular}{c||c|c|c|c|c|c|c} \hline
			Statistics         & 8EP      & 8CP      & 8EnP     & 8EL      & 8CL      & 8EnL     & 8CnL     \\ \hline \hline
			Average efficiency & $82.7\%$ & $83.1\%$ & $84.3\%$ & $76.4\%$ & $71.3\%$ & $85\%$   & $84.9\%$ \\ \hline
			Median             & $88.6\%$ & $86.2\%$ & $87\%$   & $76.8\%$ & $69.1\%$ & $85.4\%$ & $87\%$   \\ \hline
			Mode               & $89.4\%$ & $88.6\%$ & $88.6\%$ & $69.1\%$ & $68.3\%$ & $84.6\%$ & $95.9\%$ \\ \hline
			Minimal efficiency & $59.3\%$ & $30.9\%$ & $55.3\%$ & $53.7\%$ & $30.9\%$ & $56.1\%$ & $30.9\%$ \\ \hline
			Maximal efficiency & $96.7\%$ & $96.7\%$ & $98.4\%$ & $91.9\%$ & $93.5\%$ & $99.2\%$ & $99.2\%$ \\ \hline
		\end{tabular}}
\end{table}

For clustering into eight clusters, entropy was estimated for all three hundred sets of initial points.
Figure \ref{figure26a} presents the distribution of clustering efficiency, obtained for one hundred octets of initial points with the highest entropy.
When comparing Fig. \ref{figure26a} and Fig. \ref{figure26b}, one can notice some improvement in the efficiency of clustering.

\begin{table}
	\centering
	\caption{Distributions of the efficiency of the clustering algorithms - Dataset No. 4, eight clusters}\label{table23}
	\fontsize{9.5}{13.5}\selectfont{
		\begin{tabular}{c||c|c|c|c|c|c|c} \hline
			Performance levels & 8EP      & 8CP      & 8EnP     & 8EL      & 8CL      & 8EnL     & 8CnL      \\ \hline \hline
			$(90\%,100\%]$     & $28.3\%$ & $25\%$   & $34.7\%$ & $10.3\%$ & $8.3\%$  & $31\%$   & $37\% $   \\ \hline
			$(80\%,90\%]$      & $30\%$   & $39\%$   & $31\%$   & $18.7\%$ & $10.7\%$ & $46.3\%$ & $40.7\% $ \\ \hline
			$(70\%,80\%]$      & $32.7\%$ & $29.3\%$ & $28.7\%$ & $43\%$   & $24.7\%$ & $14.7\%$ & $11.7\% $ \\ \hline
			$\le 70\%$         & $9\%$    & $6.7\%$  & $5.7\%$  & $28\%$   & $56.3\%$ & $8\%$    & $10.7\% $ \\ \hline
		\end{tabular}}
\end{table}

\begin{figure}
	\centering
	\includegraphics[width=0.75\textwidth]{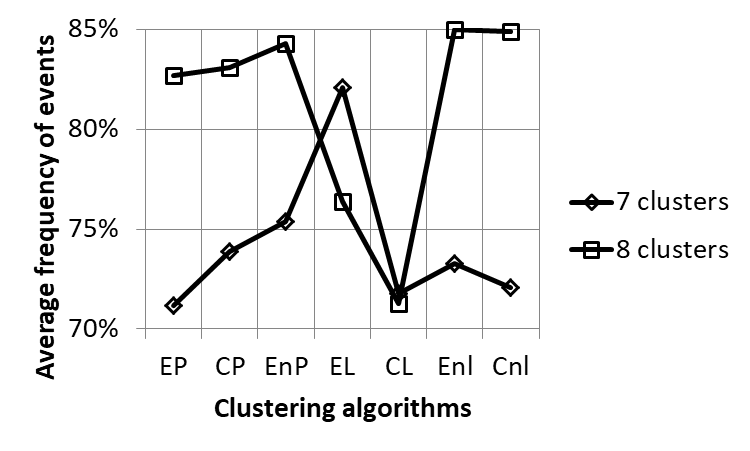}
	\caption{Average clustering efficiency for dataset No. 4}\label{figure_a4}
\end{figure}

\subsubsection{Clustering efficiency of dataset No. 4}
Tables \ref{table20}-\ref{table23} and Figures \ref{figure25} and \ref{figure26} present clustering characteristics for dataset No. 4.
Figure \ref{figure_a4} shows the average clustering efficiency obtained for this data set.

Analyzing Figures \ref{figure25} and \ref{figure26}, one can see the asymmetry in the operation of clustering algorithms.
For clustering on seven and eight clusters ($x = 7$ or $x = 8$), the $xEP$, $xCP$, $xEnP$, $xEn$L and $xCn$L algorithms work with the opposite effectiveness than the $xEL$ algorithm.
For the $xCL$ algorithm, the low efficiency for both $x = 7$ and $x = 8$ dominates.
In other words, when using the $8EP$, $8C$P, $8EnP$, $8EnL$ and $8CnL$ algorithms, results with high clustering efficiency dominate.
The $8EL$ algorithm gives results with low clustering efficiency.
For $7EP$, $7CP$, $7EnP$, $7EnL$ and $7CnL$ algorithms, results with low clustering efficiency dominate.
The $7EL$ algorithm is dominated by results with high and very high clustering efficiency.
In turn, the $8CL$ and $7CL$ algorithms work similarly.
In both cases the results with low effectiveness prevail.

Analysis of average clustering results (Figure \ref{figure_a4}) shows that $xCL$ methods ($x = 7$ or $x = 8$) give the lowest clustering results.
If the analysis did not take into account the $xEL$ and $xCL$ algorithms, one could say that clustering into eight clusters gives results clearly higher than clustering into seven clusters.

\begin{figure}
	\centering
	\includegraphics[width=0.75\textwidth]{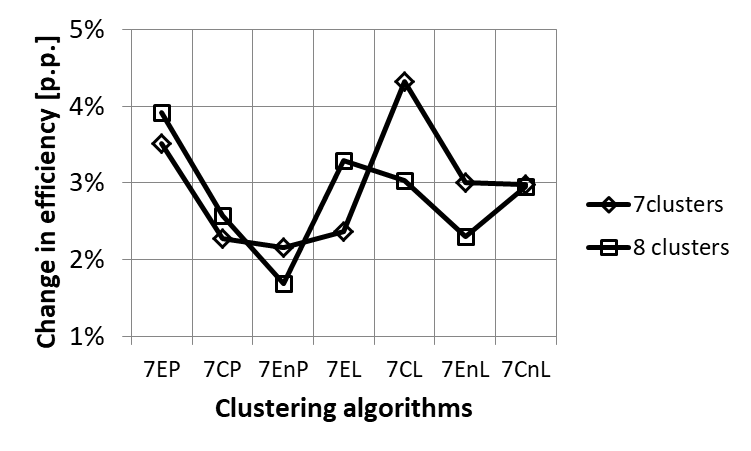}
	\caption{Change in the average efficiency of clustering (in percentage points) for initial points with the highest entropy for dataset No. 4}\label{CE_iv}
\end{figure}

When clustering the dataset No. 4, the efficiency of the algorithms also depends on entropy.
For the initial points to the k-means algorithm having greater entropy, greater clustering efficiency is obtained.
Figure \ref{CE_iv} in percentage points shows, how the average efficiency of clustering changes with the increase in entropy, for clustering into seven and eight clusters.

\section{Conclusions}
In cluster analysis, the set of elements is divided into subsets of similar elements.
In practice this is the division of a set of points located in a multidimensional space into subsets of similar points.
Clustering points, by analogy to a set of points on a plane, is called horizontal clustering.
On the other hand, points in a multidimensional space refer to variables by their coordinates.
If the variables are correlated, it can be said that they are mutually similar.
Similar variables form clusters that can be identified.
In contrast to horizontal clustering of points in space, clustering of similar variables is called vertical clustering \cite{Gniazdowski2017}.
This work uses three methods of vertical clustering:
\begin{itemize}
	\item Using the k-means method, correlated random variables were clustered due to their similarity to the principal components.
	\item Using a binary relation describing mutual similarity, variables were clustered by spectral methods.
	      The distribution of variables into clusters was manifested in the fragmentation of the graph of relation into connected components. Clustering discovers these connected components of the graph.
	\item Using a matrix of determination coefficients, variables were clustered by spectral methods.
\end{itemize}
Spectral methods were implemented in two variants. For the matrix of similarity (matrix of relation or matrix of determination coefficients), both Laplacian and normalized Laplacian were formed. Eigenproblem was solved for both Laplacians. Appropriate eigenvectors formed the space for the final use of the k-means algorithm.

Because the k-means algorithm is exploited by all the above-mentioned methods of vertical clustering, it was necessary to adopt methods of measuring distance (dissimilarity). In this work, the Euclidean metric was used, or the cosine measure of dissimilarity.

In turn, in the space of the k-means algorithm, the Euclidean distance was measured for points directly derived from the calculations, as well as for their projections on the sphere with a unit radius (the points were normalized to the unit length)

For four different datasets, the possibility of clustering using the clustering methods described above was examined.
For both points without normalization and normalized points, different ways of measuring dissimilarity (distances) were used.
Clustering was performed for different sets of initial points for each of the four datasets.
Therefore, the influence of initial points diversity (entropy) on the efficiency of clustering was also examined.

For Iris Data, one hundred percent efficiency of clustering was achieved, regardless of the methods used.
However, due to the lack of differences in results, it seems that one can not draw far-reaching conclusions and hypotheses from them.
Iris Data is important for another reason.
Thanks to its simplicity, it enables effective presentation of various versions of algorithms.
The results obtained for the remaining data sets are varied, and thus give more reasons to draw conclusions.
However, it should be noted that the conclusions obtained are not always obvious.
Analysis of the obtained results allows to state the following:
\begin{itemize}
	\item Finding connected components of the graph of similarity relation is absolutely effective and does not require additional iterations needed to find successive approximations of the centers of identified clusters.
	      The relation matrix is also the adjacency matrix of the graph.
	      The number of zero eigenvalues of Laplacian (or normalized Laplacian) is equal to the number of connected components of this graph.
	      The eigenvectors corresponding to zero eigenvalues, after their transposition, will create a space in which the k-means algorithm operates.
	      The number of points in the working space of this algorithm is equal to the number of nodes in the graph.
	      In turn, the number of different points in this space is equal to the number of connected components of the graph.
	      By identifying all points different in this space, the final centers of identified clusters are obtained.
	      Because the matrix of similarity relation is symmetric, the above conclusion is valid for any undirected graph.
	\item On the occasion of the calculations, the effect of the diversity of initial points for the k-means algorithm on the efficiency of clustering was examined.
	      On the one hand, no significant correlation between the entropy of initial points and the efficiency of clustering was noticed\footnote{Because there were no such relationships, this article does not mention their search.}.
	      Their absence means that the maximum entropy does not guarantee the maximum efficiency of clustering, and the maximum efficiency of clustering is not necessarily related to the maximum entropy of the set of initial points.
	      On the other hand, the comparison of clustering efficiency distributions for a certain population of sets of initial points, with the clustering efficiency for a subset of these sets such that the initial points have the highest entropy, allows to see that for the sets with higher entropy, slightly better clustering results are obtained.
	      Hence the conclusion that in the process of clustering can not be underestimated diversity of initial points, but rather use the sets of points with greater entropy.
	\item For Iris Data and Houses Data, maximum efficiency of clustering was obtained more often than for other datasets.
	      The relation created for these sets were equivalence relations.
	      For the remaining datasets, these were only similarity relations.
	      And so, for Iris Data, the relation had two equivalence classes, and for Houses Data, the relations had four and five equivalence classes respectively.
	      Therefore, it can be hypothesized that the maximum efficiency of clustering is more often obtained in cases where the appropriate relations are equivalence relations.
	\item In the case of Houses Data, clustering into four clusters gave better results than into five clusters.
	      It seems that the set of variables in Houses Data is more susceptible to clustering into four clusters than into five clusters.
	      Perhaps this means that clustering into four clusters is a natural clustering for variables from this dataset.
	      For dataset No. 3, clustering into three clusters was more effective than clustering into four or five clusters.
	      This may mean that dataset No. 3 is the most susceptible to clustering into three clusters.
	      Perhaps, you can also talk about the natural number of clusters for variables from this set of data.
	      In the case of dataset No. 4, if you omit the $xEL$ and $xCL$ algorithms ($x = 7$ or $x = 8$), you can see that clustering into eight clusters gives better results than clustering into seven clusters.
	      As before, variables from dataset No. 4 are more susceptible to clustering into eight clusters than into seven clusters.
	      Clustering variables in eight clusters is a natural clustering for this dataset.
	      Hence the hypothesis that for some datasets there may be a natural number of clusters, as if preferred by the dataset.
	\item The comparison between spectral methods and methods based on the analysis of the similarity of primary variables to the principal components may also be interesting.
	      For Houses Data, spectral methods are somewhat more effective.
	      For dataset No. 3, if you omit the $xEL$ and $xCL$ methods, methods based on the similarity between the primary variables and principal components are slightly more effective than the spectral methods.
	      For data set number $4$, after omitting the $xEL$ and $xCL$ methods in the analysis, there are no major differences between the two types of methods.
	\item Another thing that could have influenced the effectiveness of clustering was the method of measuring the dissimilarity (distance) in the k-means algorithm.
	      However, the analysis of clustering efficiency distributions does not allow for a clear indication that one of these methods would be better than the other.
	\item Finally, one should mention a certain incomprehensible heterogeneity in the behavior of spectral algorithms.
	      This heterogeneity consists in the fact that in some cases the $xEL$ and $xCL$ algorithms behave inversely in terms of efficiency than the $xEnL$ and $xCnL$ algorithms.
	      For dataset No. 3, this is particularly true for the $xEL$ algorithm (x=3, 4 lub 5).
	      In turn, in the case of dataset No. 4, certain heterogeneity concerns the $xEL$ and $xCL$ algorithms ($x=7$ or $x=8$).
\end{itemize}
\section*{Acknowledgments}
The main author of this work is the first author.
The role of the second author was to prepare the dataset No. 3, as well as performing principal component analysis for this dataset.

The authors would like to thank Professor Micha{\l} Grabowski for valuable suggestions regarding the possibility of using entropy as a measure of the diversity of initial points in the k-means algorithm.

\bibliography{cluster}\label{bibliography}
\bibliographystyle{unsrt}
\end{document}